\newif\ifconfver
\confvertrue        

\newif\ifonecoltab
\onecoltabtrue        

\newif\ifplainver  
\plainvertrue

\ifplainver
\confverfalse                         
\fi

\ifconfver
\documentclass[10pt,journal]{IEEEtran}
\else
\ifplainver
\documentclass[11pt]{article}
\usepackage{fullpage}
\else
\documentclass[12pt,draftcls,onecolumn]{IEEEtran}
\fi
\fi

\usepackage{calc,amsfonts,amssymb,amsmath,bm,url,color,theorem,graphicx,cite,epstopdf,nicefrac,stmaryrd}
\usepackage{psfrag,subfigure,float,epstopdf,bbold,makecell}
\usepackage[algoruled,linesnumbered]{algorithm2e}
\usepackage{multirow,comment}
\usepackage{algorithmic}
\usepackage[normalem]{ulem}
\usepackage{mdframed}
\usepackage{footnote}
\makesavenoteenv{tabular}
\definecolor{orange}{RGB}{255,107,0}

\usepackage{threeparttable,xcolor}

\theoremstyle{definition}

\newcommand{\W}{\boldsymbol{W}}

\newcommand{\D}{\boldsymbol{D}}
\newcommand{\Q}{\boldsymbol{Q}}
\newcommand{\X}{\boldsymbol{X}}
\newcommand{\C}{\boldsymbol{C}}

\newcommand{\U}{\boldsymbol{U}}

\renewcommand{\H}{\boldsymbol{H}}

\newcommand{\A}{\boldsymbol{A}}
\newcommand{\R}{\boldsymbol{R}}

\newcommand{\Z}{\boldsymbol{Z}}
\newcommand{\x}{\boldsymbol{x}}

\newcommand{\T}{{\!\top\!}}

\DeclareMathOperator{\rank}{rank}

\newtheorem{theorem}{Theorem}
\newtheorem{lemma}{Lemma}
\newtheorem{proposition}{Proposition}

\newtheorem{definition}{Definition}
\newtheorem{remark}{Remark}

\newtheorem{assumption}{Assumption}

\begin{document}

	\newcommand{\papertitle}{
		Crowdsourcing via Annotator Co-occurrence Imputation and Provable Symmetric Nonnegative Matrix Factorization
	}
	
	\newcommand{\paperabstract}{%
	Unsupervised learning of the Dawid-Skene (D\&S) model from noisy, incomplete and crowdsourced annotations has been a long-standing challenge, and is a critical step towards reliably labeling massive data.
A recent work takes a coupled nonnegative matrix factorization (CNMF) perspective, and shows appealing features: It ensures the identifiability of the D\&S model and enjoys low sample complexity, as only the estimates of the co-occurrences of annotator labels are involved.
However, the identifiability holds only when certain somewhat restrictive conditions are met in the context of crowdsourcing. Optimizing the CNMF criterion is also costly---and convergence assurances are elusive.
This work recasts the pairwise co-occurrence based D\&S model learning problem as a symmetric NMF (SymNMF) problem---which offers enhanced identifiability relative to CNMF. In practice, the SymNMF model is often (largely) incomplete, due to the lack of co-labeled items by some annotators. Two lightweight algorithms are proposed for co-occurrence imputation. Then, a low-complexity shifted {\it rectified linear unit} (ReLU)-empowered SymNMF algorithm is proposed to identify the D\&S model. 
Various performance characterizations (e.g., missing co-occurrence recoverability, stability, and convergence) and evaluations are also presented.

	}
	
	
	\ifplainver
	
	\date{\today}
	
	\title{\papertitle}
	
	\author{
		 Shahana Ibrahim and Xiao Fu
		\\ ~ \\
		School of Electrical Engineering and Computer Science\\ Oregon State University\\
		Corvallis, OR 97331, United States
		\\~\\
		(ibrahish,xiao.fu)@oregonstate.edu
	}
	
	\maketitle

	\begin{abstract}
		\paperabstract
	\end{abstract}
	
	\else
	\title{\papertitle}
	
	\ifconfver \else {\linespread{1.1} \rm \fi
		
		\author{Shahana Ibrahim and Xiao Fu
			
			\thanks{
				%
				
				
				X. Fu and S. Ibrahim are with the School of Electrical Engineering and Computer Science, Oregon State University, Corvallis, OR 97331, United States. email (xiao.fu,ibrahi)@oregonstate.edu

				This work is supported in part by the National Science Foundation under Project NSF IIS-2007836 and the Army Research Office under Project ARO W911NF-19-1-0247.

			}
		}
		
		\maketitle
		
		\ifconfver \else
		\begin{center} \vspace*{-2\baselineskip}
		\end{center}
		\fi
		
		\begin{abstract}
			\paperabstract
		\end{abstract}
		
		
		\ifconfver \else \IEEEpeerreviewmaketitle} \fi
	
	\fi
	
	\ifconfver \else
	\ifplainver \else
	\newpage
	\fi \fi

	
\section{Introduction}
\label{introduction}
Modern machine learning systems, in particular, deep learning systems, are empowered by massive high-quality {\it labeled} data \cite{najafabadi2015deep,goodfellow2016deep}.  However, massive data labeling is an arduous task---reliable data annotation requires substantial human efforts with considerable expertise, which are costly.
{\it Crowdsourcing} techniques deal with various aspects of data labeling, ranging from crowd (annotators)-based reliable annotation acquisition to effective integration of the acquired labels \cite{kittur2008crowdsourcing}. Many online platforms---such as {\it Amazon Mechanical Turk} (AMT) \cite{Buhrmester2011AmazonsMT}, {\it CrowdFlower} \cite{wazny2017crowdsourcing}, and {\it Clickworker} \cite{donna2013beyond}---have been launched for these purposes.
In platforms such as AMT, the (oftentimes self-registered) annotators do not necessarily provide reliable labels. Hence, simple integration strategies such as majority voting may work poorly \cite{karger2011iterative}.

Annotation integration is a long-existing research topic in machine learning; see, e.g., \cite{karger2011iterative,karger2013efficient,karger2014budget,karger2011budget,snow2008cheap,welinder2010multidimensional,liu2012variational, zhang2014spectral,traganitis2018blind, ibrahim2019crowdsourcing, ma2018gradient}. As an unsupervised learning task, it is often tackled from a statistical generative model identification viewpoint. The {\it Dawid-Skene} (D\&S) model \cite{dawid1979maximum} has been widely adopted in the literature. The D\&S model assumes a ground-truth label prior and assigns a ``confusion" matrix to each annotator. The entries of an annotator's confusion matrix correspond to the probabilities of the correct and incorrect annotations conditioned on the ground-truth labels. Hence, annotation integration boils down to learning the model parameters of the D\&S model. 

Perhaps a bit surprisingly, despite its popularity, the {\it identifiability} of the D\&S model had not been satisfactorily addressed until recent years. 
The model identifiability of D\&S was first shown under some special cases (e.g., binary labeling cases) \cite{ghosh2011moderates,dalvi2013aggregating,karger2013efficient}. The more general multi-class cases were discussed in \cite{zhang2014spectral,traganitis2018blind}, assuming the availability of third-order statistics of the crowdsourced annotations. 
A challenge is that the third-order statistics may be difficult to estimate reliably, especially in the sample-starved regime.
The work of \cite{ibrahim2019crowdsourcing} used pairwise co-occurrences of the annotators' responses (i.e., second-order statistics) to identify the D\&S model, which substantially improved the sample complexity, compared to the third-order statistics-based approaches.

Using second-order statistics is conceptually appealing, yet
the work in \cite{ibrahim2019crowdsourcing} still faces serious challenges in handling real large-scale crowdsourcing problems. 

\begin{enumerate}
    \item {\bf Identifiability Challenge.}
The identifiability of the methods in \cite{ibrahim2019crowdsourcing} hinges on a number of restrictive and somewhat unnatural assumptions, e.g., the existence of two {\it disjoint
} groups of annotators that both contain ``class specialists'' for all classes. 
   \item {\bf Computational Challenges.} The main algorithm in \cite{ibrahim2019crowdsourcing} is based on a {\it coupled nonnegative matrix factorization} (CNMF) approach, which has serious scalability issues. In addition, its noise robustness and convergence properties are unclear.
\end{enumerate}

\subsection{Contributions}
To overcome the challenges, we take a deeper look at the pairwise co-occurrence (second-order statistics) based D\&S model identification problem and offer an alternative approach. 
Our contributions are as follows:

\paragraph{Enhanced Identifiability.} We reformulate the pairwise annotator co-occurrence based D\&S model identification problem as a {\it symmetric nonnegative matrix factorization} (SymNMF) problem in the presence of missing ``blocks''---which are caused by the absence of some annotator co-occurrences (since not all annotators label all items). 
We show that if the missing co-occurrences can be correctly imputed, solving the subsequent SymNMF problem uniquely identifies the D\&S model under much relaxed conditions relative to those in \cite{ibrahim2019crowdsourcing}. 

\paragraph{Co-occurrence Imputation Algorithms.}
We offer two custom and recoverability-guaranteed co-occurrence imputation algorithms. 
First, we take advantage of the fact that annotator dispatch is under control in some crowdsourcing problems and devise a co-occurrence imputation algorithm using simple operations like singular value decomposition (SVD) and least squares (LS).
Second, we consider a more challenging scenario where annotator dispatch is out of reach and some observed co-occurrences are unreliably estimated.
Under this scenario, we propose an imputation criterion that is provably robust to outlying co-occurrence observations. We also propose a lightweight iterative algorithm under this setting.

\paragraph{Fast and Provable SymNMF Algorithm.} 
To identify the D\&S model from the co-occurrence-imputed SymNMF model, we propose an algorithm that is a modified version of the subspace-based SymNMF algorithm in \cite{huang2014non}. The algorithm in \cite{huang2014non} is known for its simple updates and empirically fast convergence, but understanding to its convergence properties has been elusive. We replace the nonnegativity projection step in the algorithm by a shifted {\it rectified linear unit} (ReLU) operator. Consequently, we show that the new algorithm converges {\it linearly} to the desired D\&S model parameters under some conditions---while maintaining almost the same lightweight updates. We also show that the new algorithm is provably robust to noise. 
Note that the SymNMF is an NP-hard problem, and analyzing the model estimation accuracy is challenging.
Our convergence result fills this gap.

\paragraph{Notation.} A summary of notations used in this work can be found in the supplementary material.


\section{Background}
We focus on the D\&S model identification problem in the context of crowdsourced data annotation.
Consider $N$ data items that are denoted as $\{ \bm f_n\}_{n=1}^N$, where $\bm f_n\in\mathbb{R}^D$ is a feature vector representing the data item. The corresponding (unknown) ground-truth labels are $\{y_n\}_{n=1}^N$, where $y_n\in\{1,2,\ldots,K\}$ and $K$ is the number of classes. These unlabeled data items are crowdsourced to $M$ annotators. Each annotator labels a subset of the $N$ items, and the subsets could be overlapped.
Annotator $m$'s response to item $n$ is denoted as $X_m(\bm f_n)\in\{1,\ldots,K\}$. Our interest lies in integrating $\{X_m(\bm f_n)\}_{m\in {\cal I}_n }$, where ${\cal I}_n$ is the index set of the annotators who co-labeled item $n$, to estimate the ground-truth $y_n$ for all $n\in[N]$. 
Note that na\"ive integration methods such as majority voting often work poorly \cite{karger2011iterative,salk2017limitations}, as the annotators are not equally reliable and the annotations from an annotator are normally (heavily) incomplete.

\subsection{Dawid-Skene Model}
Under the D\&S model, the ground-truth data label and the $M$ annotators' responses are assumed to be discrete random variables (RVs), which are denoted by $Y$ and $\{X_m\}_{m=1}^M$, respectively.
A key assumption is that the $X_m$'s are conditionally independent given $Y$, i.e.,
\begin{align}\label{eq:naiveBayes}
{\sf Pr}(k_1,\ldots,k_M) 
=\sum_{k=1}^K\prod_{m=1}^M{\sf Pr}(k_m|k){\sf Pr}(k),   
\end{align}
where $k_m,k\in[K]$, and we have used the shorthand notation ${\sf Pr}(k_1,\ldots,k_M)={\sf Pr}(X_1=k_1,\ldots,X_M=k_M)$, $ {\sf Pr}(k)={\sf Pr}(Y=k)$ and ${\sf Pr}(k_m|k)={\sf Pr}(X_m=k_m|Y=k)$. On the right-hand side, ${\sf Pr}(X_m=k_m|Y=k)$ when $k_m\neq k$ is referred to as the {\it confusion probability} of annotator $m$, and ${\sf Pr}(Y=k)$ for $k\in[K]$ is the prior probability mass function (PMF) of the ground-truth label. Identifying the D\&S model, i.e., the confusion probabilities and the prior, allows us to build up a maximum {\it a posteriori} probability (MAP) estimator for $y_n$.

\subsection{Related Work - From EM to Tensor Decomposition}
The work in \cite{dawid1979maximum} offered an expectation maximization (EM) algorithm for identifying the D\&S model, while no convergence or model identifiability properties were understood at the time.
Later on, a number of works considered special cases of the D\&S model and offered identifiability supports. For example, under the ``one coin'' model, the work in \cite{ghosh2011moderates} established the identifiability of the D\&S model via SVD. This work considered cases with binary labels and no missing annotations (i.e., all annotators label all data items).  The work in \cite{dalvi2013aggregating} extended the ideas to more realistic settings where missing annotations exist. Around the same time, other approaches, e.g., random graph theory \cite{karger2013efficient} and iteratively reweighted majority voting \cite{li2014error,li2015theoretical}, were also used for D\&S model identification. In \cite{welinder2010multidimensional,whitehill2009whose,zhou2012learning,zhou2015regularized}, the D\&S model was extended by modeling aspects such as ``item difficulty'' and ``annotator ability''. 
However, the identifiability of these more complex models are unclear. 


The work in \cite{traganitis2018blind,zhang2014spectral} addressed D\&S model identification with multi-class labels using third-order statistics of the annotations.
The D\&S model identification problem was recast as tensor decomposition problems.
Consequently, the uniqueness of tensor decomposition was leveraged for provably identifying the D\&S model. The key challenge lies in the sample complexity for accurately estimating the third-order statistics. The difficulty of accurately estimating the third-order statistics may make the tensor methods struggle, especially in the annotation-starved cases.
Tensor decomposition may also be costly in terms of computation; see \cite{fu2020block,fu2020computing}.

\color{black}
\subsection{Recent Development - Coupled NMF}\label{sec:coupled_nmf}
Our work is motivated by a recent development in \cite{ibrahim2019crowdsourcing}.
The work in \cite{ibrahim2019crowdsourcing} used only the estimates of ${\sf Pr}(X_m=k_m,X_j=k_j)$'s, which are much easier to estimate compared to third-order statistics in terms of sample complexity \cite{han2015minimax}. 
Define the {\it confusion matrix} of annotator $m$ (denoted by $\A_m\in\mathbb{R}^{K\times K}$) and the prior PMF $\bm \lambda \in \mathbb{R}^K$ as follows:
$
\A_m(k_m,k)={\sf Pr}(X_m=k_m|Y=k)$ and $\bm \lambda(k)={\sf Pr}(Y=k)$.
Then, by the conditional independence in \eqref{eq:naiveBayes},
the co-occurrence matrix of annotators $m,j$ can be expressed as
\begin{equation}
\bm R_{m,j}=\A_m\bm D\bm A_j^\T,
\end{equation}
where $\bm R_{m,j}(k_m,k_j)={\sf Pr}(X_m=k_m,X_j=k_j)=\sum_{k=1}^K \bm \lambda(k) \A_m(k_m,k)\A_j(k_j,k)$ and $\bm D={\sf Diag}(\bm \lambda)$. In practice, if two annotators $m$ and $j$ co-label a number of items, then the corresponding $\R_{m,j}$ can be estimated via sample averaging, i.e., 
\begin{equation}\label{eq:Rest}
\begin{aligned}
          \widehat{\R}_{m,j }(k_m ,k_j) =
     &\frac{1}{|{\cal S}_{m,j}|} \sum_{n\in{\cal S}_{m,j}} \mathbb{I}\left[X_{m}({\bm f}_n) = k_m , X_j(\bm f_n)= k_j\right],
\end{aligned}
\end{equation}
where $\mathbb{I}[\cdot]$ is an indicator function, $k_m, k_j \in [K]$, ${\cal S}_{m,j}\subseteq [N]$ holds the indices of $\bm f_n$'s that are co-labeled by annotators $m$ and $j$, and $|{\cal S}_{m,j}|$ is the number of items annotators $m$ and $j$ co-labeled.

Note that not all $\bm R_{m,j}$'s are available since some annotators $m,j$ may not have co-labeled any items. Hence, the problem boils down to estimating $\A_m$'s and $\bm \lambda$ from $\bm R_{m,j}$'s where $(m,j)\in\bm \varOmega$ with $m\neq j$, where $\bm \varOmega$ is the index set of the observed pairwise co-occurrences.

The work in \cite{ibrahim2019crowdsourcing} considered the following CNMF criterion:
\begin{subequations}\label{eq:feas}
	\begin{align}
	{\rm find}&~\{\A_m\}_{m=1}^M,\bm \lambda\\
	{\rm s.t.}&~\bm R_{m,j}=\bm A_m\D\A_j^\T,~(m,j)\in\bm \varOmega,\\
	&~\bm A_m\geq \bm 0,\bm 1^\T\A_m=\bm 1^\T,~\bm 1^\T\bm \lambda=1,\bm \lambda\geq \bm 0,
	\end{align}
\end{subequations}
where the constraints are added per the PMF interpretations of the columns of $\A_m$ and $\bm \lambda$. The word ``coupled'' comes from the fact that the co-occurrences are modeled by $\A_m\bm D\bm A_j^\T$ with shared (coupled) $\A_m$'s and $\A_j$'s.
It was shown in \cite{ibrahim2019crowdsourcing} that under some conditions,  $\A_m^\star =\A_m\bm \Pi$ and $\bm D^\star=\bm D\bm \Pi$, where $\A_m^\star$ and $\bm D^\star$ are from {\it any} optimal solution of \eqref{eq:feas} and $\bm \Pi$ is permutation matrix.
Specifically, assume that there exist two subsets of the annotators, indexed by ${\cal P}_1$ and ${\cal P}_2$, where ${\cal P}_1\cap {\cal P}_2=\emptyset$ and ${\cal P}_1\cup {\cal P}_2 \subseteq [M]$. Let 
\begin{equation}
\begin{aligned}
             &\H^{(1)} :=[\A_{m_1}^\top,\ldots,\A^\top_{m_{|{\cal P}_1|}}]^\top,\\ &\H^{(2)}:=[\A_{j_1}^\top,\ldots,\A^\top_{j_{|{\cal P}_2|}}]^\top,
\end{aligned}
\end{equation}
where $m_t\in{\cal P}_1$ and $j_\ell\in{\cal P}_2$. The most important condition used in \cite{ibrahim2019crowdsourcing} is that both $\H^{(1)}$ and $\H^{(2)}$ satisfy the \textit{sufficiently scattered condition} (SSC) (cf. Definition~\ref{as:ss}).


\paragraph{Identifiability Challenge.} 
One of our major motivations is that the conditions for D\&S identification in \cite{ibrahim2019crowdsourcing} are somewhat restrictive.
To understand this, it is critical to understand the {\it sufficiently scattered condition} (SSC) that is imposed on $\H^{(1)}$ and $\H^{(2)}$. SSC is widely used in the NMF literature \cite{fu2018nonnegative,fu2018identifiability,fu2015blind,fu2016robust,huang2014non,gillis2020nonnegative} and is defined as follows:
\begin{definition}\label{as:ss} (SSC)
	Any nonnegative matrix $\bm Z\in\mathbb{R}^{I\times K}_+$ satisfies the SSC if the conic hull of $\Z^\T$ (i.e., ${\sf cone}(\bm Z^\T)$) satisfies (i) 
		${\cal C}\subseteq {\rm cone}\{\bm Z^{\top}\}$
		where 	$\mathcal{C} = \{\x \in\mathbb{R}^K~|~ \x^\T\mathbf{1} \geq \sqrt{K-1}\|\x\|_2\}$ and (ii) ${\sf cone}\{\bm Z^{\top}\} \not\subseteq {\sf cone}\{\bm Q\} $ for any orthonormal $\Q\in\mathbb{R}^{K\times K}$ except for the permutation matrices.
\end{definition}
The SSC reflects how spread the rows of $\Z$ are in the
nonnegative orthant.
\begin{figure}
    \centering
    \includegraphics[ scale=0.3]{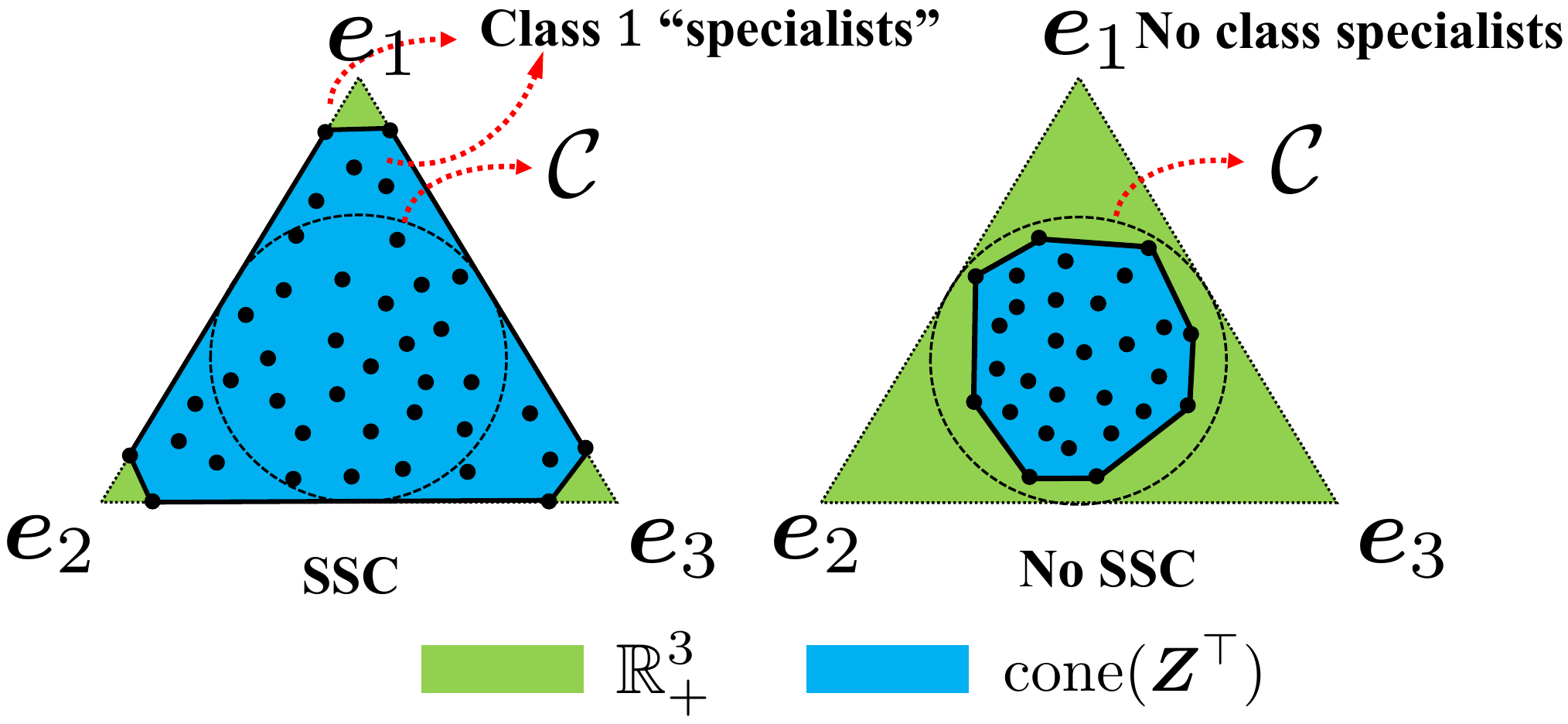}
    \caption{Illustration of $\bm Z$ satisfying the SSC and violating the SSC, respectively. The dots are the rows of $\bm Z$; the circle is the second-order cone ${\cal C}$; and the blue region with the dots is the conic hull of $\bm Z^\T$. To make $\Z$ satisfy the SSC, the blue region should cover the circle.}\label{fig:SSC}
\end{figure}
The illustration of the SSC is shown in Fig.~\ref{fig:SSC}. 
To satisfy the SSC, some rows of $\H^{(i)}$ need to be not too far away from the extreme rays of nonnegative orthant (i.e., the unit vectors $\bm e_1,\ldots,\bm e_K$). This means that some rows of certain $\A_m$'s are close to be unit vectors. If $\|\A_m(k,:)-\bm e_k^\T\|_2$ is small, it means that 
$ |\bm A_m(k,k)-1|=|{\sf Pr}(X_m=k|Y=k) -1 |$
is small; i.e., 
annotator $m$ rarely confuses data from other classes with the ones from class $k$ and is a ``class specialist'' for class $k$.
In other words, both $ \H^{(1)}$ and $\H^{(2)}$ satisfying the SSC means that the disjoint ${\cal P}_1$ and ${\cal P}_2$ both contain ``class specialists'' for all $K$ classes---which may not be a trivial condition to fulfil in practice. 


\paragraph{Computational Challenges.} 
The work in \cite{ibrahim2019crowdsourcing} recast the problem in \eqref{eq:feas} as a {\it Kullback-Leiber} (KL) divergence based model fitting problem with constraints. The iterative algorithm there often produces accurate integrated labels, but some major challenges exist. 
First, the method is hardly scalable. When the number of annotators grows, the runtime of the CNMF algorithm increases significantly.
Second, due to the nonconvexity, it is unclear if the algorithm converges to the optimal ground-truth $\A_m$ and $\bm D$, even if there is no noise.
Third, when there is noise, it is unclear how it affects the model identifiability, since the main theorem of \cite{ibrahim2019crowdsourcing} for CNMF was derived under the ideal case where no noise is present. The work in \cite{ibrahim2019crowdsourcing} offered a fast greedy algorithm for noisy cases. However, the conditions for that algorithm to work is much more restrictive, and the greedy algorithm's outputs are less accurate, as will be seen in the experiments.

\section{Proposed Approach}
Because of the appeal of its sample complexity, 
we offer an alternative way of using pairwise co-occurrences, while circumventing the challenges in the CNMF approach.
Assume that all $\bm R_{m,j}=\A_m\D\A_j^\T$ are available (including the cases where $m=j$).  Then, one can construct
\begin{align}  {\bm X}&= \begin{bmatrix} \bm R_{1,1}& \ldots & {\bm R}_{1,M} \\ \vdots & \ddots &\vdots \\ {\bm R}_{M,1} & \ldots & {\bm R}_{M,M} \end{bmatrix} =\H\H^\T,
\label{eq:Xbig} \end{align}
where $\H=[{\bm A}_{1}^\T,\ldots.{\bm A}_{M}^\T]^\T {\bm D}^{1/2}$.
Note that the above is a symmetric NMF model since $\H\geq \bm 0$ by the physical meaning of the $\A_m$'s and $\bm D$.
It is known that the model is unique if $\bm H$ satisfies the SSC \cite{huang2014non}. Hence, we have the following:
\begin{proposition}\label{col:ident}
Assume that $\H$ in \eqref{eq:Xbig} satisfies the SSC, ${\rm rank}(\H)=K$, and that $\X$ in \eqref{eq:Xbig} is available. Then, all the confusion matrices and the data prior in the D\&S model can be identified uniquely by SymNMF of $\X$ up to common column permutations; i.e.,
${\A}^\star_m=\A_m\bm \varPi$, $\forall m\in[M]$, ${\bm \lambda}^\star=\bm \varPi^\T \bm \lambda$,
where $\bm \varPi$ is a permutation matrix and ${\A}^\star_m$ denotes the $m$th column-normalized (w.r.t. the $\ell_1$ norm) block in ${\H}^\star$ that is any solution satisfying $\X=\H^\star(\H^\star)^\T$ with $\H^\star\geq \bm 0$.
\end{proposition}
The proof is a straightforward application of Theorem~4 in \cite{huang2014non}. 

\paragraph{Improved Identifiability Conditions.} Unlike in the CNMF approach, in Proposition~\ref{col:ident}, the SSC condition is imposed on $\H \in\mathbb{R}^{MK\times K}$ instead of $\H^{(i)}\in\mathbb{R}^{|{\cal P}_i|K\times K}$ for $i=1,2$.
Consequently, one only needs {\it one} set of class specialists from all the annotators instead of {\it two} sets of specialists from disjoint groups of the annotators. In addition, since $\H$ is potentially much ``taller'' than $\H^{(i)}$ (since it is often the case that $|{\cal P}_i|\ll M$), the probability that it attains the SSC condition is also much higher than that of the $\H^{(i)}$'s.
In fact, it was shown that, under a certain probabilistic generative model, for a nonnegative matrix $\bm Z\in\mathbb{R}^{I\times K}$ to satisfy the SSC with $\varepsilon$-sized error (see the detailed definition in \cite{ibrahim2019crowdsourcing}) with probability of at least $1-\mu$, one needs that $I \ge \Omega \left(\frac{(K-1)^2}{\kappa^{2(K-2)}\varepsilon^{2}}{\rm log}\left(\frac{K(K-1)}{\mu}\right)\right)$, where $\kappa>\varepsilon$ is a constant---which also asserts that $\H$ has a better chance to attain the SSC compared to the $\H^{(i)}$'s.


\paragraph{Missing Co-occurrences.} The rationale for enhancing the D\&S model identifiability using the SymNMF model in \eqref{eq:Xbig} is clear---but the challenges are also obvious. In particular, many blocks ($\bm R_{m,j}$'s) in $\X$ can be missing for different reasons. First $\bm R_{m,m}=\A_m\bm D\bm A_m^\T$ for $m=1,\ldots,M$ do not have physical meaning and thus cannot be observed or directly estimated from the data through sample averaging. Second, if annotators $m,j$ never co-labeled any items, the corresponding co-occurrence matrix $\bm R_{m,j}$ is missing.

Note that when $MK\gg K$, $\X=\H\H^\T$ is a low-rank factorization model. Imputing the unobserved $\bm R_{m,j}$'s amounts to a {\it low-rank matrix completion} (LRMC) problem \cite{Candes11}. Nonetheless, existing LRMC recoverability theory and algorithms are mostly designed under the premise that the entries (other than blocks) are missing uniformly at random---which do not cover our block missing case.
In the next two subsections, we offer two co-occurrence imputation algorithms that are tailored for the special missing pattern in the context of crowdsourcing.

\subsection{Designated Annotators-based Imputation} \label{sec:desg_imputation}
In crowdsourcing, annotators can sometimes be dispatched by the label requester.
Hence, some annotators may be {\it designated} to co-label items with other annotators.
To explain, consider the case where $\bm R_{m,n}=\A_m{\bm D}\A_{n}^\T $ is missing, i.e., $(m,n) \notin \bm \varOmega$. Assume that two annotators (indexed by $\ell$ and $r$) can be designated to label items that were labeled by annotators $m$ and $n$. This way, $\R_{m,r}$, $\R_{n,\ell}$ and $\R_{\ell ,r}$ can be {\it made} available (if there is no estimation error).
Construct $\bm C = [\R_{m,r}^{\top}, \R_{\ell, r}^{\top}]^{\top}$.
Consider the thin SVD of $\C$, i.e.,
\begin{align} \label{eq:constructC}
\bm C &= [\bm U_m^{\top}, \bm U_{\ell}^{\top}]^{\top}\bm \Sigma_{m,\ell,r}\bm V_{r}^{\top}.
\end{align}
When ${\rm rank}(\A_m)={\rm rank}(\D)=K$ for all $m\in[M]$,
it is readily seen that
$
\bm U_m = \bm A_m \bm D^{1/2} \bm \Theta$ and $\bm U_{\ell} = \bm A_{\ell} \bm D^{1/2} \bm \Theta$,
where $\bm \Theta \in \mathbb{R}^{K \times K}$ is nonsingular.
Hence, one can estimate $\bm R_{m,n}$ via
\begin{align}
\bm R_{m,n}&= \bm U_m \bm U_{\ell}^{-1}\bm R_{n,\ell}^{\top}. \label{eq:estimXmn}
\end{align}
This simple procedure also allows us to characterize the estimation error of $\R_{m,n}$ when only a finite number of co-labeled items are available: 
\begin{theorem} \label{thm:svdimputation}
    Assume that $\widehat{\bm R}_{m,n}$ is estimated by \eqref{eq:constructC}-\eqref{eq:estimXmn} using the sample-estimated $\widehat{\R}_{m,r}$, $\widehat{\R}_{n,\ell}$ and $\widehat{\R}_{\ell,r}$ [using \eqref{eq:Rest} with at least $S$ items].
    Also assume that $\kappa(\bm A_m) \le \gamma$ and ${\rm rank}(\A_m)={\rm rank}(\bm D)=K$ for all $m\in[M]$. Let $\varrho = \underset{{(m,j)\in \bm \varOmega}}{\min} \sigma_{\min}(\bm R_{m,j})$. Suppose that $S = \Omega\left(\frac{K^2\gamma^2\log(1/\delta)}{\varrho^4} \right)$ for $\delta >0$. Then, for any $(m,n)\notin \bm \varOmega$, with probability of at least $1-\delta$, we have:
	\begin{align*}
	\|\widehat{\bm R}_{m,n} -\bm R_{m,n}\|_{\rm F} &= O \left(\frac{K^2\gamma^3\sqrt{\log(1/\delta)}}{\varrho^2\sqrt{S}}  \right),
	\end{align*} 
	where $\R_{m,n}=\A_m\D\A_n^\T$ is the missing ground-truth.
\end{theorem}
The proof can be found in the supplementary material in Sec. \ref{supp:svdimputaion}.
Note that the designated annotator approach can also estimate the diagonal blocks in $\X$, i.e., $\bm R_{m,m}=\A_m\D\A_m^\T$, by asking annotators $\ell,r$ to estimate $\R_{m,\ell}$, $\R_{m,r}$, and $\R_{\ell,r}$.
The diagonal blocks can never be observed, even if every pairwise annotator co-occurrence is observed, since $\R_{m,m}$ does not have physical meaning. Hence, being able to impute the diagonal blocks is particularly important for completing the matrix $\X$.

\begin{remark}
If $\bm R_{m,r}$, $\bm R_{n,\ell}$ and $\bm R_{\ell,r}$ are observed, then $\bm R_{m,n}$ can be imputed using \eqref{eq:constructC}-\eqref{eq:estimXmn} no matter if designated annotators exist.  As will be seen, this method works reasonably well even in the absence of designated annotators, especially when the number of missing co-occurrences is not large. Nonetheless, having designated annotators guarantees that every missing co-occurrence is estimated.
\end{remark}

\subsection{Robust Co-occurrence Imputation}
In some cases, designated annotators may not exist.
More critically, the estimated co-occurrences may not be equally reliable---since the estimation accuracy of $\widehat{\R}_{m,j}$ depends on the number of items that annotators $m$ and $j$ have co-labeled [cf. Eq.~\eqref{eq:Rest}], which may be quite unbalanced across different co-occurrences.
Under such circumstances, we propose a robust co-occurrence imputation criterion, i.e.,
\begin{subequations}\label{eq:optim_problem}
\begin{align} 
&\underset{\U_m, \U_j,~\forall (m,j)\in \bm \varOmega }{\rm minimize}~\sum_{(m,j) \in \bm \varOmega}\|\widehat{\bm R}_{m,j}-{\bm U}_m {\bm U}_j^{\top}\|_{\rm F}\\
&{\rm subject~to}~\|\U_m\|_{\rm F}\leq D,~\|\U_j\|_{\rm F}\leq D,~\forall m,
\end{align}     
\end{subequations}
where $D$ is an upper bound of $\|\U_m\|_F$---which is easy to acquire in our case, as $\U_m\in{\cal R}(\A_m\D^{1/2})$ and $\A_m$'s and $\D$ are bounded.
Our formulation can be understood as a block $\ell_2/\ell_1$-mixed norm based criterion, which is often used in robust estimation for ``downweighting'' outlying data; see e.g., \cite{xu2012robust,nie2014optimal,fu2016robust}. 

\paragraph{Stability Under Finite Sample.} Our formulation is reminiscent of matrix factorization based LRMC (see, e.g., \cite{sun2016guaranteed}), but with a special block missing pattern and a co-occurrence level robustification. The existing literature of LRMC and its recoverability analysis do not cover our case. Nonetheless, we show that the proposed criterion in \eqref{eq:optim_problem} is a sound criterion for co-occurrence imputation:
\begin{theorem} \label{thm:cf_stability}
Assume that the $\widehat{\R}_{m,j}$'s are estimated using \eqref{eq:Rest} with $S_{m,j}= |{\cal S}_{m,j}|$ for all $(m,j)\in\bm \varOmega$.	
{Also assume that each $\widehat{\bm R}_{m,j}$ is observed with the same probability.}
Let $ \{\bm U_m^*,\bm U_j^\ast\}_{ (m,j) \in \bm \varOmega  }$ be any optimal solution of \eqref{eq:optim_problem}. {Define $L = M(M-1)/2$}. Then we have
	\begin{align} \label{eq:Rbound_opt1}
	\frac{1}{L}\sum_{m< j}\|{\bm U}_m^*(\bm U_{j}^*)^\T-\bm R_{m,j}\|_{\rm F} \le  C\sqrt{\frac{MK^2\log(M) }{|\bm  \varOmega|}  }
	+ \left(\frac{1}{|\bm \varOmega|}+\frac{1}{L}\right)  \sum_{(m,j) \in \bm \varOmega}\frac{1+\sqrt{M}}{\sqrt{S_{m,j}}}, 
	\end{align}
with probability of at least $1- 3\exp(-M)$, where $C>0$.
\end{theorem}
The proof can be found in the supplementary material in Sec. \ref{supp:cf_stability}. Naturally, the criterion favors more annotators and more observed pairwise co-occurrences. A remark is that the second term on the right hand side of \eqref{eq:Rbound_opt1} is proportional to $\sum \| \bm N_{m,j} \|_{\rm F}$ where $\bm N_{m,j}=\widehat{\bm R}_{m,j} -\R_{m,j}$. Unlike $\sum \|\bm N_{m,j}\|_{\rm F}^2$, this term is not dominated by large $\| \bm N_{m,j} \|_{\rm F}$'s---which reflects the criterion's robustness to badly estimated $\widehat{\bm R}_{m,j}$'s.
Also note that the result in Theorem~\ref{thm:cf_stability} does not include the diagonal blocks $\bm R_{m,m}$'s. Nonetheless, the $\bm R_{m,m}$'s can be easily estimated using \eqref{eq:constructC}-\eqref{eq:estimXmn} if every other $\R_{m,j}$ is (approximately) recovered.


\paragraph{Iteratively Reweighted Algorithm.} We propose an iteratively reweighted alternating optimization algorithm to tackle \eqref{eq:optim_problem}. 
In each iteration, we handle a series of constrained least squares subproblem w.r.t. $\U_m$ with an updated weight ($w_{m,j}$) associated with $\widehat{\R}_{m,j}$ indicating its reliability; i.e.,
\begin{align}\label{eq:itwa}
    w_{m,j}&\leftarrow\left(\|\widehat{\R}_{m,j}-\widehat{\U}_m\widehat{\U}_j^\T\|_{\rm F}^2+\xi\right)^{-\frac{1}{2}},\\
    \widehat{\U}_m &\leftarrow \arg\min_{\|\bm U_m\|_{\rm F}\leq D} \sum_{j\in {\cal S}_{m,j}}~w_{m,j}\|\widehat{\R}_{m,j}-\U_m\widehat{\U}_j^\T\|_{\rm F}^2,  \nonumber
\end{align}
for all $(m,j)\in\bm \varOmega$, where $\xi>0$ is a small number to prevent numerical issues. The procedure in \eqref{eq:itwa} is repeatedly carried out until a certain convergence criterion is met.
This algorithm is reminiscent of the classic $\ell_2/\ell_1$ mixed norm minimization \cite{Chartrand2008}; see applications of mixed-norm based the matrix and tensor factorization in \cite{nie2014optimal,fu2016robust,fu2015joint}.
Note that the subproblems are fairly easy to handle, as they are quadratic programs; see the supplementary material in Sec. \ref{supp:rob_algorithm} for more details.

\subsection{Shifted ReLU Empowered SymNMF}
Assume that $\X=\H\H^\T$ is observed (after co-occurrence imputation) with no noise. 
The task of estimating $\A_m$ for all $m$ and $\D$ boils down to estimating $\H$ from $\X$, i.e., a SymNMF problem,
as the $\A_m$'s can be ``extracted'' from $\H$ easily (cf. Proposition~\ref{col:ident}).
The work in \cite{huang2014non} offered a simple algorithm for estimating $\H\geq \bm 0$. Taking the square root decomposition $\X=\U\U^\T$, one can see that $\U=\H\Q^\T$ with an orthogonal $\Q\in\mathbb{R}^{K\times K}$.
It was shown in \cite{huang2014non} that in the noiseless case, solving the following problem is equivalent to factoring $\X$ to $\X=\H\H^\T$ with $\H\geq\bm 0$: 
\begin{subequations}\label{eq:sub_optimW}
	\begin{align}
	\underset{\H,\Q}{\rm minimize}~~& \|\bm H - \bm U\bm Q\|^2_{\rm F}\\
	 {\rm subject ~to}~~&\bm H \ge \bm 0,~\bm Q^{\top}\bm Q = \bm I.\label{eq:Qconstraint}
	\end{align}
\end{subequations}
The work in \cite{huang2014non} proposed an alternating optimization algorithm for handling \eqref{eq:sub_optimW}. The algorithm is effective, but it is unclear if it converges to the ground-truth $\H$---even without noise.
To establish convergence assurances, we propose a simple tweak of the algorithm in \cite{huang2014non} as follows:
\begin{subequations} \label{eq:algo_updates}
	\begin{align}
	\H_{(t+1)}& \leftarrow {\sf ReLU}_{\alpha_{(t)}}\left( {\bm U} \bm Q_{(t)} \right) \label{eq:Ht}\\
	{\bm W}_{(t+1)}{\bm \Sigma}_{(t+1)}{\bm V}_{(t+1)}^\T &\leftarrow {\sf svd}\left(  \bm H_{(t+1)}^\T{\bm U} \right)\label{eq:WVt}\\
	\Q_{(t+1)}&\leftarrow  {\bm V}_{(t+1)}{\bm W}_{(t+1)}^\T\label{eq:Qt},
	\end{align}
\end{subequations}
where ${\sf ReLU}_{\alpha}(\cdot):\mathbb{R}^{MK\times K}\rightarrow\mathbb{R}^{MK\times K}$ is an elementwise shifted {\it rectified linear activation function} (ReLU) and is defined as
\begin{align*}
[{\sf ReLU}_{\alpha}(\Z)]_{i,k} = \begin{cases} \Z(i,k),~&\text{if } \bm Z(i,k) \ge \alpha,\\
0,~&\text{o.w.}, \end{cases}
\end{align*}
where $\alpha\geq 0$. 
The step in \eqref{eq:Ht} is orthogonal projection of each element of $\bm U\bm Q_{(t)}$ to $[\alpha_{(t)},+\infty)$.
The two steps \eqref{eq:WVt} and \eqref{eq:Qt} give the optimal solution to the $\bm Q$-subproblem, which is often referred to as the {\it Procrustes projection}. 
The key difference between our algorithm and the original version in \cite{huang2014non} is that we use a shifted ReLU function (with a pre-defined sequence $\{\alpha_{(t)}\}$) for the $\H$ update, while \cite{huang2014non} always uses $\alpha_{(t)}=0$. The modification is simple, yet it allows us to offer desirable convergence guarantees.
To proceed, we make the following assumption on $\H$:
\begin{assumption}\label{as:H}
The nonnegative factor $\H \in\mathbb{R}^{MK\times K}_+$ satisfies:
(i) ${\rank}(\H)=K$ and {$\|\H\|_{\rm F}  = \sigma$};  (ii) $\frac{ \|\H(j,:) \bm \Theta\|_2^2 }{ \| \H\bm \Theta \|_{\rm F}^2 } \leq \zeta,~\forall j,~\forall \bm \Theta\in\mathbb{R}^{K\times K} $; (iii) the locations of the nonzero elements {of $\H$ are uniformly distributed over $[MK]\times[K]$, and the set $\bm \varDelta = \{ (j,k) : [\bm H]_{j,k} > 0\} $}
has the following cardinality bound
\begin{equation}\label{eq:density}
 |\bm \varDelta|  = O\left(\frac{MK\gamma_{0}^2}{(1+MK\zeta ) \sigma^4}\right);   
\end{equation}
and (iv) $0<\gamma_0 \leq \min_{1\leq k\leq K}\{  \beta^2_k -\beta^2_{k+1} \}$, where $\beta_k$ is the $k$th singular value of $\H$ and $\beta_{K+1}=0$.
\end{assumption}
{
{Assumption (ii) means that the energy of the range space of $\bm H$ is well spread over its rows.}
Assumption (iii) means that the nonzero support of $\bm H$ is not too dense. This
reflects the fact that sparsity of the latent factors is often favorable in NMF problems, for both enhancing model identifiability and accelerating computation \cite{huang2014non,huang2014putting,fu2018nonnegative}.
Assumption (iv) means that $\H$'s singular values are sufficiently different, which is often useful in characterizing SVD-based operations when noise is present [cf. Eq.~\eqref{eq:WVt}].
}
With these assumptions, we show the following theorem:

\begin{theorem}\label{thm:converge}
Under Assumption \ref{as:H},
consider $\widehat{\U}=\H\Q^\T+\bm N$, where $\Q\in\mathbb{R}^{K\times K}$ is orthogonal, and apply \eqref{eq:algo_updates}. Denote $\nu=\|\bm N\|_{\rm F}$, $h_{(t)}=	\|\bm H_{(t)}-\bm H \bm \varPi\|_{\rm F}^2$ and $q_{(t)}=	\|\bm Q_{(t)}-\bm Q \bm \varPi\|_{\rm F}^2$, where $\bm \varPi$ is any permutation matrix. Suppose that $\nu \le \sigma\min\{(1-\rho)\sqrt{\eta} q_{(0)},1\}$ for {$\rho := O(\nicefrac{K\eta \sigma^4}{\gamma_0^2})\in(0,1)$}, where $\eta = (\nicefrac{|\bm \varDelta|}{MK^2})(1+MK\zeta )$, and that
\begin{equation}\label{eq:initcond}
    2\sigma q_{(0)} + 2\nu<\min_{(j,k) \in \bm \varDelta} [\bm H]_{j,k} .
\end{equation}  
Then, there exists $\alpha_{(t)}=\alpha > 0$ 
such that with probability of at least $1-\delta$, the following holds:
\begin{subequations}
\begin{align}
  q_{(t)} &\leq \rho q_{(t-1)} + O\left( \nicefrac{  K \sigma^2\nu^2}{\gamma_0^2}\right), \label{eq:Qerror}\\
  h_{(t)} &\leq 2\eta \sigma^2 q_{(t-1)} + 2\nu^2,  \label{eq:Herror}
      \end{align}
      \end{subequations}
 where $\delta = 2\exp\left(   - \nicefrac{ 2|\bm \varDelta|     }{K^2\left(1-\frac{|\bm \varDelta|-1}{MK^2}\right)  }\right) $. 

\end{theorem}
{The proof is relegated to the supplementary material in Sec. \ref{supp:convergethm}. 
Theorem~\ref{thm:converge} can be understood as that the solution sequence produced by the algorithm in \eqref{eq:algo_updates} converges {\it linearly} to neighborhoods of the ground-truth latent factors (up to a column permutation ambiguity)---and the neighborhoods have zero volumes if noise is absent. 
Specifically, Eq.~\eqref{eq:Qerror} means that, with high probability, the estimation error of $\Q$ decreases by a factor of $\rho$ after each iteration---which corresponds to a linear (geometric) rate. Consequently, by Eq.~\eqref{eq:Herror}, the estimation error of $\H$ also declines in the same rate. 

The theorem is also consistent with some long-existing empirical observations from the NMF literature. 
For example, the parameter $\eta$ is proportional to the number of nonzero elements in the latent factor $\H$. Apparently, a sparser $\H$ induces a smaller $\eta$, and thus a smaller $\rho$---which means faster convergence. The fact that NMF algorithms in general are in favor of sparser latent factors was previously observed and articulated from multiple perspectives \cite{huang2014putting,gillis2012sparse,huang2014non}.

A remark is that the convergence result in Theorem~\ref{thm:converge} holds if the initialization is reasonable [cf. Eq.~\eqref{eq:initcond}].
Nevertheless, our experiments show that simply using $\bm Q_{(0)}=\bm I$ works well in practice. We also find that using a diminishing sequence of $\{\alpha_{(t)}\}$ often helps accelerate convergence; see more discussions in the supplementary material in Sec. \ref{supp:symnmf_convergence}.


Convergence analysis for (Sym)NMF algorithms is in general challenging due to the NP-hardness, even without any noise \cite{vavasis2010complexity}. Provable NMF algorithms without relying on restrictive conditions like ``separability'' (see definition in \cite{donoho2003does}) are rarely seen in the literature. Notably, the work in \cite{li2016recovery,li2017provable} also used ${\sf ReLU}_{\alpha}(\cdot)$ for guaranteed NMF---but their algorithms are not for SymNMF and the analyses cannot be applied to our orthogonality-constrained problem.

}

\paragraph{Complexity.} The steps in \eqref{eq:Ht} and \eqref{eq:WVt} and the Procrustes projection in \eqref{eq:Qt} both cost $O(MK^3)$ flops. The SVD in \eqref{eq:WVt} requires $O(K^3)$ flops. Note that in crowdsourcing, $K$ is the number of classes, which is normally small relative to $M$ (the number of annotators). Hence, the algorithm often runs with a competitive speed.



\begin{table}[t]
	\centering
  \caption{Classification error (\%) and runtime (sec.) on the UCI Connect4 dataset ($N=20,561$, $M=10$, $K=3$). The ``\texttt{SymNMF}'' family are the proposed methods.}
		\resizebox{0.6\linewidth}{!}{
	\begin{tabular}{l|c|c|c|c}
		\hline
		\textbf{Algorithms}& \multicolumn{1}{l|}{$p_m=0.3$} & \multicolumn{1}{l|}{\makecell{$p_m \in (0.3,0.5),$ \\  $p_d=0.8$ }} & \multicolumn{1}{l|}{\makecell{$p_m \in (0.5,0.7),$ \\  $p_d=0.8$ }} & \multicolumn{1}{l}{\textbf{Time(s)}} \\
		\hline
		\hline
		\texttt{RobSymNMF} & \textbf{33.26} & 33.06 & 32.16 & 0.142 \\
		\hline
		\texttt{RobSymNMF-EM} & {34.27} & 33.20 & 32.11 & 0.191 \\
		\hline
		\texttt{RobSymNMF } ($w_{m,j}=1$) &\textbf{33.14}  & 34.60& 33.91 & 0.132  \\
		\hline
		\texttt{DesSymNMF} & {33.45} & \textbf{32.18} & \textbf{31.42} & 0.061 \\
		\hline
		\texttt{DesSymNMF-EM} & 33.94 & \textbf{32.50} & \textbf{31.40} & 0.128 \\
		\hline
		\texttt{SymNMF} (w/o imput.) & 34.87& 35.71 & 32.00 & 0.052 \\
		\hline
		\texttt{MultiSPA} & 47.78 & 42.24 & 49.54 & 0.020\\
		\hline
		\texttt{CNMF} & 36.26 & 39.55 & 34.70 & 4.741 \\
		\hline
		\texttt{TensorADMM} & 36.20 & 34.34 & 35.18 & 5.183 \\
		\hline
		\texttt{Spectral-D\&S} & 64.28 & 66.95 & 71.97 & 20.388 \\
		\hline
		\texttt{MV-EM} & 34.14 & 34.17 & 34.19 & 0.107 \\
		\hline
		\texttt{MinimaxEntropy} & 36.20 & 36.17 & 35.46 & 27.454 \\
		\hline
		\texttt{KOS}   & 54.55 & 43.21 & 39.41 & 12.798 \\
		\hline
		\texttt{Majority Voting} & 37.76 & 36.88 & 36.75 & - \\
		\hline
		\hline
	\end{tabular}%
}
	\label{tab:uci_connect4}%
\end{table}%

\begin{table}[t]
  \centering
  \caption{Classification error (\%) and runtime (sec.) on the UCI Credit dataset ($N=540$, $M=10$, $K=2$). The ``\texttt{SymNMF}'' family are the proposed methods.}
\resizebox{0.6\linewidth}{!}{
    \begin{tabular}{l|c|c|c|c}
    \hline
	\textbf{Algorithms}& \multicolumn{1}{l|}{$p_m=0.3$} & \multicolumn{1}{l|}{\makecell{$p_m \in (0.3,0.5),$ \\  $p_d=0.8$ }} & \multicolumn{1}{l|}{\makecell{$p_m \in (0.5,0.7),$ \\  $p_d=0.8$ }} & \multicolumn{1}{l}{\textbf{Time(s)}} \\
    \hline
    \hline
    	\texttt{RobSymNMF} & \textbf{16.31} & \textbf{13.99} & {13.74} & 0.152 \\
    \hline
    	\texttt{RobSymNMF-EM} & 16.76 & 13.96 & 14.06 & 0.160 \\
    \hline
	\texttt{RobSymNMF} ($w_{m,j}=1$) & \textbf{16.32} &\textbf{13.99} & \textbf{13.72} & 0.062  \\
	\hline
  	\texttt{DesSymNMF} & {16.37} & \textbf{13.83} & \textbf{13.67} & 0.052 \\
    \hline
 	\texttt{DesSymNMF-EM}  & 16.80 & {14.07} & 13.77 & 0.059 \\
    \hline
	\texttt{SymNMF} (w/o imput.) & 16.51& 13.94 & 13.85 & 0.039 \\
	\hline
	\texttt{MultiSPA} & 16.74 & 14.28 & 14.60 & 0.003 \\
    \hline
    	\texttt{CNMF} & 16.74 & 14.24 & 14.40 & 3.273 \\
    \hline
    	\texttt{TensorADMM} & 16.70 & 14.31 & {13.87} & 3.405 \\
    \hline
    	\texttt{Spectral-D\&S} & 16.98 & 14.24 & 14.00 & 1.790 \\
    \hline
    	\texttt{MV-EM} & 44.54 & 26.20 & 14.00 & 0.007 \\
    \hline
    	\texttt{MinimaxEntropy} & 17.50 & 17.00 & 16.78 & 0.728 \\
    \hline
    	\texttt{KOS}   & 17.28 & 14.22 & 14.89 & 0.009 \\
    \hline
    	\texttt{GhoshSVD} & 17.07 & 14.76 & 14.80 & 0.009 \\
    \hline
    	\texttt{EigenRatio} & 17.17 & 14.43 & 14.44 & 0.003 \\
    \hline
    	\texttt{Majority Voting} & 18.22 & 15.95 & 14.83 & - \\
    \hline
    \hline
    \end{tabular}%
	}
  \label{tab:uci_credit}%
\end{table}%


\begin{table}[t]
  \centering
  \caption{Classification error (\%) and runtime (sec.) on the UCI Car dataset ($N=1,352$, $M=10$, $K=4$). The ``\texttt{SymNMF}'' family are the proposed methods. }
\resizebox{0.6\linewidth}{!}{
  \begin{tabular}{l|c|c|c|c}
    \hline
    \textbf{Algorithms} & \multicolumn{1}{c|}{${\sf Miss}=70\%$} & \multicolumn{1}{c|}{${\sf Miss}=50\%$} & \multicolumn{1}{c|}{${\sf Miss}=30\%$} & \multicolumn{1}{c}{\textbf{Time (s)}} \\
    \hline
    \hline
   	\texttt{RobSymNMF} & \textbf{24.01} & \textbf{23.17} & \textbf{22.05} & 0.108 \\
    \hline
   	\texttt{RobSymNMF-EM} & 24.93 & 23.71 & \textbf{22.03} & 0.123 \\
    \hline
		\texttt{RobSymNMF} ($w_{m,j}=1$) &\textbf{24.01} &\textbf{23.40} & 22.16 &0.100  \\
	\hline
   	\texttt{DesSymNMF} & {24.50} & {23.41} & 23.00 & 0.048 \\
    \hline
    	\texttt{DesSymNMF-EM} & 24.91 & 24.59 & 23.45 & 0.060 \\
    \hline
	\texttt{SymNMF} (w/o imput.) &24.43 & 24.03 & 24.40 & 0.031 \\
	\hline
    \texttt{MultiSPA} & 47.12 & 47.14 & 33.84 & 0.002 \\
    \hline
      \texttt{CNMF} & 43.65 & 41.49 & 30.55 & 3.666 \\
    \hline
      \texttt{TensorADMM} & 36.67 & 39.32 & 37.38 & 4.900 \\
    \hline
      \texttt{Spectral-D\&S} & 31.20 & 29.67 & 29.14 & 47.800 \\
    \hline
      \texttt{MV-EM} & 30.27 & 29.96 & 29.65 & 0.013 \\
    \hline
      \texttt{MinimaxEntropy} & 28.22 & 25.73 & 24.68 & 12.664 \\
    \hline
      \texttt{KOS }  & 48.87 & 49.87 & 41.83 & 0.104 \\
    \hline
      \texttt{Majority Voting} & 43.88 & 43.08 & 42.40 & - \\
    \hline
    \hline
    \end{tabular}%
    }
  \label{tab:uci_car}%
\end{table}%

\begin{table}[t]
  \centering
  \caption{Classification error (\%) and runtime (sec.) on the AMT datasets ``Bluebird'' and ``Dog''.The ``\texttt{SymNMF}'' family are the proposed methods. }
  			\resizebox{0.6\linewidth}{!}{
    \begin{tabular}{l|c|c|c|c}
    \hline
    \textbf{Algorithms} & \multicolumn{2}{c|}{\makecell{\textbf{Bluebird} \\  ($N=108$, $M=39$, $K=2$) }} & \multicolumn{2}{c}{\makecell{\textbf{Dog} \\  ($N=807$, $M=52$, $K=4$) }} \\
    \hline
    \hline
          & \multicolumn{1}{c|}{\textbf{Error} (\%)} & \multicolumn{1}{c|}{\textbf{Time} (s)} & \multicolumn{1}{c|}{\textbf{Error} (\%)} & \multicolumn{1}{c|}{\textbf{Time} (s)} \\
    \hline    
	\hline
   \texttt{RobSymNMF} & 11.11 & 0.72  & \textbf{16.10}  & 0.41 \\
    \hline
    \texttt{RobSymNMF-EM} & 11.11 & 0.79  & \textbf{15.86} & 0.48 \\
    \hline
   \texttt{RobSymNMF} ($w_{m,j}=1$) & 11.11 & 0.38  & \textbf{16.10}  & 0.38 \\
    \hline
	\texttt{DesSymNMF} & \textbf{10.18} & 0.15  & 16.35 & 0.11 \\
    \hline
    \texttt{DesSymNMF-EM} & \textbf{10.18} & 0.19  & \textbf{15.86} & 0.16 \\
    \hline
	\texttt{SymNMF} (w/o imput.) & \textbf{10.18} & 0.12  & 16.72 & 0.10 \\
    \hline
	\texttt{MultiSPA} & 13.88 & 0.10  & 17.96 & 0.09 \\
    \hline
	\texttt{CNMF} & 11.11 & 6.76  & \textbf{15.86} & 17.14 \\
    \hline
    \texttt{TensorADMM} & 12.03 & 85.56 & 18.01 & 613.93 \\
    \hline
	\texttt{Spectral-D\&S} & 12.03 & 1.97  & 17.84 & 43.88 \\
    \hline
	\texttt{MV-EM} & 12.03 & 0.02  &\textbf{ 15.86} & 0.06 \\
    \hline
	\texttt{MinimaxEntropy} & \textbf{8.33}  & 3.43  & 16.23 & 4.6 \\
    \hline
	\texttt{KOS}   & 11.11 & 0.11  & 31.84 & 0.17 \\
    \hline
    \texttt{GhoshSVD} & 27.77 & 0.02  & N/A   & N/A \\
    \hline
    \texttt{EigenRatio} & 27.77 & 0.03  & N/A   & N/A \\
    \hline
    \texttt{PG-TAC} & 24.07 & 0.04  & 18.21  & 21.11 \\
    \hline
    \texttt{CRIA}$_V$ & 24.07 & 0.05  & 17.10   & 18.48 \\
    \hline
	\texttt{Majority Voting} & 21.29 & N/A   & 17.91 & N/A \\
    \hline
    \hline
    \end{tabular}%
    }
  \label{tab:AMT1}%
\end{table}%

\begin{table}[t]
  \centering
  \caption{Classification error (\%) and runtime (sec.) on the AMT datasets ``RTE'' and ``TREC''. The ``\texttt{SymNMF}'' family are the proposed methods.}
    			\resizebox{0.6\linewidth}{!}{
    \begin{tabular}{l|c|c|c|c}
    \hline
    \textbf{Algorithms} & \multicolumn{2}{c|}{\makecell{\textbf{RTE} \\  ($N=800$, $M=164$, $K=2$) }} & \multicolumn{2}{c}{\makecell{\textbf{TREC} \\  ($N=19,033$, $M=762$, $K=2$) }}\\
    \hline
    \hline
          & \textbf{Error} (\%)  & \textbf{Time} (s)& \textbf{Error} (\%) & \textbf{Time} (s) \\
    \hline
    \texttt{RobSymNMF} & \textbf{7.25}  & 2.31  & 30.68 & 64.99 \\
    \hline
   \texttt{RobSymNMF-EM} & \textbf{7.12}  & 2.4   & 29.62 & 67.39 \\
    \hline
   \texttt{RobSymNMF} ($w_{m,j}=1$) & 7.37 & 1.35  & {33.23}  & 62.33 \\
    \hline
    \texttt{DesSymNMF} & 13.87 & 3.32  & 36.75 & 71.31 \\
    \hline
    \texttt{DesSymNMF-EM} & \textbf{7.25} & 3.43  & \textbf{29.36} & 72.13 \\
    \hline
	\texttt{SymNMF} (w/o imput.) & {48.75} & 0.23  & 35.47 & 57.60 \\
    \hline
    \texttt{MultiSPA} & 8.37  & 0.18  & 31.56 & 51.34 \\
    \hline
    \texttt{CNMF} & \textbf{7.12 } & 18.12 & 29.84 & 536.86 \\
    \hline
    \texttt{TensorADMM} & N/A   & N/A   & N/A   & N/A \\
    \hline
    \texttt{Spectral-D\&S} & \textbf{7.12}  & 6.34  & \textbf{29.58} & 919.98 \\
    \hline
    \texttt{MV-EM} & \textbf{7.25}  & 0.09  & 30.02 & 3.12 \\
    \hline
    \texttt{MinimaxEntropy} & 7.5   & 6.4   & 30.89   & 356.32 \\
    \hline
    \texttt{KOS}   & 39.75 & 0.07  & 51.95 & 8.53 \\
    \hline
    \texttt{GhoshSVD} & 49.12 & 0.06  & 43.03 & 7.18 \\
    \hline
    \texttt{EigenRatio} & 9.01  & 0.07  & 43.95 & 1.87 \\
    \hline
    \texttt{PG-TAC} & 8.12 & 50.41  & 33.89  & 917.21 \\
    \hline
    \texttt{CRIA}$_V$ & 9.37 & 49.04  & 34.59   & 900.34 \\
    \hline
    \texttt{Majority Voting} & 10.31 & N/A   & 34.85 & N/A \\
    \hline
    \hline
    \end{tabular}%
    }
  \label{tab:AMT2}%
\end{table}%

\section{Experiments}
\paragraph{Baselines.} We denote the proposed robust co-occurrence imputation-assisted SymNMF algorithm as \texttt{RobSymNMF} and the designated annotators-based imputation-based SymNMF as \texttt{DesSymNMF}.   To benchmark our methods, we employ a number of crowdsourcing algorithms, namely, \texttt{MultiSPA}, \texttt{CNMF} \cite{ibrahim2019crowdsourcing}, \texttt{TensorADMM} \cite{traganitis2018blind} \texttt{Spectral-D\&S} \cite{zhang2014spectral}, \texttt{KOS} \cite{karger2013efficient}, \texttt{EigenRatio} \cite{dalvi2013aggregating}, \texttt{GhoshSVD} \cite{ghosh2011moderates}, and \texttt{MinimaxEntropy} \cite{zhou2014aggregated}. 
We also employ \texttt{EM} \cite{dawid1979maximum} initialized by \texttt{majority voting} (denoted as \texttt{MV-EM}) as a baseline. 
Note that \texttt{CNMF} is the state-of-the-art, which uses pairwise co-occurrences as our methods do.
We also use our proposed methods to initialize \texttt{EM} (\texttt{RobSymNMF-EM} and \texttt{DesSymNMF-EM}).
For all the D\&S model-based algorithms, we construct an MAP predictor for $y_n$ after the model is learned.

\paragraph{Synthetic Data Experiments. } The synthetic data experiments are presented in the supplementary material in Sec. \ref{supp:synthetic}.

\paragraph {UCI Data Experiments. }
We consider a number of UCI datasets, namely, ``Connect4'', ``Credit'' and ``Car''.  
We choose different classifiers from the
MATLAB machine learning toolbox, e.g., support vector machines and decision tree; {see Sec. \ref{supp:uci} of the supplementary material for details.} These classifiers serve as annotators in our experiments. We partition the datasets randomly in every trial, with a training to testing ratio being 1/4---which means that the annotators are not extensively trained. Each classifier (annotator) is then allowed to label a test item with probability $p_m \in (0,1]$.

Tables \ref{tab:uci_connect4} and \ref{tab:uci_credit} show the performance of the algorithms on Connect4 and Credit, respectively.
In the first column of the tables, $p_m$ is fixed for all $M$ annotators. In the second and third columns, we designate two annotators $\ell$ and $r$, and let them label the data items with higher probabilities (i.e., $p_d$). This way, the designated annotators can co-label items with many other annotators---which can help impute missing co-occurrences using \eqref{eq:constructC}-\eqref{eq:estimXmn}. The designated annotators $\ell$ and $r$ are chosen from the $M$ annotators randomly in each trial. The probability $p_m$ is also randomly chosen from a pre-specified range as indicated in the tables. We use this setting to simulate realistic scenarios in crowdsourcing where incomplete, noisy, and unbalanced labels are present. The results are averaged from 20 trials. 

From Tables \ref{tab:uci_connect4} and \ref{tab:uci_credit}, one can observe that the proposed methods show promising classification performance in all cases.
The proposed methods exhibit clear improvements upon the \texttt{CNMF}---especially in the more challenging case in Table~\ref{tab:uci_connect4}. 
The proposed methods also outperform the the third-order statistics-based ones (\texttt{TensorADMM} and \texttt{Spectral-D\&S}) under most settings, articulating the advantages of using second-order statistics.
In terms of the runtime performance, the proposed \texttt{SymNMF} family are also about 20 to 50 times faster compared to \texttt{CNMF} in these two tables. 
There are $10\%$ of co-occurrences missing in the cases corresponding to the first columns of Tables \ref{tab:uci_connect4} and \ref{tab:uci_credit}. \texttt{DesSymNMF} using \eqref{eq:constructC}-\eqref{eq:estimXmn} is able to impute all the missing ones, although we did not assign any designated annotator.
In both tables, \texttt{RobSymNMF} slightly (but consistently) outperforms \texttt{DesSymNMF} when there is no designated annotators, showing some advantages in such cases.
In the above experiments, our robust imputation algorithm in \eqref{eq:itwa} offers labeling errors that are smaller than or equal to its non-robust version (with $w_{m,j}=1$) in 5 out of 6 settings.

Table \ref{tab:uci_car} presents the performance of the algorithms on the Car dataset under different proportions of missing co-occurrences; {see Sec. \ref{supp:uci} of the supplementary material for the details of generating such cases. In this experiment, we do not assign designated annotators. If $\bm R_{m,n}$ cannot be completed by observed co-occurrences using \eqref{eq:constructC}-\eqref{eq:estimXmn}, we leave it as an all-zero block.} Using \eqref{eq:constructC} and \eqref{eq:estimXmn}, \texttt{DesSymNMF} still improves the missing proportions to 17\%, 9\% and 0\% for the columns from left to right, respectively.
One can see that the proposed method largely outperforms the baselines, especially in the cases where 70\% of the $\bm R_{m,j}$'s are not observed. However, \texttt{CNMF} is not able to produce competitive results in this experiment.


%


\paragraph{AMT Data Experiments. }
We also evaluate the algorithms using various AMT datasets, namely ``Bluebird", ``Dog", ``RTE" and ``TREC", which are annotated by human annotators. The AMT datasets are more challenging, in the sense that we have no control for annotation acquisition and no designated annotators are available. Similar as before, for \texttt{DesSymNMF}, we leave the co-occurrences that cannot be recovered by \eqref{eq:constructC}-\eqref{eq:estimXmn} as all-zero blocks.
{In the AMT experiments, we include two additional baselines based on tensor completion, namely, \texttt{PG-TAC} \cite{zhou2016crowdtensor} and \texttt{CRIA}$_V$ \cite{li2018multi}---both of which reported good performance over AMT datasets.
}



Table \ref{tab:AMT1} and \ref{tab:AMT2} present the evaluation results over the AMT datasets. The \texttt{TensorADMM} algorithm could not run with large $M$ due to scalablity issues.  
The results are consistent with those observed in the UCI experiments. The proposed methods' labeling accuracy is either comparable with or better than that of \texttt{CNMF},  but is order-of-magnitude faster.
The proposed methods are also observed to most effectively initialize the \texttt{EM} algorithm \cite{dawid1979maximum}. 
{An observation is that there are 2.5\%, 14.0\%, 90.68\%, and 96.57\% of the pairwise co-occurrences missing in Bluebird, Dog, RTE and TREC, respectively. \texttt{DesSymNMF} is able to bring down the missing proportions to 0.00\%. 11.34\%, 50.15\%, and 92.18\%, respectively. The \texttt{DesSymNMF} imputation can sometimes improve the final accuracy significantly; see the Dog and RTE columns. In addition, our robust imputation criterion \eqref{eq:optim_problem} and the algorithm in \eqref{eq:itwa} often exhibit visible improvements upon the equally weighted (non-robust) version, as in the UCI case.}

\paragraph{Comparison with Deep Learning-based Methods. }
{We present an additional experiment and compare the proposed approaches with two deep learning (DL)-based crowdsourcing methods in \cite{Rodrigues2018deep}. The details can be found in the supplementary material in Sec. \ref{supp:realdata}.}




\section{Conclusion}
We proposed a D\&S model identification-based crowdsourcing method that uses sample-efficient pairwise co-occurrences of annotator responses. 
We advocated a SymNMF-based framework that offers strong identifiability of the D\&S model under reasonable conditions. To realize the SymNMF framework, we {proposed} two lightweight algorithms for provably imputing missing co-occurrences when the annotations are incomplete. We also proposed a computationally economical SymNMF algorithm, and analyzed its convergence properties. We tested the framework on UCI and AMT data and observed promising performance. The proposed algorithms are typically order-of-magnitude faster than other high-performance baselines. 

\section{Acknowledgement}
This work is supported in part by the National Science Foundation under Project NSF IIS-2007836 and the Army Research Office under Project ARO W911NF-19-1-0247.


\bibliographystyle{IEEEtran}

\appendix

\section{Notation} \label{sup:notations}

\begin{table}[H]
	\centering
	\label{tab:phi}
	\begin{tabular}{c c  }
	\textbf{Notation} & \textbf{Definition} \\
		\hline
		\hline
		$x$  & scalar in $\mathbb{R}$ \\
		\hline
        $\bm x$ & vector in $\mathbb{R}^n$, i.e., $\x=[x_1,\ldots,x_n]^\T$ \\
        \hline
        $\bm X$  & matrix in $\mathbb{R}^{m\times n}$ with $\X(i,j)=x_{i,j}$ \\
        \hline
        $[\bm X]_{i,j}$ or $\bm X(i,j)$   & $(i,j)$th entry of $\bm X$ \\
        \hline
        $\bm X \ge \bm 0$  & $\X(i,j)\geq 0$ $\forall~(i,j)$\\
        \hline
        $\kappa(\bm X)$  & condition number of $\bm X$  \\
        \hline
        $\sigma_{\max}(\bm X)$  & maximum singular value of $\bm X$ \\
        \hline
        $\sigma_{\min}(\bm X)$  & minimum singular value of $\bm X$  \\
        \hline
        $\|\bm X\|_2$  & 2-norm of $\bm X$ (same as $\sigma_{\max}(\X)$)  \\
        \hline 
        $\|\bm X\|_{\rm F}$  & Frobenius norm of $\bm X$  \\
        \hline
        ${\mathcal R}(\bm X)$ &  range space of $\bm X$ \\
         \hline
         ${\sf cone}(\bm X)$  & conic hull of $\X$: $\{ \bm y~|~\bm y= \bm X\bm \theta,~\forall \bm \theta\geq \bm 0\}$  \\
         \hline
         $\|\bm x\|_2$  & $\ell_2$-norm of $\bm x$ \\
         \hline
          $\|\bm x\|_1$  & $\ell_1$-norm of $\bm x$ \\
          \hline
          ${\sf Diag}(\bm x)$  & diagonal matrix with $x_1,\ldots,x_n$ in the diagonal\\
         \hline
          ${\dagger}$  & pseudo-inverse   \\
        \hline
        $\top$  & transpose\\
        \hline
         $|\mathcal{C}|$  & the cardinality of the set $\mathcal{C}$\\
         \hline
         $[T]$  & $\{1,\dots,T\}$ for an integer $T$\\
         \hline
         $\bm I$ & identity matrix with proper size\\
         \hline
         $\bm 1$ & all-one vector with proper size\\
         \hline
         $\bm 0$ & all-zero vector or matrix with proper size\\
         \hline
         $\bm e_i$ & unit vector with the $i$th element being 1\\
         \hline
          $\mathbb{R}_+^n$ & nonnegative orthant of $\mathbb{R}^n$ \\
         \hline
        \hline
	\end{tabular}
\end{table}

\section{More Details of The Robust Co-occurrence Imputation Algorithm} \label{supp:rob_algorithm}
\subsection{Iteratively Reweighted Algorithm for Robust Co-occurrence Imputation}
In order to design an algorithm for solving Problem \eqref{eq:optim_problem}, we approximate \eqref{eq:optim_problem} using a smooth version of the objective function. Specifically, we propose to use
\begin{subequations}\label{eq:optim_problem_smooth}
\begin{align} 
&\underset{\U_m, \U_j,~\forall (m,j)\in \bm \varOmega }{\rm minimize}~\sum_{(m,j) \in \bm \varOmega}\left(\|\widehat{\bm R}_{m,j}-{\bm U}_m {\bm U}_j^{\top}\|^2_{\rm F}+\xi\right)^{\frac{1}{2}}\\
&{\rm subject~to}~\|\U_m\|_{\rm F}\leq D,~\|\U_j\|_{\rm F}\leq D,~\forall m,
\end{align}     
\end{subequations}
where $\xi >0$ is a small number.



We update $\bm U_m$ by fixing $\{w_{m,j}\}_{(m,j)\in \bm \varOmega}$ and $\bm U_j$'s where $j \neq m$. Then, we can update $\{w_{m,j}\}_{(m,j)\in \bm \varOmega}$, by fixing $\bm U_m$ and $\bm U_j$, for all $(m,j) \in \bm \varOmega$. 
In each iteration $t$, the sub-problem to solve $\bm U_m$ can be written as
\begin{subequations}\label{eq:optim_problem_smooth_mod_sub}
\begin{align} 
&\underset{\U_m }{\rm minimize}~ \sum_{j\in{\cal S}_{m}}w^{(t)}_{m,j}\|\widehat{\bm R}_{m,j}-{\bm U}_m ({\bm U}^{(t)}_j)^{\top}\|^2_{\rm F} \label{eq:optim_problem_smooth_object}\\
&{\rm subject~to}~\|\U_m\|_{\rm F}\leq D,
\end{align}     
\end{subequations}
where ${\cal S}_{m}=\{j~|~\text{ $\widehat{\bm R}_{m,j}$ or $\widehat{\bm R}_{j,m}$ is observed}\}$.
The problem in \eqref{eq:optim_problem_smooth_mod_sub} is a second-order cone-constrained quadratic program, and can be solved using any off-the-shelf convex optimization algorithm.
We propose to use the
\textit{projected gradient descent} (PGD) algorithm due to its simplicity. Specifically, in iteration $r$ of the PGD conducted during the $t$th outer iteration, $\bm U_m$ is updated via
\begin{align*}
    \bm U_m^{(t,r+1)} \leftarrow {\sf Proj}_{\cal D}\left(\bm U_m^{(t,r)} - \beta \bm G^{(t,r)}_m  \right),
\end{align*}
where $\beta>0$ is the step size, ${\sf Proj}_{\cal D}\left(\cdot\right):\mathbb{R}^{K\times K} \rightarrow \mathbb{R}^{K\times K}$ denotes the orthogonal projection onto the set
$  {\cal D} =\{ \X\in\mathbb{R}^{K\times K}~|~\|\X\|_{\rm F}\leq D \}, $
and $\bm G^{(t,r)}_m$ is the gradient of the objective function \eqref{eq:optim_problem_smooth_object} w.r.t to $\bm U_m$. 
Specifically, we have
\[  {\bm G^{(t,r)}_m =  \sum_{j\in{\cal S}_{m} } w^{(t)}_{m,j} \left(     \bm U_m^{(t,r)} (\bm U_j^{(t)})^\T \bm U_j^{(t)}  -\widehat{\bm R}_{m,j}\bm U_j^{(t)}    \right)  } .  \]
The step size is selected as the inverse of the Lipschitz constant of the gradient.
In addition, the projection is simply re-scaling; i.e., for any $\bm Z \in \mathbb{R}^{K \times K}$,
\[  {\sf Proj}_{\cal D}(\Z) =\begin{cases}  \Z,\quad & \Z \in {\cal D}\\  \frac{\Z}{\|\Z\|_F},\quad & \Z \notin {\cal D}.            \end{cases} \]

Note that we let
\[  \U_m^{(t+1)} \leftarrow \U_m^{(t,r^\star_t)}, \quad   \U_m^{(t+1,0)} \leftarrow    \U_m^{(t+1)}, \]
where $r_t^\star$ is the number of iterations where the PGD stops for updating $\bm U_m^{(t)}$.
After $\bm U_m$ for all $m$ are updated using PGD, we update $w_{m,j}$, for all $(m,j) \in \bm \varOmega$, by the following:
\begin{align*}
    w_{m,j}^{(t+1)}&\leftarrow\left(\|\widehat{\R}_{m,j}-{\U}^{(t)}_m({\U}^{(t)}_j)^\T\|_{\rm F}^2+\xi\right)^{-\frac{1}{2}},~\forall (m,j)\in\bm \varOmega.
\end{align*}

\subsection{Complexity and Convergence}

The per-iteration complexity of the algorithm is often not large, due to its first-order optimization nature. The complexity-dominating step are the computation of the step size and constructing the gradient, which both cost $O(MK^3)$ flops. This is acceptable since $K$ is normally small. 

Iteratively reweighted algorithms' stationary-point convergence properties have been well understood. By a connection between the algorithm and the {\it block successive upper bound minimization} (BSUM) \cite{razaviyayn2013unified}, it is readily seen that the solution sequence converges to a stationary point of \eqref{eq:optim_problem_smooth}. Although global optimality of the algorithm may be much harder to establish, such a procedure often works well in practice---which presents a valuable heuristic for tackling the stability-guaranteed co-occurrence imputation criterion in Theorem~\ref{thm:cf_stability}, i.e., Problem~\eqref{eq:optim_problem}.

\section{More Details of Experiments}

\paragraph {Parameters.} The stopping criterion for all the iterative algorithms in the experiments is set such that the algorithms are terminated when the relative change of their respective cost functions is less than $10^{-6}$. { For the proposed SymNMF algorithm, we set $\alpha_{(0)}= 10^{-6}$, and we use two $\alpha_{(t)}$ scheduling rules in our simulations and real data  experiments, respectively. Specifically, for simulations that demonstrate the convergence properties of the proposed algorithm, we use $\alpha_{(t)} = \psi^{t+1}$ where $0 <\psi<1 $.
For the rest of the simulations and real data experiments, we let $\alpha_{(t)}=\alpha_{(0)}$ for simplicity.} 
We run all the experiments in Matlab 2018b on Windows 10 on an Intel I7 CPU running at 3.40 GHZ. 

\subsection{Synthetic Data Simulations} \label{supp:synthetic}

\subsubsection{Identifiability}
In this section, we analyze the D\&S model identifiability of the proposed framework using synthetic data experiments. 

First, we consider the noiseless case where we directly generate $\bm R_{m,j}=\bm A_m\bm D\bm A_j^\T$ for $(m,j)\in\bm \varOmega$  and observe if the confusion matrices and the prior can be identified by the algorithms up to a common column permutation. We fix $M=25$ annotators and the number of classes $K=3$. An annotator is chosen randomly from $M$ annotators and is made as a ``class specialist" of all the classes $1,\dots,K$. This is achieved by setting its confusion matrix $\bm A_m$ to be close to an identity matrix. Specifically, for the chosen ``class specialist", we set $ \|\bm A_m(k,:)-\bm e_k^{\top}\|_2 \le \varepsilon. \forall k,$
with $\varepsilon = 0.10$. In this way, the $\bm H$ matrix as defined in \eqref{eq:Xbig} approximately satisfy the SSC (see Definition \ref{as:ss}).
The columns of the confusion matrices for the rest of the annotators  and the prior probability vector $\bm \lambda \in \mathbb{R}^K$ are generated using Dirichlet distribution with parameter $\bm \mu = \bm 1 \in \mathbb{R}^K$.  
We generate different missing proportions by observing each pairwise blocks with a probability smaller than one. Using these observed pairwise blocks, the proposed algorithms are run and the mean squared error (MSE) of the confusion matrices and the prior vector are estimated. The MSE is computed as follows:
    \begin{align}
    \text{MSE} = \underset{\bm \varPi}{\text{min}} \frac{1}{MK+1}\left( \|\bm \varPi^{\top} \bm \lambda - \widehat{\bm \lambda}\|^2_2+\sum_{m=1}^M  \lVert {\A}_m\bm \varPi-\widehat{{\A}}_m\rVert_{\rm F}^2\right),
\end{align}
where $\bm \varPi$ is a permutation matrix and $\widehat{\bm A}_m,m = [M]$ and $\widehat{\bm \lambda}$ are the outputs by the algorithms.

\begin{table}[t]
  \centering
  \small
  \caption{Average MSE of the proposed methods for $M=25$, $K=3$ with different block missing proportions (noiseless case).}
  \resizebox{0.5\textwidth}{!}{
    \begin{tabular}{l|r|r|r}
    \hline
    \textbf{Algorithms} & \multicolumn{1}{l|}{{\sf Miss}=70\%} & \multicolumn{1}{l|}{{\sf Miss}=50\%} & \multicolumn{1}{l}{{\sf Miss}=30\%} \\
    \hline
    \hline
    \texttt{RobSymNMF} & $4.10 \times 10^{-3}$ & $1.70\times 10^{-3}$ & $3.44\times 10^{-4}$ \\
    \hline
    \texttt{DesSymNMF} & $2.84\times 10^{-4}$ & $4.59\times 10^{-4}$ & $3.05\times 10^{-4}$ \\
    \hline
    \hline
    \end{tabular}%
    }
  \label{tab:synthetic1}%
\end{table}%

Table \ref{tab:synthetic1} presents the MSE of the proposed methods for different proportions of the missing co-occurrences, averaged over 20 different trials. Both the proposed methods output low MSE values in all the cases. One can see that the MSE of the \texttt{RobSymNMF} decreases when more blocks are observed, which is consistent with Theorem~\ref{thm:cf_stability}. Since we consider the noiseless case by observing ${\bm R_{m,j}} = \bm A_m \bm D \bm A_j^{\top}$ for all $(m,j) \in \bm \varOmega$, the algorithm \texttt{DesSymNMF} is able to impute all the missing pairwise co-occurrences accurately via \eqref{eq:constructC}-\eqref{eq:estimXmn}. Therefore, the MSE of the \texttt{DesSymNMF} is more or less unaffected with changing co-occurrence missing proportions. 

\begin{table}[t]
  \centering
  \caption{Average MSE and the runtime of the proposed methods and baselines for $M=25, K=3, p =0.3$ for different values of $N$}
    \resizebox{0.5\textwidth}{!}{
    \begin{tabular}{l|c|c|c|c}
    \hline
    \textbf{Algorithms} & \multicolumn{1}{c|}{$N=1000$} & \multicolumn{1}{c|}{$N=5000$} & \multicolumn{1}{c|}{$N=10000$} & \multicolumn{1}{c}{Time (s)}\\
    \hline
    \hline
    \texttt{RobSymNMF} & \textbf{0.0099} & \textbf{0.0019} & \textbf{0.0012} &0.342\\
    \hline
    \texttt{DesSymNMF} & \textbf{0.0127} & 0.0038 & 0.0029&0.072 \\
    \hline
    \texttt{MultiSPA} & 0.2248 & 0.1645 & 0.1575&0.0148 \\
    \hline
    \texttt{CNMF} & 0.0314 & \textbf{0.0036} & \textbf{0.0009}&22.475 \\
    \hline
    \texttt{TensorADMM} & 0.0218 & 0.0041 & 0.0011&27.263 \\
    \hline
    \texttt{Spectral-D\&S} & 0.0465 & 0.0259 & 0.0050&17.492 \\
    \hline
    \texttt{MV-EM} & 0.0495 & 0.0866 & 0.1051&0.055 \\
    \hline
    \hline
    \end{tabular}%
    }
  \label{tab:synthetic2}%
\end{table}%

Table \ref{tab:synthetic2} presents the average MSE and the runtime of the methods under test using various numbers of data items.
We fix $M=25$, $K=3$ and vary the number of data items $N$. The generating process for the confusion matrices and the prior vector is the same as that used in Table \ref{tab:synthetic1}. Once the confusion matrices $\bm A_m, m=[M]$ are generated, the labels from each annotator $m$ for a data item with true label $c \in [K]$ is randomly chosen from $[K]$ using the probability distribution $\bm A_m(:,c)$.  An annotator label for each data item is retained with probability $p<1$ which is fixed at 0.3. Using such labels, the co-occurrences are estimated via \eqref{eq:Rest}. In all the cases in Table \ref{tab:synthetic2}, there are 4\% of the pairwise co-occurrences missing. One can see that the proposed methods, especially \texttt{RobSymNMF}, outperform the other methods in most of the cases and also enjoy promising runtime performance. The \texttt{DesSymNMF} imputes all the missing blocks, even though there are no designated annotators and still provides good performance. 
This is because most co-occurrences are available and the conditions for using \eqref{eq:constructC}-\eqref{eq:estimXmn} are almost always satisfied.
Particularly, the MSEs of the proposed methods are at least 40\% lower than the best-performing baseline, when the number of data items are small (see $N=1000$). 
This shows the advantages of the pairwise co-occurrence based methods in the sample-starved regime.
As $N$ increases, the MSEs of all the methods become better and closer. 

\subsubsection{Convergence} \label{supp:symnmf_convergence}
In this section, we compare the convergence behaviors of the proposed SymNMF algorithm [cf. Eq.~\eqref{eq:algo_updates}] and the SymNMF algorithm proposed in \cite{huang2014non}. The proposed algorithm uses a shifted ReLU function for the $\H$ update with $\alpha_{(t)}>0$. The algorithm in \cite{huang2014non} has nonnegative thresholding, i.e., a ReLU function with $\alpha_{(t)}=0$ for all $t$. 

In our proof of Theorem~\ref{thm:converge}, we assume that $\alpha_{(t)}$ is chosen such that a key condition is always satisfied; see Eqs.~\eqref{eq:cond_1} and \eqref{eq:newcond}.
In practice, these conditions may not be checkable. A heuristic way of selecting $\{ \alpha_{(t)} \}$ is to use a diminishing sequence $\{ \alpha_{(t)} \}$. 
In simulations, we found that using such sequences $\{ \alpha_{(t)} \}$ can often accelerate convergence.

We consider a nonnegative matrix $\bm H \in \mathbb{R}_+^{J \times K}$ and control its sparsity (i.e., the number of zero entries in $\bm H$) using a parameter $\varphi$ such that $1-\varphi = {\sf Pr}([\bm H]_{j,k} =0)$. The nonzero entries are randomly sampled from a uniform distribution between 0 and 1. Using the matrix $\bm H$, the symmetric nonnegative matrix $\bm X \in \mathbb{R}_+^{J \times J}$ is formed by $\bm X = \bm H\bm H^{\top}$ and its rank-$K$ square root decomposition is performed, i.e., $\bm X = \bm U \bm U^{\top}$. The matrix $\bm U \in \mathbb{R}^{J \times K} $ resulted from the rank-$K$ square root decomposition is input to the algorithms. Both the algorithms are initialized by $\bm Q_{(0)}=\bm I$.  

\begin{figure}[t]
	\centering
	\subfigure[$\varphi=0.7, \psi=0.75$]{
    \includegraphics[width=0.32\linewidth]{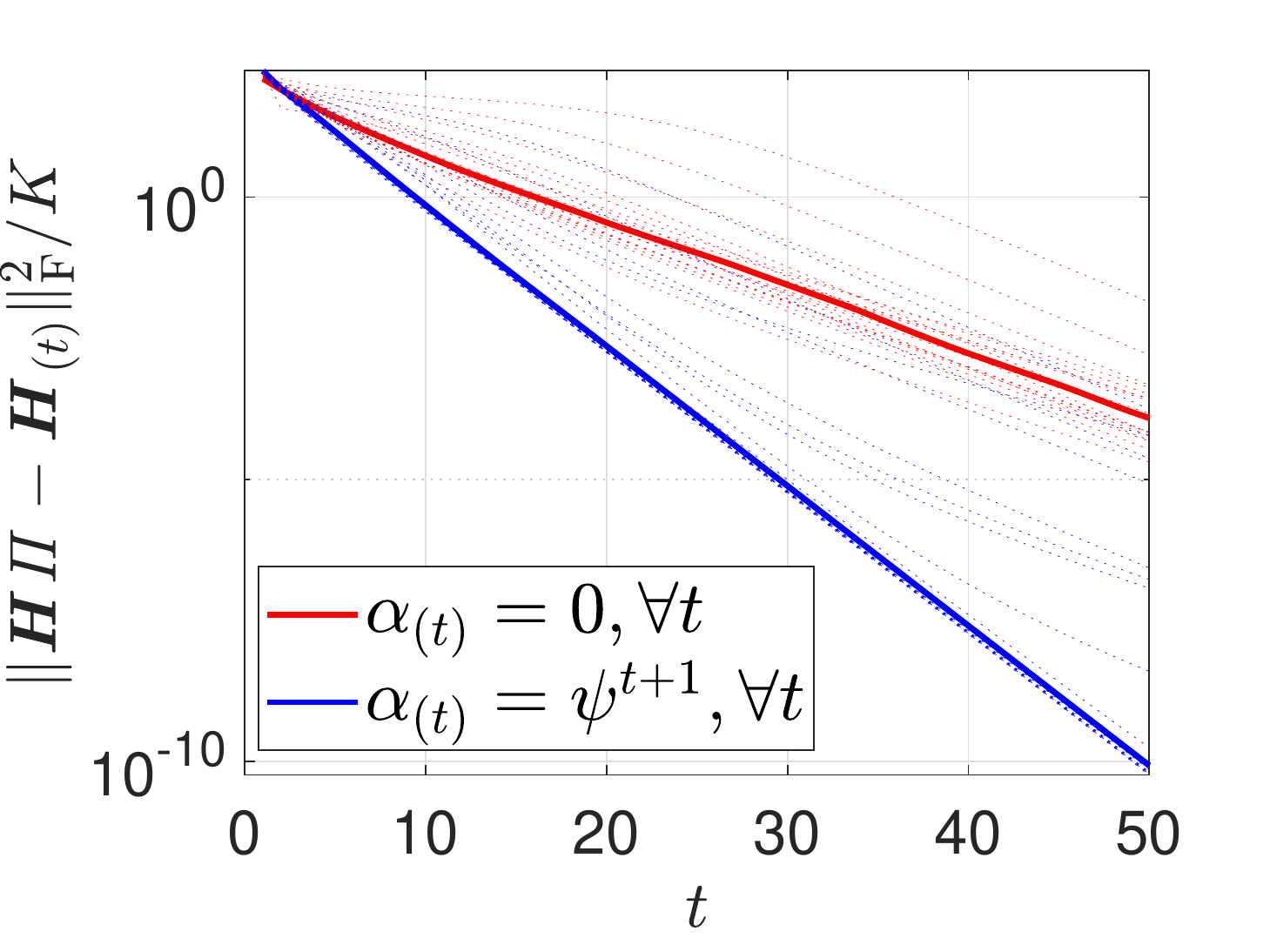}}
    \subfigure[$\varphi=0.5,  \psi=0.75$]{
    \includegraphics[width=0.32\linewidth]{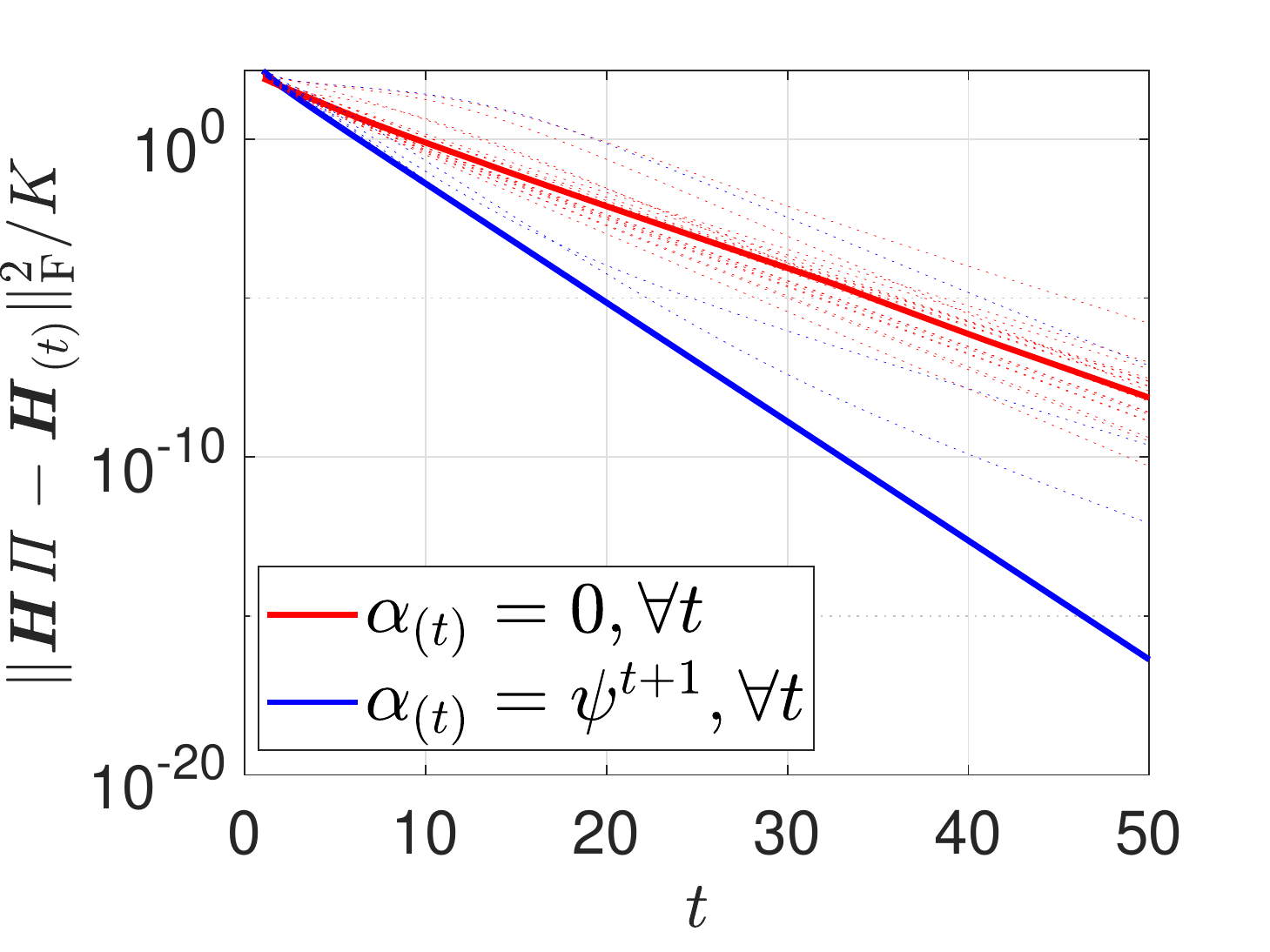}}
    \subfigure[$\varphi=0.3,  \psi=0.75$]{
    \includegraphics[width=0.32\linewidth]{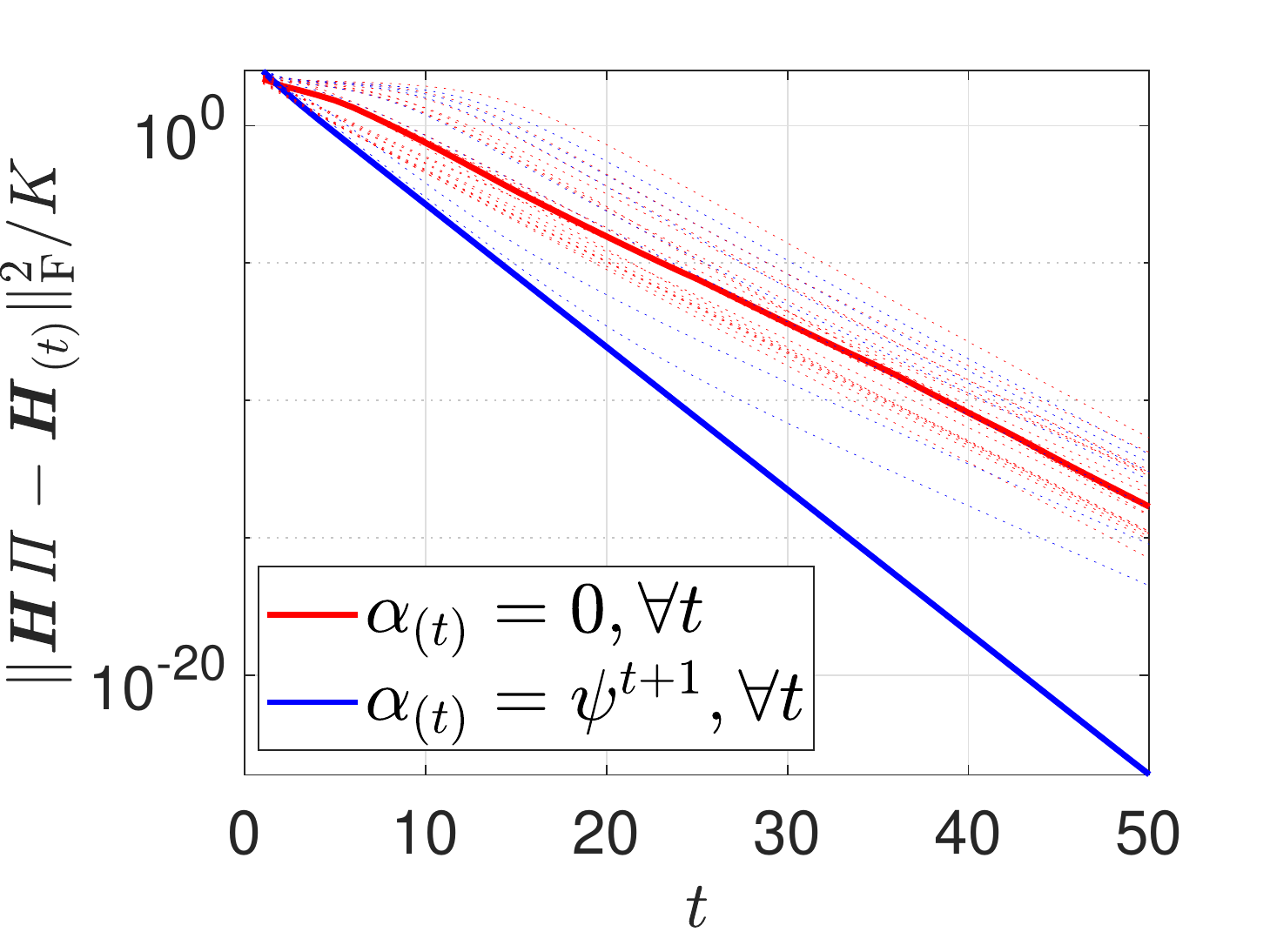}}
	\caption{Convergence of the SymNMF algorithm with $\alpha_{(t)}= \psi^{t+1}$ (proposed)  and $\alpha_{t}=0$ for different levels of sparsity of $\bm H \in \mathbb{R}^{1000\times 3}$ (noiseless case). Dashed line represents each trial and the bold line denotes the median of the 20 independent trials.}
		\label{fig:conv_symnmf1}
\end{figure}

{Fig.~\ref{fig:conv_symnmf1} shows $\|\bm H\bm \varPi-\bm H_{(t)}\|^2_{\rm F}/K$, where $\bm \varPi$ is a permutation matrix, against the iteration index $t$. One can see that for different sparsity levels, the proposed SymNMF algorithm converges faster. It can also be observed that as the sparsity level increases (i.e., $\varphi$ decreases), both SymNMF algorithms converge quickly to low MSE levels.  
}


\begin{figure}[t]
	\centering
	\subfigure[$\varphi=0.7, \psi=0.60$]{
    \includegraphics[width=0.31\linewidth]{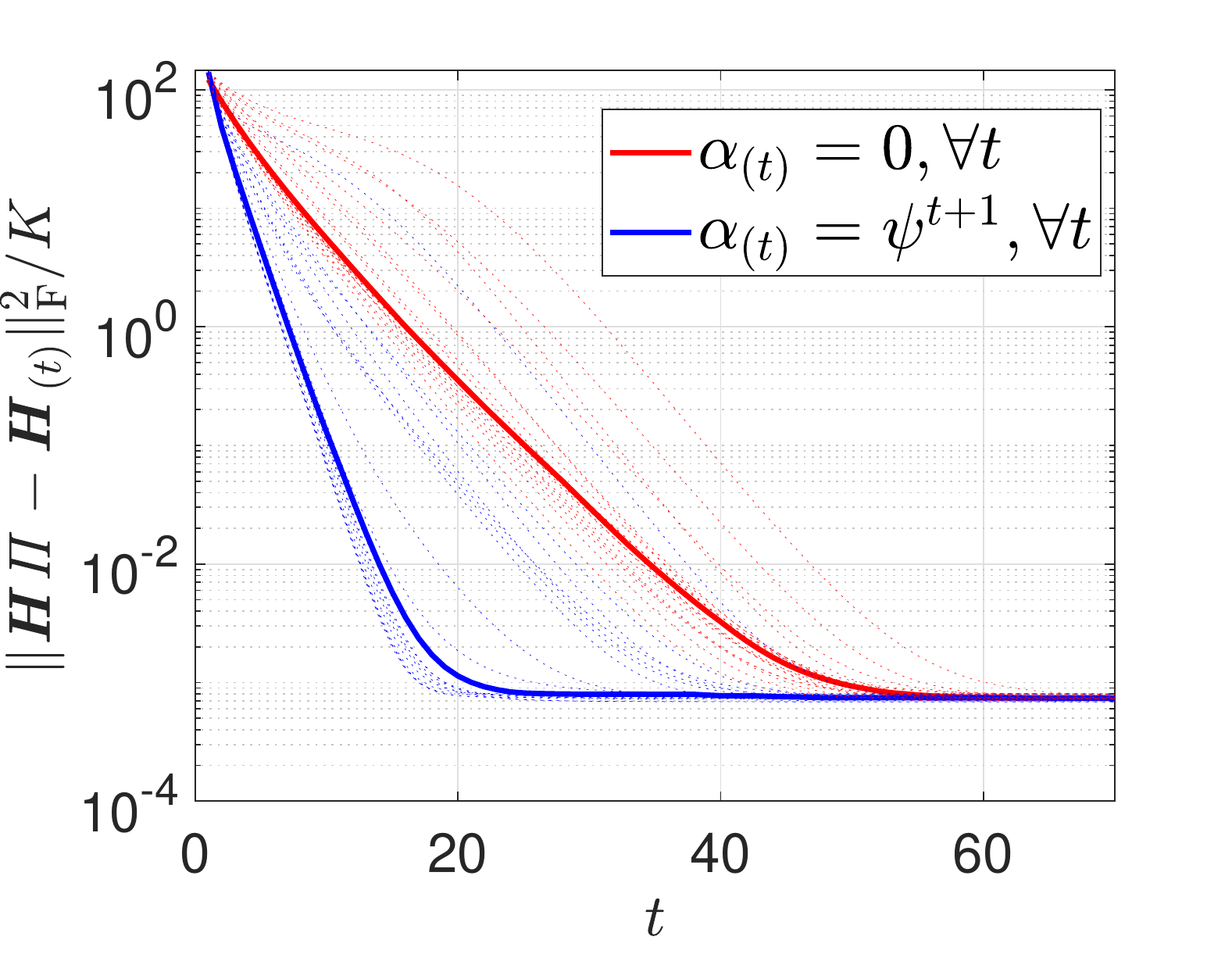}}
    \subfigure[$\varphi=0.5, \psi=0.60$]{
    \includegraphics[width=0.32\linewidth]{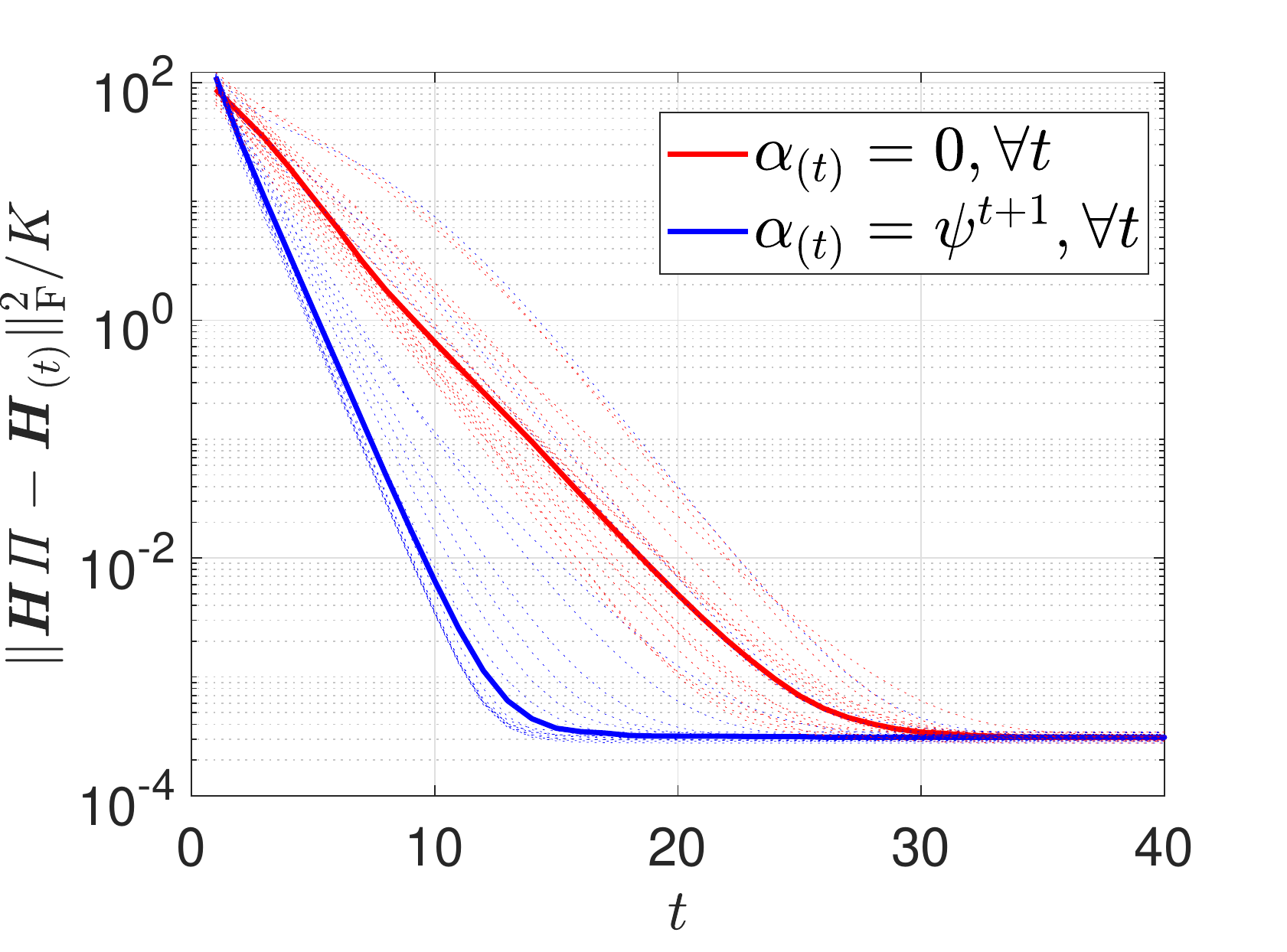}}
    \subfigure[$\varphi=0.3, \psi=0.60$]{
    \includegraphics[width=0.32\linewidth]{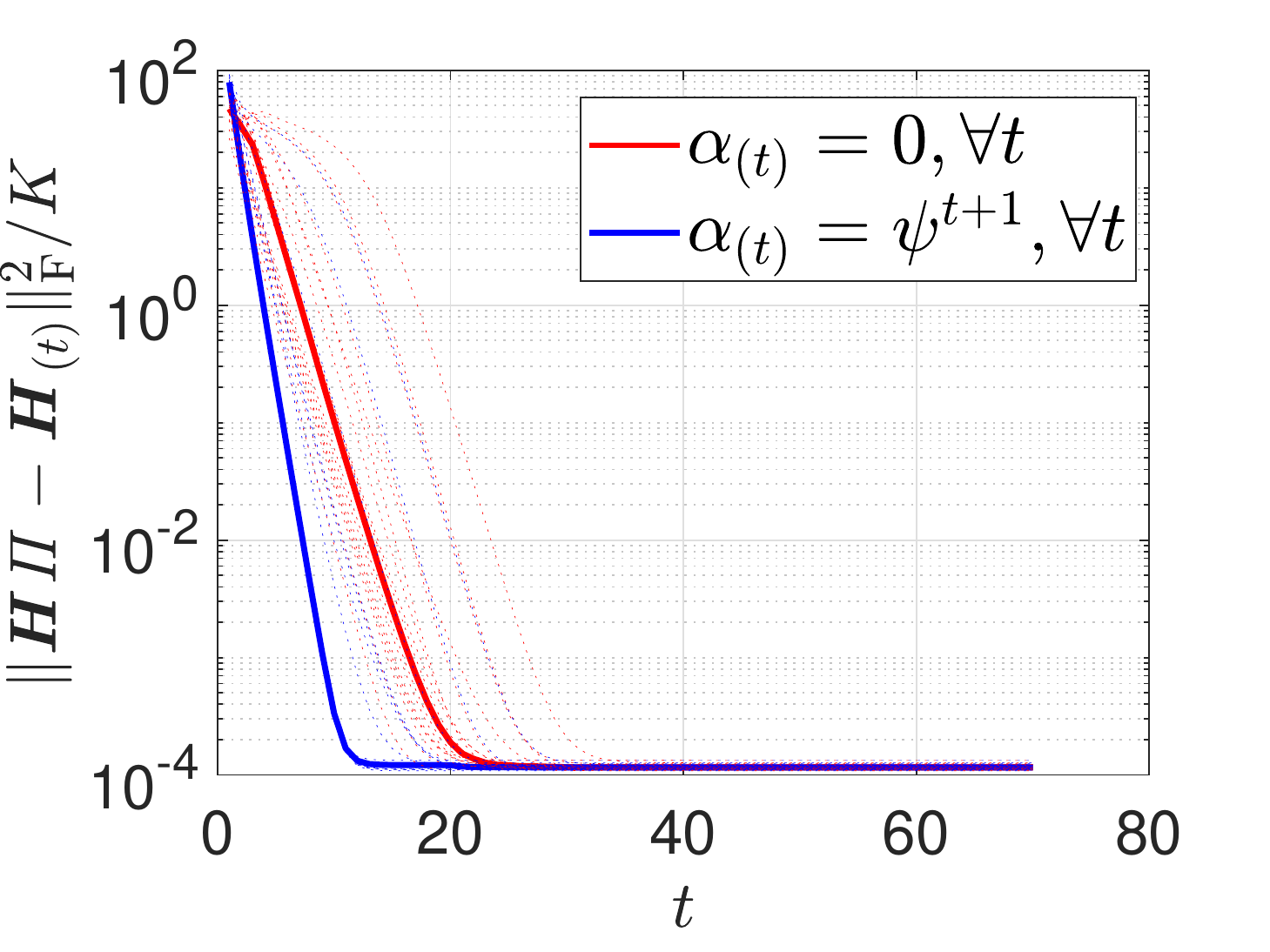}}
	\caption{Convergence of the SymNMF algorithm with $\alpha_{(t)}= \psi^{t+1}$ (proposed) and $\alpha_{t}=0$ for different levels of sparsity of $\bm H \in \mathbb{R}^{1000\times 3}$ and ${\sf SNR}$=30dB. Dashed line represents each trial and the bold line denotes the median of the 20 independent trials.}
	\label{fig:conv_symnmf2}
\end{figure}

Fig.~\ref{fig:conv_symnmf2} shows the convergence behaviour of the algorithms when zero-mean i.i.d. Gaussian noise with variance $\sigma^2$ is added to the matrix $\bm X$. The signal-to-noise ratio (SNR) in dB is defined as ${\sf SNR}=10\log_{10}\left(\frac{\|\bm X\|_{\rm F}^2/J^2 }{\sigma^2} \right)$. The rank-$K$ square root decomposition is performed on the resulted noisy matrix $\widehat{\bm X}$, i.e., $\widehat{\bm X} = \widehat{\bm U}\widehat{\bm U}^{\top}$ and the matrix $\widehat{\bm U}$ is input to the algorithms. In this case as well, one can observe faster convergence for the proposed SymNMF for different sparsity levels.


\subsection{Details of The UCI Data Experiments} \label{supp:uci}
\paragraph{MATLAB Classifiers for UCI Data Experiments.} For UCI data (\url{https://archive.ics.uci.edu/ml/datasets.php}) experiments, we choose 10 different classifiers from the MATLAB statistics and machine learning toolbox (\url{https://www.mathworks.com/products/statistics.html}); see Table~\ref{tab:classifiers}.

\begin{table}[t]
  \centering
  	\caption{Ten Classifiers used As Machine Annotators.}\label{tab:classifiers}
	\begin{tabular}{c   }
		\hline
		\hline
     Coarse $k$-nearest neighbor classifier\\
     \hline
     Medium $k$-nearest neighbor classifier\\
     \hline
    Fine $k$-nearest neighbor classifier\\
    \hline
	Cosine $k$-nearest neighbor classifier\\
	\hline
	Coarse decision tree classifier\\
	\hline
	Medium decision tree classifier\\
	\hline
	Fine decision tree classifier\\
	\hline
	Linear support vector machine (SVM) classifier\\
	\hline
	Quadratic support vector machine (SVM) classifier\\
	\hline
	Coarse Gaussian support vector machine (SVM) classifier\\
    \hline
    \hline
	\end{tabular}
\end{table}

\paragraph{Simulation Setup of Table \ref{tab:uci_car}.}
For the experiment in Table \ref{tab:uci_car}, we employ the following strategy in order to generate different proportions of the missing blocks:  
\begin{enumerate}
    \item Consider $N$ items to be labeled by the annotators (machine classifiers). We split the test data into three disjoint parts having sizes of $0.1N,0.3N$ and $0.6N$, respectively. 
    \item Each disjoint part of the test data is co-labeled by only $P$ annotators, which are chosen randomly from $M$ available annotators and $P \ll M$.  We also make sure that every annotator labels at least one part out of the three test data parts. 
\end{enumerate}
By varying $P$ for the three test data parts, we are able to control the proportions of missing co-occurrences. For each column of the table, we adjust $P$ and generate the cases such that the corresponding missing proportion ({\sf Miss}) is achieved.

In addition, since we have chosen different sizes for the three sets, different annotator pairs co-label varying number of data items. This makes the estimation accuracy for the pairwise statistics $\widehat{\bm R}_{m,j}$'s unbalanced---and we use this setting to test the robustness of our co-occurrence imputation algorithm.

\subsection{Additional Real-Data Experiment} \label{supp:realdata}
In this section, we present an additional real-data experiment. Specifically, we compare the proposed algorithms with a number of deep learning (DL)-based crowdsourcing methods, namely, \texttt{CrowdLayer} and \texttt{DL-MV} from the work in \cite{Rodrigues2018deep}. 

Note that the DL-based methods are implemented under fairly different settings relative to classic D\&S learning methods.
For example,
both DL baselines train a deep neural networks using data items (e.g., images) as (part of the) input, whereas the classic D\&S methods do not need to know or see the data items.


\begin{table}[t]
  \centering
  \caption{Classification error (\%) and runtime (sec.) on the LabelMe dataset ($N=1000$, $M=59$, $K=8$). The ``\texttt{SymNMF}'' family are the proposed methods.}
    			\resizebox{0.33\linewidth}{!}{
    \begin{tabular}{l|c|c}
    \hline
    \textbf{Algorithms} &  \textbf{Error} (\%)  & \textbf{Time} (s) \\ 
    \hline
    \hline
    \texttt{RobSymNMF} & {32.10}  & 1.25 \\
    \hline
    \texttt{RobSymNMF-EM} & {22.10}  & 1.29  \\
    \hline
    \texttt{DesSymNMF} & 29.10 & 0.11   \\
    \hline
    \texttt{DesSymNMF-EM} & {22.20} & 0.20   \\
    \hline
   \texttt{CrowdLayer} & {20.90}  & 15.80  \\
     \hline
  \texttt{DL-MV} & 23.10 & 14.31  \\
    \hline
    \hline
    \end{tabular}%
    }
  \label{tab:deep}%
\end{table}%

The dataset used in this experiment is the LabelMe data that is posted by the authors of \cite{Rodrigues2018deep}. We use 1,000 data items that belong to 8 classes and are labeled by 59 annotators. The methods \texttt{CrowdLayer} and \texttt{DL-MV} are trained with 50 epochs. Table \ref{tab:deep} presents the results of the algorithms under test. 
In the table, the results of \texttt{CrowdLayer} and our method are averaged from 100 trials (to observe performance under random initialization and stochastic algorithms). We observed that \texttt{CrowdLayer}'s and our method's average error rates are close, but \texttt{CrowdLayer} has an almost 10 times larger standard deviation (\texttt{RobSymNMF-EM} $22.1\%\pm 0.5\%$ v.s. \texttt{CrowdLayer} $20.9\% \pm  4.7\%$). 
The proposed method is also around 12 times faster (1.3 sec. vs 15.8 sec.).

\begin{figure}
\centering
\includegraphics[width=0.5\linewidth]{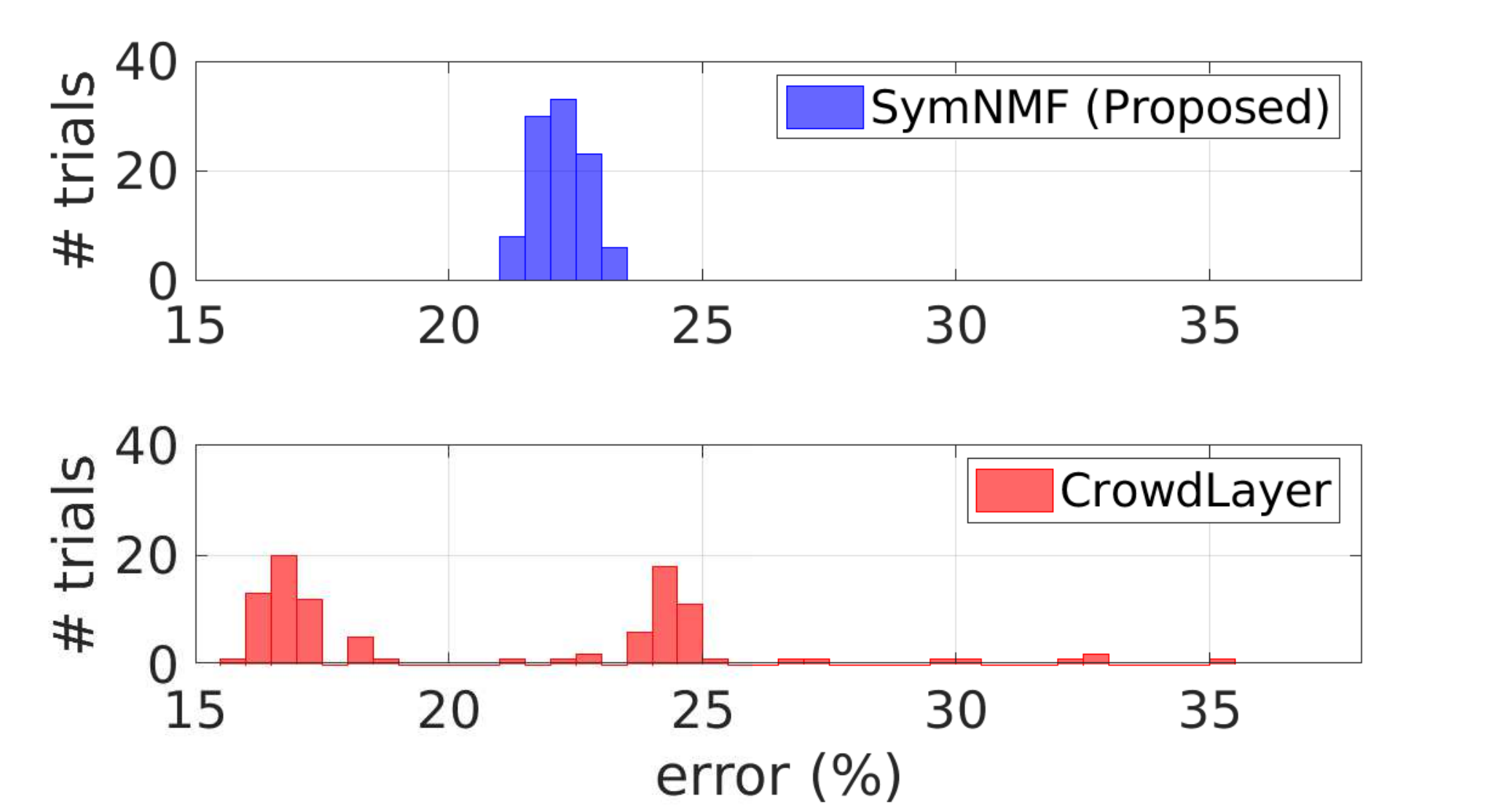}
\caption{ Histograms of error rates from 100 trials. The \texttt{CrowdLayer} method could work very well to attain low error rate in some trials, but multiple failed trials with error rate$\geq 30\%$ are also observed.}\label{fig:hist}
\end{figure}

Fig.~\ref{fig:hist} presents the histogram of the error rates for our method and \texttt{CrowdLayer}. From Fig.~\ref{fig:hist}, one can see that there are trials where \texttt{CrowdLayer} offers impressively low error rate, but there are also multiple trials where \texttt{CrowdLayer} gives high error rates ($\sim 30\%-37\%$).
The large variance is perhaps because DL methods' computational problem is more challenging, since DL algorithms such as SGD/Adam may not always converge well.
However, the proposed method with convergence guarantees offers stable results. 

\section{Proof of Theorem \ref{thm:svdimputation}} \label{supp:svdimputaion}

\vspace{.25cm}

\begin{mdframed}[backgroundcolor=gray!10,topline=false,
	rightline=false,
	leftline=false,
	bottomline=false]
\noindent
{\bf Theorem 1}
{\it
    Assume that $\widehat{\bm R}_{m,n}$ is estimated by \eqref{eq:constructC}-\eqref{eq:estimXmn} using the sample-estimated $\widehat{\R}_{m,r}$, $\widehat{\R}_{n,\ell}$ and $\widehat{\R}_{\ell,r}$ [using \eqref{eq:Rest} with at least $S$ items].
    Also assume that $\kappa(\bm A_m) \le \gamma$ and ${\rm rank}(\A_m)={\rm rank}(\bm D)=K$ for all $m\in[M]$. Let $\varrho = \underset{{(m,j)\in \bm \varOmega}}{\min} \sigma_{\min}(\bm R_{m,j})$. Suppose that $S = \Omega\left(\frac{K^2\gamma^2\log(1/\delta)}{\varrho^4} \right)$ for $\delta >0$. Then, for any $(m,n)\notin \bm \varOmega$, with probability of at least $1-\delta$, we have:
	\begin{align*}
	\|\widehat{\bm R}_{m,n} -\bm R_{m,n}\|_{\rm F} &= O \left(\frac{K^2\gamma^3\sqrt{\log(1/\delta)}}{\varrho^2\sqrt{S}}  \right),
	\end{align*} 
	where $\R_{m,n}=\A_m\D\A_n^\T$ is the missing ground-truth.}
\end{mdframed}

The missing pairwise co-occurrence $\bm R_{m,n}$ is imputed by \eqref{eq:constructC}-\eqref{eq:estimXmn} using available co-occurrences $\R_{m,r}$, $\R_{n,\ell}$ and $\R_{\ell ,r}$.  In practice, we do not observe the true pairwise co-occurrences $\R_{m,r}$, $\R_{n,\ell}$ and $\R_{\ell ,r}$. Therefore, we first form the matrix $\widehat{\bm C}$ by using the corresponding sample estimated co-occurrences as below:
\begin{align*}
\widehat{\bm C} &= [\widehat{\R}_{m,r}^{\top}, \widehat{\R}_{\ell,r}^{\top}]^{\top}.
\end{align*}
To characterize $\|\widehat{\bm C} - \bm C\|_{\rm F}$, we use Lemma 13 from \cite{zhang2014spectral} which gives the result that, with probability at least $1-\delta$, 
\begin{align} \label{eq:Nij2}
\|\widehat{\R}_{m,j}-{\R}_{m,j}\|_{\rm F} \le \frac{1+\sqrt{\log(1/\delta)}}{\sqrt{S}} := \phi,~\forall m \neq j,
\end{align}
where $S>0$ is the number of samples that the annotators $m$ and $j$ have co-labeled.
Then we have
\begin{align}
\|\widehat{\bm C} - \bm C\|^2_{\rm F} &= \|\widehat{\R}_{m,r}-{\R}_{m,r}\|_{\rm F}^2 + \|\widehat{\R}_{\ell,r}-{\R}_{\ell,r}\|_{\rm F}^2 \le 2\phi^2 \nonumber\\
\implies \|\widehat{\bm C} - \bm C\|_{\rm F} &\le \sqrt{2}\phi. \label{eq:Chatbound}
\end{align}
Let us denote the thin SVD operation on $\widehat{\bm C}$ as follows:
\begin{align} \label{eq:svdChat}
\widehat{\bm C} &= [\widehat{\bm U}_m^{\top},\widehat{\bm U}_{\ell}^{\top}]^{\top} \widehat{\bm \Sigma}_{m,\ell,r}\widehat{\bm V}_{r}^{\top}.
\end{align}
We consider the below lemma to characterize this SVD operation:
\begin{lemma} \cite{yu2014useful}\label{lem:svds1}
	Let $\bm C \in \mathbb{R}^{m \times n}$ and $\widehat{\bm C} \in \mathbb{R}^{m \times n}$ have singular values $\sigma_1 \ge \sigma_2 \ge \dots \ge \sigma_{\min(m,n)}$ and $\widehat{\sigma}_1 \ge \widehat{\sigma}_2 \ge \dots \widehat{\sigma}_{\min(m,n)}$, respectively. Fix $1\le t \le s \le {\rm rank}(\bm C)$ and assume that $\min(\sigma_{t-1}^2-\sigma_t^2,\sigma_s^2-\sigma_{s+1}^2) > 0$, where $\sigma_0^2 := \infty$ and $\sigma_{{\rm rank}(\bm C)+1} := 0$. Let $q := s-t+1$ and let $\bm U = \begin{bmatrix}\bm u_t, \bm u_{t+1} ,\ldots, \bm u_{s} \end{bmatrix} \in \mathbb{R}^{m \times q}$ and  $\widehat{\bm U} = \begin{bmatrix}\widehat{\bm u}_t, \widehat{\bm u}_{t+1}, \ldots,  \widehat{\bm u}_{s} \end{bmatrix} \in \mathbb{R}^{m \times q}$ have orthonormal columns satisfying $\bm C^{\top} \bm u_j = \sigma_j \bm v_j$ and $\widehat{\bm C}^{\top} \widehat{\bm u}_j = \widehat{\sigma}_j \widehat{\bm v}_j$ for $j = t,t+1,\dots,s$ and let $\bm V = \begin{bmatrix}\bm v_t, \bm v_{t+1}, \ldots, \bm v_{s} \end{bmatrix} \in \mathbb{R}^{n \times q}$ and  $\widehat{\bm V} = \begin{bmatrix}\widehat{\bm v}_t, \widehat{\bm v}_{t+1} , \ldots, \widehat{\bm v}_{s} \end{bmatrix} \in \mathbb{R}^{n \times q}$ have orthonormal columns satisfying $\bm C \bm v_j = \sigma_j \bm u_j$ and $\widehat{\bm C} \widehat{\bm v}_j = \widehat{\sigma}_j \widehat{\bm u}_j$ for $j = t,t+1,\dots,s$. Then there exists an orthogonal matrix ${\bm O} \in \mathbb{R}^{q \times q}$ such that 
	\begin{align*}
	\|\widehat{\bm U}  - \bm U{\bm O}\|_{\rm F} &\le \frac{2^{3/2}(2\sigma_1+\|\widehat{\bm C}-\bm C\|_{2})\min(q^{1/2}\|\widehat{\bm C}-\bm C\|_{2},\|\widehat{\bm C}-\bm C\|_{\rm F})}{\min(\sigma_{t-1}^2-\sigma_t^2,\sigma_s^2-\sigma_{s+1}^2)}
	\end{align*}
	and the same upper bound holds for $\|\widehat{\bm V}  - \bm V{\bm O}\|_{\rm F}$.
\end{lemma}

For now, let us assume
\begin{align} \label{eq:condC_S}
{\rm rank}(\bm C) = K,\quad \|\widehat{\bm C}-\bm C\|_2 \le\|\bm C\|_2=\sigma_{\max}(\bm C).
\end{align}
By applying Lemma \ref{lem:svds1} in \eqref{eq:svdChat},
we get
\begin{align*}
\|\widehat{{\bm U}}_{m}  - {{\bm U}}_{m}{\bm O}\|_{\rm F} 
&\le \frac{2^{3/2}\sqrt{K}3\sigma_{\max}(\bm C)\|\widehat{\bm C}-\bm C\|_{2}}{\sigma_{\min}^2(\bm C)}.
\end{align*}
where $\bm O \in \mathbb{R}^{K \times K}$ is orthogonal.

By substituting the bound \eqref{eq:Chatbound} in the above, we get that with probability of at least $1-\delta$,
\begin{align}
\|\widehat{{\bm U}}_{m}  - {{\bm U}}_{m}{\bm O}\|_{\rm F} &\le \frac{12\sqrt{K}\sigma_{\max}(\bm C)\phi}{\sigma_{\min}^2(\bm C)} \label{eq:Um_noise},\\
\|\widehat{{\bm U}}_{\ell}  - {{\bm U}}_{\ell}{\bm O}\|_{\rm F} &\le \frac{12\sqrt{K}\sigma_{\max}(\bm C)\phi}{\sigma_{\min}^2(\bm C)}. \label{eq:Uell_noise}
\end{align}
The missing co-occurrence ${\bm R}_{m,n}$ is imputed by using $\widehat{\bm U}_m$, $\widehat{\bm U}_{\ell}$ and $\widehat{\bm R}_{n,\ell}$ via the following operation:
\begin{align*}
\widehat{\bm R}_{m,n} = \widehat{\bm U}_m \widehat{\bm U}_{\ell}^{-1}\widehat{\bm R}_{n,\ell}^{\top}.
\end{align*} 

The first term $\widehat{\bm U}_m$ is characterized by \eqref{eq:Um_noise}. To characterize the term $\widehat{\bm U}_{\ell}^{-1}$, we use the following lemma:

\begin{lemma}  \label{lem:inverse}
	Consider any matrices $\bm Y,\bm Z , \bm E \in \mathbb{R}^{K \times K}$ such that $\bm Z = \bm Y+\bm E$ and $\bm Y$ is invertible. Suppose that ${\rm rank}(\bm Z) = {\rm rank}(\bm Y)$ and that $\|\bm E \|_2 \le \sigma_{\min}(\bm Y)/2 $ . Then, we have
	\begin{align*}
	\|\bm Z^{-1}-\bm Y^{-1}\|_2 \le \frac{2\|\bm E \|_2}{\sigma_{\min}^2(\bm Y)}.
	\end{align*}
\end{lemma}
The proof of the lemma can be found in Section \ref{supp:lem_inverse}.

Applying Lemma~\ref{lem:inverse} by letting $\bm Y := {{\bm U}}_{\ell}\bm O$ and $\bm Z :=\widehat{{\bm U}}_{\ell}$, we get
\begin{align} \label{eq:Upseudo}
\|\widehat{{\bm U}}_{\ell}^{-1} - {({\bm U}}_{\ell}\bm O)^{-1}\|_2 \le \frac{2}{\sigma_{\min}^2({{\bm U}}_{\ell})} \|\widehat{{\bm U}}_{\ell}- {({\bm U}}_{\ell}\bm O)\|_2.
\end{align}

We proceed to characterize $\sigma_{\min}({{\bm U}}_{\ell})$ in the above relation by utilizing the following result:
\begin{lemma} \label{lem:sigmaU}
	Suppose that $\kappa(\bm A_m) \le \gamma$, for all $m$. Then,  we have
	\begin{align*}
	\sigma_{\min}({\bm U}_{\ell}) &\ge  \frac{1}{\sqrt{2K}\gamma}, ~~ \sigma_{\max}({\bm U}_{\ell}) \le   \gamma 
	\end{align*}
	and the above bounds are applicable for $\bm U_m$ as well.
\end{lemma}
The proof of the lemma can be found in Section \ref{supp:lem_sigmaU}.

Applying Lemma \eqref{lem:sigmaU} in \eqref{eq:Upseudo}, we get 
\begin{align} 
\|\widehat{{\bm U}}_{\ell_1}^{-1} - {({\bm U}}_{\ell_1}\bm O)^{-1}\|_2 &\le  {4K\gamma^2} \|\widehat{{\bm U}}_{\ell_1}- {({\bm U}}_{\ell_1}\bm O)\|_2 \nonumber\\
&\le \frac{48K\sqrt{K}\gamma^2\sigma_{\max}(\bm C)\phi}{\sigma_{\min}^2(\bm C)},\label{eq:leastsquares2}
\end{align}
where we applied \eqref{eq:Uell_noise} in the last inequality.

Using the above derived upper bounds, we proceed to bound the following term:
\begin{align*}
\|\widehat{\bm R}_{m,n} -\bm R_{m,n}\|_2 = \|\widehat{\bm U}_m \widehat{\bm U}_{\ell}^{-1}\widehat{\bm R}_{n,\ell}^{\top} - {\bm U}_m {\bm U}_{\ell}^{-1}{\bm R}_{n,\ell}^{\top}\|_2,
\end{align*}
    where we can see that ${\bm U}_m {\bm U}_{\ell}^{-1}{\bm R}_{n,\ell}^{\top} = {\bm U}_m \bm O ({\bm U}_{\ell}\bm O)^{-1}{\bm R}_{n,\ell}^{\top}$ for the orthogonal matrix $\bm O$.
   To simplify the notations, let us define $\bm Z_1 : = {\bm U}_m \bm O$, $\bm Z_2 := ({\bm U}_{\ell}\bm O)^{-1}$ and $\bm Z_3 := {\bm R}_{n,\ell}^{\top}$.  We also define $\widehat{\bm Z}_1 : = \widehat{\bm U}_m$, $\widehat{\bm Z}_2 := \widehat{\bm U}_{\ell}^{-1}$ and $\bm Z_3 := \widehat{\bm R}_{n,\ell}^{\top}$. Using these notations, we have the following set of relations:
\begin{align}
\left\|\widehat{\bm Z}_1\widehat{\bm Z}_2\widehat{\bm Z}_3-\bm Z_1\bm Z_2\bm Z_3\right\|_{2}
&= \left\|\widehat{\bm Z}_1\widehat{\bm Z}_2\widehat{\bm Z}_3-\bm Z_1\bm Z_2\bm Z_3 - \widehat{\bm Z}_1\bm Z_2\bm Z_3+\widehat{\bm Z}_1\bm Z_2\bm Z_3\right\|_{2} \nonumber\\
&= \left\|\left(\widehat{\bm Z}_1-\bm Z_1\right)\bm Z_2\bm Z_3+\widehat{\bm Z}_1\left(\widehat{\bm Z}_2\widehat{\bm Z}_3-\bm Z_2\bm Z_3\right)\right\|_2 \nonumber\\
&\le \left\|\left(\widehat{\bm Z}_1-\bm Z_1\right)\bm Z_2\bm Z_3\right\|_2+\left\|\widehat{\bm Z}_1\left(\widehat{\bm Z}_2\widehat{\bm Z}_3-\bm Z_2\bm Z_3\right)\right\|_2 \nonumber\\
&= \left\|\left(\widehat{\bm Z}_1-\bm Z_1\right)\bm Z_2\bm Z_3\right\|_2 +\left\|\widehat{\bm Z}_1(\widehat{\bm Z}_2-\bm Z_2)\bm Z_3+\widehat{\bm Z}_1\widehat{\bm Z}_2\left(\widehat{\bm Z}_3-\bm Z_3\right)\right\|_2 \nonumber\\
&\le \|\bm Z_2\|_2\|\bm Z_3\|_2\left\|\widehat{\bm Z}_1-\bm Z_1\right\|_2+\|\widehat{\bm Z}_1\|_2\|\bm Z_3\|_2 \left\|\widehat{\bm Z}_2-\bm Z_2\right\|_2  +\|\widehat{\bm Z}_1\|_2\|\widehat{\bm Z}_2\|_2 \left\|\widehat{\bm Z}_3-\bm Z_3\right\|_2,\nonumber 
\end{align}
where we have used triangle inequality to obtain the first inequality and used the fact that $\|\bm X\bm Y\|_2 \le \|\bm X\|_2 \|\bm Y\|_2$ in the last inequality. Applying this result, we get
\begin{align}
\|\widehat{\bm U}_m \widehat{\bm U}_{\ell}^{-1}\widehat{\bm R}_{n,\ell}^{\top} - {\bm U}_m \bm O ({\bm U}_{\ell}\bm O)^{-1}{\bm R}_{n,\ell}^{\top}\|_2 &\le \|({\bm U}_{\ell}{\bm O})^{-1}\|_2\|{\bm R}_{n,\ell}\|_2\|\widehat{\bm U}_m -{\bm U}_{m}{\bm O}\|_2 \nonumber\\
&\quad \quad+\|\widehat{\bm U}_m\|_2\|{\bm R}_{n,\ell}\|_2 \|\widehat{\bm U}_{\ell}^{-1}-({\bm U}_{\ell}{\bm O})^{-1}\|_2 \nonumber \\
&\quad \quad  +\|\widehat{\bm U}_m\|_2\|\widehat{\bm U}_{\ell}^{-1}\|_2 \|\widehat{\bm R}_{n,\ell}-{\bm R}_{n,\ell}\|_2. \label{eq:product_bound}
\end{align}
In \eqref{eq:product_bound}, we need to apply the below characterizations to derive the final bound:
\begin{enumerate}
	\item \textbf{Upper bound for $\|\widehat{{\bm U}}_{m}\|_2$}
	\begin{align*}
	\|\widehat{{\bm U}}_{m}\|_2 
	&= \|\widehat{{\bm U}}_{m} - {{\bm U}}_{m}\bm O+{{\bm U}}_{m}\bm O\|_2 \le \|\widehat{{\bm U}}_{m} - {{\bm U}}_{m}\bm O\|_2 + \|{{\bm U}}_{m}\bm O\|_2\\
	&\le \sigma_{\max}( {{\bm U}}_{m}) + \sigma_{\max}( {{\bm U}}_{m}) = 2\sigma_{\max}( {{\bm U}}_{m}) \le 2\gamma.
	\end{align*}
	where we have used triangle inequality for the first inequality, used the assumption that $\|\widehat{{\bm U}}_{m} - {{\bm U}}_{m}\bm O\|_2 \le \sigma_{\min}( {{\bm U}}_{m})/2$ for the second inequality and invoked Lemma~\ref{lem:sigmaU} for the last inequality.
	
	\item \textbf{Upper bound for $\|({{\bm U}_{\ell}{\bm O})^{-1}}\|_2$}
	\begin{align*}
\|({{\bm U}_{\ell}{\bm O})^{-1}}\|_2 = 1/\sigma_{\min}({\bm U}_{\ell}) \le \sqrt{2K}\gamma,
	\end{align*}
	where we have applied Lemma~\ref{lem:sigmaU} for the last inequality.
	\item \textbf{Upper bound for $\|\widehat{{\bm U}}_{\ell}^{-1}\|_2$}
	\begin{align*}
\|\widehat{{\bm U}}_{\ell}^{-1}\|_2 
	&= 1/\sigma_{\min}(\widehat{{\bm U}}_{\ell}) \le 2/\sigma_{\min}({\bm U}_{\ell})  \le 2\sqrt{2K}\gamma,
	\end{align*}
	where we have used the assumption that $\|\widehat{{\bm U}}_{\ell} - {{\bm U}}_{\ell}\bm O\|_2 \le \sigma_{\min}( {{\bm U}}_{\ell})/2$ for the first inequality and invoked Lemma~\ref{lem:sigmaU} for the last inequality.
	
	\item \textbf{Upper bound for $\| {\bm R}_{n,\ell}\|_2$}
	\begin{align*}
	\| {\bm R}_{n,\ell}\|_2 \le \| {\bm R}_{n,\ell}\|_{\rm F} \le 1,
	\end{align*}
	where we used the fact that the entries of the matrix ${\bm R}_{n,\ell}$ are nonnegative and sum to one and therefore $\| {\bm R}_{n,\ell}\|_{\rm F}^2 \le 1$.
\end{enumerate}

Applying these upper bounds to \eqref{eq:product_bound}, we attain the following:
\begin{align*}
\|\widehat{\bm R}_{m,n} -\bm R_{m,n}\|_2 
&\le \sqrt{2K} \gamma \|\widehat{\bm U}_m -{\bm U}_m \bm O  \|_2 + 2\gamma \|\widehat{\bm U}_{\ell}^{-1}-  ({\bm U}_{\ell}{\bm O})^{-1} \|_{2} + 4\sqrt{2K} \gamma^2\|\widehat{\bm R}_{n,\ell}-{\bm R}_{n,\ell}\|_2\\
\implies\|\widehat{\bm R}_{m,n} -\bm R_{m,n}\|_{\rm F} &\le \sqrt{2}K \gamma \|\widehat{\bm U}_m -{\bm U}_m \bm O  \|_{\rm F} + 2\sqrt{K}\gamma \|\widehat{\bm U}_{\ell}^{-1}-  ({\bm U}_{\ell}{\bm O})^{-1} \|_{2} + 4\sqrt{2}K\gamma^2\|\widehat{\bm R}_{n,\ell}-{\bm R}_{n,\ell}\|_{\rm F},
\end{align*}
where we used the matrix norm equivalence $\|\bm X\|_2 \le \|\bm X\|_{\rm F} \le \sqrt{K}\|\bm X\|_2$, for a matrix $\bm X$ of rank $K$, in the last inequality.

By substituting the bounds \eqref{eq:Nij2}, \eqref{eq:Um_noise} and \eqref{eq:leastsquares2} in the above, we get
\begin{align*}
\|\widehat{\bm R}_{m,n} -\bm R_{m,n}\|_{\rm F} &\le \frac{12\sqrt{2K}K\gamma\sigma_{\max}(\bm C)\phi}{\sigma_{\min}(\bm C)^2} +  \frac{96K^2\gamma^3\sigma_{\max}(\bm C)\phi}{\sigma_{\min}^2(\bm C)} + 4\sqrt{2}K \gamma^2 \phi.
\end{align*}
where we have $\bm C =[\R_{m,r}^{\top}, \R_{\ell,r}^{\top}]^{\top}$ and can immediately see that $\|\bm C\|^2_{\rm F} \le 2$, which implies that $\sigma_{\max}(\bm C) \le \|\bm C\|_{\rm F} \le \sqrt{2}$ and $\phi =\frac{1+\sqrt{\log(1/\delta)}}{\sqrt{S}}$.
Combining this, we get that with probability at least $1-\delta$, for a certain constant $C_1>0$,
\begin{align}
\|\widehat{\bm R}_{m,n} -\bm R_{m,n}\|_{\rm F} &\le \frac{C_1K^2\gamma^3\sqrt{\log(1/\delta)}}{\sigma^2_{\min}(\bm C)\sqrt{S}}  . \label{eq:Rmnbound}
\end{align}


Finally, we will summarize the conditions to be satisfied to obtain \eqref{eq:Rmnbound}. From \eqref{eq:condC_S}, we can see that the below condition needs to be satisfied:
\begin{align}
 \|\widehat{\bm C}-\bm C\|_2 \le\|\bm C\|_2&=\sigma_{\max}(\bm C)~\implies \sqrt{2}\phi \le \sigma_{\max}(\bm C) \nonumber\\
  \implies S &\le \frac{2(1+\sqrt{\log(1/\delta)})^2}{\sigma^2_{\max}(\bm C)}.\label{eq:Sbound1}
\end{align}
From Lemma \ref{lem:inverse}, the condition to be satisfied is:
\begin{align} \label{eq:assump_Uell}
    \|\widehat{{\bm U}}_{\ell} - {{\bm U}}_{\ell}\bm O\|_2 \le \sigma_{\min}( {{\bm U}}_{\ell})/2.
\end{align}
By applying \eqref{eq:Uell_noise} and Lemma \ref{lem:sigmaU} in the left and right hand sides of \eqref{eq:assump_Uell}, respectively, the condition to be satisfied can be re-written as:
\begin{align}
\frac{12\sqrt{K}\sigma_{\max}(\bm C)\phi}{\sigma_{\min}(\bm C)^2} \le \frac{1}{\sqrt{2K}\gamma}~&\implies \frac{12\sqrt{2}\sqrt{K}\sigma_{\max}(\bm C)(1+\sqrt{\log(1/\delta))}}{\sigma_{\min}(\bm C)^2\sqrt{S}} \le \frac{1}{\sqrt{2K}\gamma},\nonumber\\
\implies S &\ge\frac{C_2K^2\gamma^2\log(1/\delta)}{\sigma_{\min}(\bm C)^2}, \label{eq:Sbound2}
\end{align}
for a certain constant $C_2 >0$. Combining \eqref{eq:Sbound1} and \eqref{eq:Sbound2}, we get the final condition on $S$ as stated in the theorem.

\section{Proof of Theorem \ref{thm:cf_stability}} \label{supp:cf_stability}

\vspace{.25cm}

\begin{mdframed}[backgroundcolor=gray!10,topline=false,
	rightline=false,
	leftline=false,
	bottomline=false]
\noindent
{\bf Theorem 2}
{\it
Assume that the $\widehat{\R}_{m,j}$'s are estimated using \eqref{eq:Rest} with $S_{m,j}= |{\cal S}_{m,j}|$ for all $(m,j)\in\bm \varOmega$.	
{Also assume that each $\widehat{\bm R}_{m,j}$ is observed with the same probability.}
Let $ \{\bm U_m^*,\bm U_j^\ast\}_{ (m,j) \in \bm \varOmega  }$ be any optimal solution of \eqref{eq:optim_problem}. {Define $L = M(M-1)/2$}. Then we have
	\begin{align} 
	&\frac{1}{L}\sum_{m< j}\|{\bm U}_m^*(\bm U_{j}^*)^\T-\bm R_{m,j}\|_{\rm F} \le  C\sqrt{\frac{MK^2\log(M) }{|\bm  \varOmega|}  }+ \left(\frac{1}{|\bm \varOmega|}+\frac{1}{L}\right)  \sum_{(m,j) \in \bm \varOmega}\frac{1+\sqrt{M}}{\sqrt{S_{m,j}}}, \nonumber
	\end{align}
with probability of at least $1- 3\exp(-M)$, where $C>0$.}

\end{mdframed}

Let $\bm R^*_{m,j} = \bm U_m^*\bm U_j^{*\top}$, where $ \{\bm U_m^*,\bm U_j^\ast\}_{ (m,j) \in \bm \varOmega  }$ be any optimal solution of \eqref{eq:optim_problem} and $\bm N_{m,j} = \widehat{\bm R}_{m,j}-\bm R_{m,j}$ for every $m,j$.
{Note that we treat $\bm N_{m,j}=\bm 0$ for $(m,j)\notin \bm \varOmega$, since the co-occurrences are unobserved.}
We define the following quantity that will be useful in our proof:
\begin{align} \label{eq:tau}
\tau (\bm \varOmega) = \left|\frac{1}{|\bm \varOmega|}\sum_{(m,j) \in \bm \varOmega}\|\widehat{\bm R}_{m,j}-\bm R_{m,j}^*\|_{\rm F}- \frac{1}{L}\sum_{m<j}\|\widehat{\bm R}_{m,j}-\bm R_{m,j}^*\|_{\rm F}\right|,
\end{align}
where $L = M(M-1)/2$.
Then we have
\begin{align}
\frac{1}{L}\sum_{m<j}\|{\bm R}^*_{m,j}-\bm R_{m,j}\|_{\rm F} &= \frac{1}{L}\sum_{m<j}\|{\bm R}^*_{m,j}-\widehat{\bm R}_{m,j}+\bm N_{m,j}\|_{\rm F} \nonumber\\
&\stackrel{(a)}{\le} \frac{1}{L}\sum_{m<j}\|{\bm R}^*_{m,j}-\widehat{\bm R}_{m,j}\|_{\rm F}+\frac{1}{L}\sum_{m<j}\|\bm N_{m,j}\|_{\rm F} \nonumber\\
&\stackrel{(b)}{\le} \frac{1}{|\bm \varOmega|}\sum_{(m,j) \in \bm \varOmega}\|\widehat{\bm R}_{m,j}-\bm R_{m,j}^*\|_{\rm F} + \tau (\bm \varOmega)+\frac{1}{L}\sum_{m<j}\|\bm N_{m,j}\|_{\rm F}\nonumber\\
&\stackrel{(c)}{\le} \frac{1}{|\bm \varOmega|}\sum_{(m,j) \in \bm \varOmega}\|\widehat{\bm R}_{m,j}-\bm R_{m,j}\|_{\rm F} + \tau (\bm \varOmega)+\frac{1}{L}\sum_{m<j}\|\bm N_{m,j}\|_{\rm F}\nonumber\\
&=  \frac{1}{|\bm \varOmega|}\sum_{(m,j) \in \bm \varOmega}\|\bm N_{m,j}\|_{\rm F} + \tau (\bm \varOmega)+\frac{1}{L}\sum_{m<j}\|\bm N_{m,j}\|_{\rm F} \label{eq:rmse},
\end{align}
where $(a)$ is due to triangle inequality, $(b)$ is due to the definition of $\tau(\bm \varOmega)$ and triangle inequality, and $(c)$ is due to the fact that $\bm R^*_{m,j}$ is the optimal solution of \eqref{eq:optim_problem}.

Next, we will characterize $\tau (\bm \varOmega)$. For this, let us define the set
\begin{align*}
\mathcal{S}_K = \{\bm X = \bm U \bm V^{\top} \in \mathbb{R}^{MK \times MK} : {\rm rank}(\bm X) \le K,~ \|\bm U \|_{\rm F} \le B, ~\|\bm V \|_{\rm F} \le B \},
\end{align*}
where the constant $B = \sqrt{M}D$ and $D$ is the constant from Problem \eqref{eq:optim_problem}. 

If $\|\bm U \|_{\rm F} \le B$ and $\|\bm V \|_{\rm F} \le B $, then $\|\bm X\|_F \le \|\bm U\|_{\rm F} \|\bm V\|_{\rm F} = B^2$.
Therefore, we can rewrite the definition of the set $\mathcal{S}_K$ as below:
\begin{align} \label{eq:SKset}
\mathcal{S}_K = \{\bm X \in \mathbb{R}^{MK \times MK} : {\rm rank}(\bm X) \le K,~ \|\bm X\|_{\rm F} \le B^2\}.
\end{align}

We will invoke the following lemma to characterize the covering number of the set $\mathcal{S}_K$.

\begin{lemma} \cite{wang2012stability} \label{lem:covering}
	Let $\mathcal{S}_r = \{\bm X \in \mathbb{R}^{n_1 \times n_2} : {\rm rank}(\bm X) \le r,~ \|\bm X\|_{\rm F} \le C  \}$. Then there exists an $\epsilon$-net $\overline{\mathcal{S}}_r$ for the Frobenius norm obeying
	\begin{align*}
	|\overline{\mathcal{S}}_r(\epsilon)| \le (9C/\epsilon)^{(n_1+n_2+1)r}.
	\end{align*}
\end{lemma}

By denoting the $\epsilon$-net of $\mathcal{S}_K$ defined in \eqref{eq:SKset} as $\overline{\mathcal{S}}_K(\epsilon)$ and applying Lemma~\ref{lem:covering}, we get that
\begin{align} \label{eq:covering1}
|\overline{\mathcal{S}}_K(\epsilon)| \le (9B^2/\epsilon)^{(2MK+1)K}.
\end{align}
Let $\widetilde{\bm X} \in \overline{\mathcal{S}}_K(\epsilon)$ and we can define the following:
\begin{subequations}\label{eq:calLs}
	\begin{align}
	\widehat{\mathcal{L}}(\widetilde{\bm X}) &= \frac{1}{|\bm \varOmega|}\sum_{(m,j) \in \bm \varOmega}\|\widehat{\bm R}_{m,j}-\widetilde{\bm X}_{m,j}\|_{\rm F}  \\
	{\mathcal{L}}(\widetilde{\bm X}) &= \frac{1}{L}\sum_{m<j}\|\widehat{\bm R}_{m,j}-\widetilde{\bm X}_{m,j}\|_{\rm F},
	\end{align}
\end{subequations}
where $\widetilde{\bm X}_{m,j} \in \mathbb{R}^{K \times K}$ is the $(m,j)$th block of $\widetilde{\bm X} \in \mathbb{R}^{MK \times MK}$.

To proceed, consider the below lemma:
\begin{lemma} \cite{serfling1974probability}\label{lem:hoeffding}
	Let $X =[X_1,\dots, X_n]$ be a set of samples taken without replacement from a set $\{x_1,\dots,x_N\}$ with mean $u$ where $n \le N$. Denote $a := \max_i x_i$ and $b := \max_i b_i$. Then, we have
	\begin{align*}
	{\sf Pr}\left(\left|\frac{1}{n}\sum_{i=1}^n X_i- u\right| \ge t \right) \le 2 \exp\left(-\frac{2nt^2}{\left(1-\frac{n-1}{N}\right)(b-a)^2} \right).
	\end{align*}
\end{lemma}

Notice that $\{\|\widehat{\bm R}_{m,j}-\widetilde{\bm X}_{m,j}\|_{\rm F}\}_{m<j}$ forms a set of $L$ elements with ${\mathcal{L}}(\widetilde{\bm X})$ as its mean and $\widehat{\mathcal{L}}(\widetilde{\bm X})$ as the mean estimated from $|\bm \varOmega|$ samples, drawn without replacement. Also, we have
\begin{align*}
\max_{m,j} \|\widehat{\bm R}_{m,j}-\widetilde{\bm X}_{m,j}\|_{\rm F}&\le \max_{m,j} \|\widehat{\bm R}_{m,j}\|_{\rm F}+\|\widetilde{\bm X}_{m,j}\|_{\rm F}\le 2, \\
\min_{m,j} \|\widehat{\bm R}_{m,j}-\widetilde{\bm X}_{m,j}\|_{\rm F} &= 0 .
\end{align*}
Therefore, by applying Lemma~\ref{lem:hoeffding}, we have
\begin{align*}
{\sf Pr}\left(\left|\widehat{\mathcal{L}}(\widetilde{\bm X})-{\mathcal{L}}(\widetilde{\bm X})  \right| \ge t  \right)\le 2 \exp\left(-\frac{2|\bm \varOmega|t^2}{\left(1-\frac{|\bm \varOmega|-1}{L}\right)4} \right).
\end{align*}

Applying union bound over every $\widetilde{\bm X} \in  \overline{\mathcal{S}}_K(\epsilon)$, we get
\begin{align*}
{\sf Pr}\left(\underset{\widetilde{\bm X} \in  \overline{\mathcal{S}}_K(\epsilon)}{\rm sup}\left|\widehat{\mathcal{L}}(\widetilde{\bm X})-{\mathcal{L}}(\widetilde{\bm X})  \right| \ge t  \right)\le 2|\overline{\mathcal{S}}_K(\epsilon)| \exp\left(-\frac{2|\bm \varOmega|t^2}{\left(1-\frac{|\bm \varOmega|-1}{L}\right)4} \right).
\end{align*}

By letting $|\overline{\mathcal{S}}_K(\epsilon)| \exp\left(-\frac{2|\bm \varOmega|t^2}{\left(1-\frac{|\bm \varOmega|-1}{L}\right)4} \right) = \exp(-M)$, we get that
\begin{align*}
\log~ |\overline{\mathcal{S}}_K(\epsilon)| &-\frac{2L|\bm \varOmega|t^2}{\left(L-|\bm \varOmega|+1\right)4}  = -M\\
\implies M + \log~ |\overline{\mathcal{S}}_K(\epsilon)| & = \frac{2L|\bm \varOmega|t^2}{\left(L-|\bm \varOmega|+1\right)4}\\
\implies t & = \sqrt{\frac{( M + \log~ |\overline{\mathcal{S}}_K(\epsilon)|) (L-|\bm \varOmega|+1)4 }{2L|\bm \varOmega|}  } 
\end{align*}

Therefore, we get that with probability at least $ 1- 2\exp(-M)$, we have
\begin{align*}
\underset{\widetilde{\bm X} \in  \overline{\mathcal{S}}_K(\epsilon)}{\rm sup}\left|\widehat{\mathcal{L}}(\widetilde{\bm X})-{\mathcal{L}}(\widetilde{\bm X})  \right| \le \sqrt{\frac{( M + \log~ |\overline{\mathcal{S}}_K(\epsilon)|) (L-|\bm \varOmega|+1)4 }{2L|\bm \varOmega|}  } .
\end{align*}

By applying \eqref{eq:covering1}, we have
\begin{align}\label{eq:Lepsilonbound}
\underset{\widetilde{\bm X} \in  \overline{\mathcal{S}}_K(\epsilon)}{\rm sup}\left|\widehat{\mathcal{L}}(\widetilde{\bm X})-{\mathcal{L}}(\widetilde{\bm X})  \right| \le  \sqrt{\frac{( M + (2MK+1)K\log(9B^2/\epsilon)) (L-|\bm \varOmega|+1)4 }{2L|\bm \varOmega|}  } : = \zeta .
\end{align}

%
With the above result, we proceed to relate $\mathcal{S}_K$ and $\overline{\mathcal{S}}_K(\epsilon)$. Let $\bm X \in \mathcal{S}_K$ and for every $\bm X$, there exists $\widetilde{\bm X} \in \overline{\mathcal{S}}_K(\epsilon) $ satisfying $\|\bm X- \widetilde{\bm X}\|_{\rm F} \le \epsilon$.
This implies that 
\begin{align*}
|\mathcal{L}(\bm X) - \mathcal{L}(\widetilde{\bm X})| &= \left | \frac{1}{L}\sum_{m<j}\|\widehat{\bm R}_{m,j}-\bm X_{m,j}\|_{\rm F} - \frac{1}{L}\sum_{m<j}\|\widehat{\bm R}_{m,j}-\widetilde{\bm X}_{m,j}\|_{\rm F}\right| \\
&= \left | \frac{1}{L}\sum_{m<j}\left(\|\widehat{\bm R}_{m,j}-\bm X_{m,j}\|_{\rm F} - \|\widehat{\bm R}_{m,j}-\widetilde{\bm X}_{m,j}\|_{\rm F}\right)\right| \\
&\le \left | \frac{1}{L}\sum_{m<j}\left(\|\bm X_{m,j} -\widetilde{\bm X}_{m,j}\|_{\rm F}\right)\right| \le \epsilon
\end{align*}
where we have used the relation that $\|\bm X_{m,j} -\widetilde{\bm X}_{m,j}\|_{\rm F} \le \|\bm X- \widetilde{\bm X}\|_{\rm F} \le \epsilon$.
Similarly, we have
\begin{align*}
|\widehat{\mathcal{L}}(\bm X) - \widehat{\mathcal{L}}(\widetilde{\bm X})| &= \left | \frac{1}{|\bm \varOmega|}\sum_{(m,j) \in \bm \varOmega}\|\widehat{\bm R}_{m,j}-\bm X_{m,j}\|_{\rm F} - \frac{1}{|\bm \varOmega|}\sum_{(m,j) \in \bm \varOmega}\|\widehat{\bm R}_{m,j}-\widetilde{\bm X}_{m,j}\|_{\rm F}\right| \\
&= \left | \frac{1}{|\bm \varOmega|}\sum_{(m,j) \in \bm \varOmega}\left(\|\widehat{\bm R}_{m,j}-\bm X_{m,j}\|_{\rm F} - \|\widehat{\bm R}_{m,j}-\widetilde{\bm X}_{m,j}\|_{\rm F}\right)\right| \\
&\le \left | \frac{1}{|\bm \varOmega|}\sum_{(m,j) \in \bm \varOmega}\left(\|\bm X_{m,j} -\widetilde{\bm X}_{m,j}\|_{\rm F}\right)\right| \le \epsilon.\\
\end{align*}
From the above results, we further have
\begin{align*}
\underset{\bm X \in \mathcal{S}_K}{\rm sup}\left| \widehat{\mathcal{L}}(\bm X)- \mathcal{L}(\bm X)\right| &\le \underset{\bm X \in \mathcal{S}_K}{\rm sup}\left(\left| \widehat{\mathcal{L}}(\bm X)-\widehat{\mathcal{L}}(\widetilde{\bm X})\right| + \left|\mathcal{L}(\widetilde{\bm X})- \mathcal{L}(\bm X)\right| +\left|\widehat{\mathcal{L}}(\widetilde{\bm X})-{\mathcal{L}}(\widetilde{\bm X})\right|\right)\\
&\le \epsilon + \epsilon + \underset{\widetilde{\bm X} \in \overline{\mathcal{S}}_K(\epsilon)}{\rm sup}\left|\widehat{\mathcal{L}}(\widetilde{\bm X})-{\mathcal{L}}(\widetilde{\bm X})\right| \\
&\le 2\epsilon + \zeta,
\end{align*}
where we have applied \eqref{eq:Lepsilonbound} in the last inequality.

Setting $\epsilon = 1/{L}$, we get the below with probability at least $ 1- 2\exp(-M)$:
\begin{align*}
\underset{\bm X \in \mathcal{S}_K}{\rm sup}\left| \widehat{\mathcal{L}}(\bm X)- \mathcal{L}(\bm X)\right| 
&\le 2\frac{1}{{L}} + \sqrt{\frac{( M + (2MK+1)K\log(9LB^2)) (L-|\bm \varOmega|+1)4 }{2L|\bm \varOmega|}  } \\
&\le 2\frac{1}{{L}} + \sqrt{\frac{( M + 3MK^2\log(9LB^2)) (L-|\bm \varOmega|+1)4 }{2L|\bm \varOmega|}  }\\
&\le 2\frac{1}{|\bm \varOmega|} + \sqrt{\frac{2( M + 3MK^2\log(9LB^2)) }{|\bm \varOmega|}  }\\
&\le  2\frac{1}{|\bm \varOmega|} + \sqrt{\frac{2( M + 3MK^2\log(9M^2B^2)) }{|\bm \varOmega|}  }
\end{align*}
 where we have used the relation that $L = M(M-1)/2$ in the last inequality. Note that $B$ is defined such that $\|\bm X\|_{\rm F} \le B^2$, where $\bm X \in \mathcal{S}_K$. In our case, we have ${\bm R}_{m,j} \ge \bm 0,~ \sum_{p,q} {\bm R}_{m,j}(p,q)=1$ and therefore we get $\|{\bm R}_{m,j}\|^2_{\rm F} \le 1$ for all $m,j$. It implies that all the elements $\bm X$ of the feasible set $\mathcal{S}_K$ can be set to have $\|{\bm X}\|_{\rm F}^2 \le M^2$. Therefore, we can set $B^2=M$.

Using the definition of $\tau(\bm \varOmega)$ given by \eqref{eq:tau}, $\widehat{\mathcal{L}}(\bm X)$ and $\mathcal{L}(\bm X)$ given by \eqref{eq:calLs} and setting $B^2=M$, we can then see that, there exists a constant $C>0$ such that 
\begin{align} \label{eq:taubound}
\tau(\bm \varOmega) \le C\sqrt{\frac{MK^2\log(M) }{|\bm \varOmega|}  }.
\end{align}
Substituting \eqref{eq:taubound} in \eqref{eq:rmse}, we get that with probability at least $  1- 2\exp(-M)$,
\begin{align} \label{eq:Rij}
\frac{1}{L}\sum_{m<j}\|{\bm R}^*_{m,j}-\bm R_{m,j}\|_{\rm F} \le\frac{1}{|\bm \varOmega|}\sum_{(m,j) \in \bm \varOmega}\|\bm N_{m,j}\|_{\rm F}  +\frac{1}{L}\sum_{m<j}\|\bm N_{m,j}\|_{\rm F} + C\sqrt{\frac{MK^2\log(M) }{|\bm \varOmega|}  }.
\end{align}
Using Lemma 13 from \cite{zhang2014spectral}, we get that with probability at least $1-\delta$, 
\begin{align} \label{eq:Nij}
\|\bm N_{m,j}\|_{\rm F} \le \frac{1+\sqrt{\log(1/\delta)}}{\sqrt{S_{m,j}}},~\text{if } (m,j) \in \bm \varOmega,
\end{align}
where $S_{m,j}$ is the (nonzero) number of samples that the annotators $m$ and $j$ have co-labeled. Also, without loss of any generality, we can let $\widehat{\bm R}_{m,j} = \bm R_{m,j}$, for all $(m,j) \notin \bm \varOmega$. Therefore, we have
\begin{align}\label{eq:Nij1}
    \|\bm N_{m,j}\|_{\rm F} = 0, ~(m,j) \notin \bm \varOmega.
\end{align}
By substituting $\delta = \exp(-M)$, combining \eqref{eq:Rij}-\eqref{eq:Nij1} with union bound,  we have the below with probability at least  $1-3\exp(-M)$,
\begin{align} \label{eq:Rbound_opt}
\frac{1}{L}\sum_{m<j}\|{\bm R}^*_{m,j}-\bm R_{m,j}\|_{\rm F} \le \left(\frac{1}{|\bm \varOmega|}+\frac{1}{L}\right) \sum_{(m,j) \in \bm \varOmega}\frac{1+\sqrt{M}}{\sqrt{S_{m,j}}}  + C\sqrt{\frac{MK^2\log(M) }{|\bm \varOmega|}  },
\end{align}
where $L = M(M-1)/2$.

{\section{Proof of Theorem \ref{thm:converge}} \label{supp:convergethm}

	We restate the assumptions and the convergence theorem here:
	
	\vspace{.25cm}
	
	\begin{mdframed}[backgroundcolor=gray!10,topline=false,
		rightline=false,
		leftline=false,
		bottomline=false]
		\noindent
		{\bf Assumption 1}
		{
			{\it
		The nonnegative factor $\H \in\mathbb{R}^{MK\times K}_+$ satisfies:
(i) ${\rank}(\H)=K$ and {$\|\H\|_{\rm F}  = \sigma$};  (ii) $\frac{ \|\H(j,:) \bm \Theta\|_2^2 }{ \| \H\bm \Theta \|_{\rm F}^2 } \leq \zeta,~\forall j,~\forall \bm \Theta\in\mathbb{R}^{K\times K} $; (iii) the locations of the nonzero elements {of $\H$ are uniformly distributed over $[MK]\times[K]$, and the set $\bm \varDelta = \{ (j,k) : [\bm H]_{j,k} > 0\} $}
has the following cardinality bound
\begin{equation}\label{eq:supp_density}
 |\bm \varDelta|  = O\left(\frac{MK\gamma_{0}^2}{(1+MK\zeta ) \sigma^4}\right);   
\end{equation}
and (iv) $0<\gamma_0 \leq \min_{1\leq k\leq K}\{  \beta^2_k -\beta^2_{k+1} \}$, where $\beta_k$ is the $k$th singular value of $\H$ and $\beta_{K+1}=0$.
				
			}
			
		}
	\end{mdframed}

	\begin{mdframed}[backgroundcolor=gray!10,topline=false,
		rightline=false,
		leftline=false,
		bottomline=false]
		\noindent
		{\bf Theorem 3}
		Under Assumption \ref{as:H},
consider $\widehat{\U}=\H\Q^\T+\bm N$, where $\Q\in\mathbb{R}^{K\times K}$ is orthogonal, and apply \eqref{eq:algo_updates}. Denote $\nu=\|\bm N\|_{\rm F}$, $h_{(t)}=	\|\bm H_{(t)}-\bm H \bm \varPi\|_{\rm F}^2$ and $q_{(t)}=	\|\bm Q_{(t)}-\bm Q \bm \varPi\|_{\rm F}^2$, where $\bm \varPi$ is any permutation matrix. Suppose that $\nu \le \sigma\min\{(1-\rho)\sqrt{\eta} q_{(0)},1\}$ for {$\rho := O(\nicefrac{K\eta \sigma^4}{\gamma_0^2})\in(0,1)$}, where $\eta = (\nicefrac{|\bm \varDelta|}{MK^2})(1+MK\zeta )$, and that
\begin{equation}\label{eq:supp_initcond}
    2\sigma q_{(0)} + 2\nu<\min_{(j,k) \in \bm \varDelta} [\bm H]_{j,k} .
\end{equation}  
Then, there exists $\alpha_{(t)}=\alpha > 0$ 
such that with probability of at least $1-\delta$ the following holds:
\begin{subequations}
\begin{align}
  q_{(t)} &\leq \rho q_{(t-1)} + O\left( \nicefrac{  K \sigma^2\nu^2}{\gamma_0^2}\right), \label{eq:supp_Qerror}\\
  h_{(t)} &\leq 2\eta \sigma^2 q_{(t-1)} + 2\nu^2,  \label{eq:supp_Herror}
      \end{align}
      \end{subequations}
 where $\delta = 2\exp\left(   - \nicefrac{ 2|\bm \varDelta|     }{K^2\left(1-\frac{|\bm \varDelta|-1}{MK^2}\right)  }\right) $. 
	\end{mdframed}
	
	\noindent

	Let $\widehat{\X}$ be the estimated $\X$ in \eqref{eq:Xbig}.
	Consider the rank-$K$ square root decomposition of $\widehat{\X} \in \mathbb{R}^{MK \times MK}$:  $$\widehat{\X}=\widehat{\U}\widehat{\U}^\T.$$
	It can be shown that $\widehat{\bm U} = \bm U + \bm N = \bm H \bm Q^{\top} +\bm N$ with bounded noise $\bm N$, if $\widehat{\X}$ is a reasonable estimate for $\X$ (cf. Lemma~\ref{lem:svds1}).
	
	Using $\widehat{\U}\in\mathbb{R}^{MK\times K}$, the proposed SymNMF algorithm has the following updates:
	\begin{subequations}
		\begin{align}
		\H_{(t+1)}& \leftarrow {\sf ReLU}_{\alpha_{(t)}}\left( \widehat{\bm U} \bm Q_{(t)} \right) \label{eq:Htsupp}\\
		{\bm W}_{(t+1)}{\bm \Sigma}_{(t+1)}{\bm V}_{(t+1)}^\T &\leftarrow {\sf svd}\left(  \bm H_{(t+1)}^\T{\widehat{\bm U}} \right)\label{eq:WVtsupp}\\
		\Q_{(t+1)}&\leftarrow  {\bm V}_{(t+1)}{\bm W}_{(t+1)}^\T\label{eq:Qtsupp},
		\end{align}
	\end{subequations}
	where $\alpha_{(t)} >0$. 
	
	In the proof, we omit the permutation notation $\bm \varPi$ for notation simplicity, since all the column-permuted version of $\H$ and $\Q$ are considered equally good---i.e., the column permutation ambiguity in NMF problems is intrinsic; see \cite{fu2018nonnegative,huang2014non}.
	
	Suppose that $\bm  Q^{\top}\bm Q_{(t)} = \bm I +   \bm E_{\bm Q_{(t)}}$. Note that
	\[ \| \bm E_{\bm Q_{(t)}}\|_{\rm F}  = \| \bm  Q -\bm Q_{(t)}  \|_{\rm F} \]
	per the orthogonality of $\Q$ and $\Q_{(t)}$.
	
	{
	
	Also define $$\bm E_{\bm H_{(t+1)}} := \H_{(t+1)} - \bm H.$$

	\subsection{The $\bm H$-update}
	From the update in \eqref{eq:Htsupp}, the below set of relations can be obtained:
	\begin{align}
||\bm E_{\bm H_{(t+1)}} \|_{\rm F}  &= \left\lVert{\sf ReLU}_{\alpha_{(t)}}\left( \widehat{\bm U} \bm Q_{(t)} \right) - \bm H\right\rVert_{\rm F} \nonumber\\
&= \left\lVert{\sf ReLU}_{\alpha_{(t)}}\left(  (\bm H \bm Q^{\top} +\bm N)\bm Q_{(t)} \right) - \bm H\right\rVert_{\rm F} \nonumber\\
&= \left\lVert{\sf ReLU}_{\alpha_{(t)}}\left(  {\bm H}(\bm Q^{\top} \bm Q_{(t)}) + \bm N\bm Q_{(t)} \right) - \bm H\right\rVert_{\rm F} \nonumber\\
&= \left\lVert{\sf ReLU}_{\alpha_{(t)}}\left(  {\bm H}(\bm I +   \bm E_{\bm Q_{(t)}}) +\bm N\bm Q_{(t)} \right) - \bm H\right\rVert_{\rm F} \nonumber\\
&= \left\lVert{\sf ReLU}_{\alpha_{(t)}}\left(  {\bm H} +   \bm H\bm E_{\bm Q_{(t)}}+\bm N\bm Q_{(t)}\right) - \bm H\right\rVert_{\rm F}. \label{eq:EHt}
\end{align}
Recall that $\bm \varDelta : = \{(j,k) : [\bm H]_{j,k} > 0\}$. Assume that the following conditions are satisfied for $\alpha_{(t)}$ (at the end of the proof, using Lemma \ref{lem:alpha_feas}, we will establish the feasibility of $\alpha_{(t)}$ satisfying the below conditions),
	\begin{align}
&\alpha_{(t)} \leq[\bm H+\bm H\bm E_{\bm Q_{(t)}}+\bm N\bm Q_{(t)}]_{j,k} ,\quad \forall (j,k) \in \bm \varDelta , \label{eq:cond_1}\\
&	{\alpha_{(t)} \geq [\bm H\bm E_{\bm Q_{(t)}}+\bm N\bm Q_{(t)}]_{j,k},\quad \forall j,k. } \label{eq:newcond}
	\end{align}
}

Then, we have
		\begin{align}
&\|\bm E_{\bm H_{(t+1)}} \|^2_{\rm F} \nonumber \\ &=\sum_{(j,k) \in \bm \varDelta}\left|[{\sf ReLU}_{\alpha_{(t)}}\left(  {\bm H} +   \bm H\bm E_{\bm Q_{(t)}}+\bm N\bm Q_{(t)}\right)]_{j,k} - [\bm H]_{j,k}\right|^2   + \sum_{(j,k) \notin \bm \varDelta}\left|[{\sf ReLU}_{\alpha_{(t)}}\left(  {\bm H} +   \bm H\bm E_{\bm Q_{(t)}}+\bm N\bm Q_{(t)}\right)]_{j,k} - [\bm H]_{j,k}\right|^2  \nonumber\\
&= \sum_{(j,k) \in \bm \varDelta}\left|[{\sf ReLU}_{\alpha_{(t)}}\left(  {\bm H} +   \bm H\bm E_{\bm Q_{(t)}}+\bm N\bm Q_{(t)}\right)]_{j,k} - [\bm H]_{j,k}\right|^2   +\sum_{(j,k) \notin \bm \varDelta}  \left|[{\sf ReLU}_{\alpha_{(t)}}\left(   \bm H\bm E_{\bm Q_{(t)}}+\bm N\bm Q_{(t)}\right)]_{j,k} \right|^2 \nonumber\\
&= \sum_{(j,k) \in \bm \varDelta}\left|[   \bm H\bm E_{\bm Q_{(t)}}+\bm N\bm Q_{(t)}]_{j,k}\right|^2, \label{eq:exp_EH_bound1}
\end{align}
where we used $[\bm H]_{j,k} =0, \forall (j,k) \notin \bm \varDelta$ to get the second equality and applied the conditions in \eqref{eq:cond_1} and \eqref{eq:newcond} to obtain the last equality.

Note that the below holds:
\begin{align}
\left|[   \bm H\bm E_{\bm Q_{(t)}}+\bm N\bm Q_{(t)}]_{j,k} \right|^2 &= |[\bm H\bm E_{\bm Q_{(t)}}]_{j,k}|^2+|[\bm N\bm Q_{(t)}]_{j,k}|^2 + 2[\bm H\bm E_{\bm Q_{(t)}}]_{j,k}[\bm N\bm Q_{(t)}]_{j,k} \nonumber\\
&\le |[\bm H\bm E_{\bm Q_{(t)}}]_{j,k}|^2+|[\bm N\bm Q_{(t)}]_{j,k}|^2 + |[\bm H\bm E_{\bm Q_{(t)}}|^2+|[\bm N\bm Q_{(t)}]_{j,k}|^2 \nonumber\\
&= 2|[\bm H\bm E_{\bm Q_{(t)}}]_{j,k}|^2+2|[\bm N\bm Q_{(t)}]_{j,k}|^2, \label{eq:HEQ_NQ}
\end{align}
where we have applied the Young's inequality in the first inequality.

Combining \eqref{eq:exp_EH_bound1} and \eqref{eq:HEQ_NQ}, we get that
\begin{align}
\|\bm E_{\bm H_{(t+1)}} \|^2_{\rm F} \le 2\sum_{(j,k) \in \bm \varDelta}\left|[   \bm H\bm E_{\bm Q_{(t)}}]_{j,k}\right|^2 + 2\sum_{(j,k) \in \bm \varDelta}\left|[   \bm N\bm Q_{(t)}]_{j,k}\right|^2. \label{eq:exp_EH_bound2}
\end{align}
Next, we consider the following lemma to bound the first term in \eqref{eq:exp_EH_bound2}.
\begin{lemma} \cite{serfling1974probability}\label{lem:hoeffding}
	Let $X =[X_1,\dots, X_n]$ be a set of samples taken without replacement from a set $\{x_1,\dots,x_N\}$ with mean $u$ where $n \le N$. Denote $a := \min_i x_i$ and $b := \max_i x_i$. Then, we have
	\begin{align*}
	{\sf Pr}\left(\left|\frac{1}{n}\sum_{i=1}^n X_i- u\right| \ge s \right) \le 2 \exp\left(-\frac{2ns^2}{\left(1-\frac{n-1}{N}\right)(b-a)^2} \right).
	\end{align*}
\end{lemma}

Applying Lemma \ref{lem:hoeffding}, and by the assumption that nonzero elements of $\bm H$ are located over $[MK]\times [K]$ uniformly at random, we get
\begin{align*}
{\sf Pr}\left(  \frac{1}{|\bm \varDelta|}\sum_{(j,k)\in \bm \varDelta } | [\bm H\bm E_{\bm Q_{(t)}}]_{j,k} |^2  - \frac{1}{JK} \|\bm H\bm E_{\bm Q_{(t)}}\|_{\rm F}^2 \geq s  \right) \leq 2\exp\left(   - \frac{ 2|\bm \varDelta| s^2     }{ (1-\frac{|\bm \varDelta|-1}{JK}) (b-a)^2 }  \right),
\end{align*}
where $J = MK$.
Using the assumption that $\frac{ \|\H(j,:) \bm \Theta\|_2^2 }{ \| \H\bm \Theta \|_{\rm F}^2 } \leq \zeta,~\forall j,~\forall \bm \Theta\in\mathbb{R}^{K\times K} $,  we get
\begin{subequations} \label{eq:bound_a_b}
\begin{align}
b &= \max_{j,k}~|[\bm H\bm E_{\bm Q_{(t)}}]_{j,k}|^2 \le \max_{j}~\|\bm H(j,:)\bm E_{\bm Q_{(t)}}\|_2^2 \le \zeta \|\bm H\bm E_{\bm Q_{(t)}}\|_{\rm F}^2\\
a &= \min_{j,k}~|[\bm H\bm E_{\bm Q_{(t)}}]_{j,k}|^2 \ge 0.
\end{align}
\end{subequations}
Using the bounds \eqref{eq:bound_a_b} and by letting $s = \zeta \|\bm H\bm E_{\bm Q_{(t)}}\|_{\rm F}^2/K$, we get that 
\begin{align*}
{\sf Pr}\left(  \frac{1}{|\bm \varDelta|}\sum_{i,k\in \bm \varDelta } | [\bm H\bm E_{\bm Q_{(t)}}]_{j,k} |^2  - \frac{1}{JK} \|\bm H\bm E_{\bm Q_{(t)}}\|_{\rm F}^2 \geq \frac{1}{K} \zeta \|\bm H\bm E_{\bm Q_{(t)}}\|_{\rm F}^2  \right) \leq 2\exp\left(   - \frac{ 2|\bm \varDelta|    }{ K^2(1-\frac{|\bm \varDelta|-1}{JK})  }  \right).
\end{align*}
It implies that with probability at least $1-2\exp\left(   - \frac{ 2|\bm \varDelta|     }{K^2 (1-\frac{|\bm \varDelta|-1}{JK})  }\right)$, we get
\begin{align}
\sum_{(j,k)\in \bm \varDelta } | [\bm H\bm E_{\bm Q_{(t)}}]_{j,k} |^2 \le \frac{|\bm \varDelta|}{JK}(1+J\zeta )\|\bm H\bm E_{\bm Q_{(t)}}\|_{\rm F}^2  \le \frac{|\bm \varDelta|}{JK}(1+J\zeta )\|\bm H\|_{\rm F}^2\|\bm E_{\bm Q_{(t)}}\|_{\rm F}^2. \label{eq:exp_EH_bound3}
\end{align}

Letting $\eta =\frac{|\bm \varDelta|}{JK}(1+J\zeta )$ and applying \eqref{eq:exp_EH_bound3} in \eqref{eq:exp_EH_bound2}, we get that
\begin{align}
\|\bm E_{\bm H_{(t+1)}} \|^2_{\rm F} 
&\le 2\eta\|\bm H\|_{\rm F}^2 \|\bm E_{\bm Q_{(t)}}\|_{\rm F}^2 + 2\sum_{(j,k) \in \bm \varDelta}\left|[   \bm N\bm Q_{(t)}]_{j,k}\right|^2 \nonumber\\
&\le 2\eta\|\bm H\|_{\rm F}^2 \|\bm E_{\bm Q_{(t)}}\|_{\rm F}^2 + 2\|  \bm N\bm Q_{(t)}\|_{\rm F}^2 \nonumber\\
&=  2\eta\|\bm H\|_{\rm F}^2 \|\bm E_{\bm Q_{(t)}}\|_{\rm F}^2 + 2\nu^2,\label{eq:exp_EH_bound4}
\end{align}
where we have used $\|\bm N\|_{\rm F} = \nu$ and the orthogonality of $\bm Q_{(t)}$ in the last equality.

	\subsection{The $\Q$-update}
	
	We will now consider the update in \eqref{eq:WVtsupp}:
	\begin{align*}
	\bm H_{(t+1)}^\T\widehat{\bm U} & = (\bm H + {\bm E}_{\bm H_{(t+1)}})^{\top}({\bm U}+\bm N)
	\\&= \bm H^{\top} \bm U +  {\bm E}^{\top}_{\bm H_{(t+1)}}\bm U +\bm H^{\top}\bm N + {\bm E}^{\top}_{\bm H_{(t+1)}}\bm N .
	\end{align*}
	We bound the below:
	\begin{align}
\|\bm H_{(t+1)}^\T\widehat{\bm U}-\bm H^{\top} \bm U\|^2_{\rm F} &= \|{\bm E}^{\top}_{\bm H_{(t+1)}}\bm U + \H^\T\bm N +  {\bm E}^{\top}_{\bm H_{(t+1)}}\bm N   \|^2_{\rm F} \nonumber\\
	& \leq 3 \|{\bm E}^{\top}_{\bm H_{(t+1)}}\bm U \|_{\rm F}^2   +  3 \| \H^\T\bm N  \|_{\rm F}^2  + 3 \| {\bm E}^{\top}_{\bm H_{(t+1)}}\bm N   \|^2_{\rm F}   \nonumber\\
	&\leq 3\|\bm H\|_{\rm F}^2 \| \bm E_{\H_{(t+1)}} \|_{\rm F}^2 + 3 \|\bm H\|_{\rm F}^2 \nu^2 + 3 \nu^2  \| \bm E_{\H_{(t+1)}} \|_{\rm F}^2\nonumber\\
	&= 3(\|\bm H\|_{\rm F}^2 +\nu^2) \| \bm E_{\H_{(t+1)}} \|_{\rm F}^2 + 3 \|\bm H\|_{\rm F}^2\nu^2\nonumber\\
	&\leq  3(\|\bm H\|_{\rm F}^2 +\nu^2) \left( 2\eta\|\bm H\|_{\rm F}^2 \| {\bm E}_{\bm Q_{(t)}}\|^2_{\rm F} + 2\nu^2\right) +  3 \|\bm H\|_{\rm F}^2\nu^2\nonumber\\
	& =  6\eta(\|\bm H\|_{\rm F}^2 +\nu^2)  \|\bm H\|_{\rm F}^2 \| {\bm E}_{\bm Q_{(t)}}\|^2_{\rm F} + 6 (\|\bm H\|_{\rm F}^2 +\nu^2) \nu^2  +   3 \|\bm H\|_{\rm F}^2\nu^2 \nonumber\\
	& \le 12\eta\|\bm H\|_{\rm F}^4   \| {\bm E}_{\bm Q_{(t)}}\|^2_{\rm F} + 15 \|\bm H\|_{\rm F}^2\nu^2 , 
\label{eq:shrinking}
	\end{align}
	where we have used {the Young's inequality for the first inequality}, used the fact that $\|\bm U\|_{\rm F}=\|{\bm H}\|_{\rm F}$ for the second inequality, applied the result in \eqref{eq:exp_EH_bound4} for the third inequality and used the assumption that $\| \bm N \|_{\rm F} = \nu \le \|\bm H\|_{\rm F}$ for the last inequality.
	
	Let us proceed to characterize the SVD operation in \eqref{eq:WVtsupp}. Denote the full SVD of $\bm H^{\top} \bm U$ using the following notation:
	\begin{align*}
	{\bm W}{\bm \Sigma}{\bm V}^\T &= {\sf svd}\left(  \bm H^\T{\bm U} \right).
	\end{align*}
	We invoke the below lemma:
	
	\begin{lemma} \cite{wedin1972perturbation,mirsky1960quater,fan2018eigenvector}\label{lem:svds2}
		Let $\bm C \in \mathbb{R}^{m \times n}$ and $\widehat{\bm C} \in \mathbb{R}^{m \times n}$ have singular values $\sigma_1 \ge \sigma_2 \ge \dots \ge \sigma_{\min(m,n)}$ and $\widehat{\sigma}_1 \ge \widehat{\sigma}_2 \ge \dots \widehat{\sigma}_{\min(m,n)}$, respectively. {Let $r \le \min\{m,n\}$}. Denote $\bm w_{1} ,\ldots, \bm w_{r}  \in \mathbb{R}^{m }$ and  $\widehat{\bm w}_1, \ldots,  \widehat{\bm w}_{r}  \in \mathbb{R}^{m}$ as the orthonormal columns satisfying $\bm C^{\top} \bm w_i = \sigma_i \bm v_i$ and $\widehat{\bm C}^{\top} \widehat{\bm w}_i = \widehat{\sigma}_i \widehat{\bm v}_i$ for $i = 1,\dots,r$ and let $\bm v_1, \ldots, \bm v_{r}  \in \mathbb{R}^{n}$ and  $\widehat{\bm v}_1 , \ldots, \widehat{\bm v}_{r} \in \mathbb{R}^{n}$ are orthonormal columns satisfying $\bm C \bm v_i = \sigma_i \bm w_i$ and $\widehat{\bm C} \widehat{\bm v}_i = \widehat{\sigma}_i \widehat{\bm w}_i$ for $i = 1,\dots,r$. Denote ${\gamma}_0 = \min \{\sigma_i-\sigma_{i+1}: i=1,\dots,r\}$ where $\sigma_{r+1}=0$. Then, if $\|\widehat{\bm C}-\bm C\|_2 \le {\gamma}_0/2$, we have
		\begin{align} \label{eq:svd_bound_wedin}
		\underset{1\le i \le r}{\max}\{\|\widehat{\bm w}_i-\bm w_i\|_2 \vee \|\widehat{\bm v}_i-\bm v_i\|_2\}
		&\le \frac{2\sqrt{2}\|\widehat{\bm C}-\bm C\|_{2}}{{\gamma}_0},
		\end{align}
		where the operation $a \vee b = \max\{a,b\}$.
	\end{lemma}
	A short proof of how the bound in \eqref{eq:svd_bound_wedin} is obtained from the classic result in \cite{wedin1972perturbation} is given in Section \ref{supp:lem_svds2}.
	
	By letting $\bm C := \bm H^{\top}\bm U$, $\widehat{\bm C} :=\bm H_{(t+1)}^\T\widehat{\bm U}$ and applying Lemma~\ref{lem:svds2}, we have
	\begin{align}
	\|{\bm W}_{(t+1)}-{\bm W}\|_{\rm F} &\le \frac{2\sqrt{2K}\|\bm H_{(t+1)}^\T\widehat{\bm U}-\bm H^{\top}\bm U\|_{\rm F}}{\gamma_0},\label{eq:Wbound_1}\\
	\|{\bm V}_{(t+1)}-{\bm V}\|_{\rm F} &\le \frac{2\sqrt{2K}\|\bm H_{(t+1)}^\T\widehat{\bm U}-\bm H^{\top}\bm U\|_{\rm F}}{\gamma_0}, \label{eq:Vbound_1}
	\end{align}
	where we have used the fact that for any matrix $\bm \Theta = [\bm \theta_1,\dots,\bm \theta_K]$, the equality $\|\bm \Theta\|_{\rm F} = \sqrt{ \sum_{i=1}^K \|\bm \theta_i\|_2^2}$ holds. We have also applied matrix norm equivalence $\|\bm \Theta\|_2 \le \|\bm \Theta\|_{\rm F}$.
	Note that since the singular values of $\bm H^{\top}\bm U$ are the same as that of $\bm H^{\top} \bm H$, we re-define $\gamma_0$ as
	\[  \gamma_0 =  \min_{1\leq k\leq K}\{\beta^2_{k}-\beta^2_{k+1}\}, \]
	where $\beta_k$'s are the singular values of $\bm H$.


	By squaring the term in the right hand side of \eqref{eq:Wbound_1}, we get
	\begin{align}
	\|{\bm W}_{(t+1)}-{\bm W}\|^2_{\rm F} &\le \frac{8K\|\bm H_{(t+1)}^\T\widehat{\bm U}-\bm H^{\top}\bm U\|^2_{\rm F}}{\gamma_0^2}\nonumber\\
	& \le \frac{8K  \left(12\eta\|\bm H\|_{\rm F}^4   \| {\bm E}_{\bm Q_{(t)}}\|^2_{\rm F} + 15 \|\bm H\|_{\rm F}^2\nu^2\right)}{\gamma_0^2},\label{eq:Wbound1}
	\end{align}
	where we applied \eqref{eq:shrinking} to obtain the last inequality.
	We can similarly get that
	\begin{align} \label{eq:Vbound1}
\|{\bm V}_{(t+1)}-{\bm V}\|^2_{\rm F} &\le \frac{8K  \left(12\eta\|\bm H\|_{\rm F}^4   \| {\bm E}_{\bm Q_{(t)}}\|^2_{\rm F} + 15 \|\bm H\|_{\rm F}^2\nu^2\right)}{\gamma_0^2}.
	\end{align}
	
	

	Consider $\bm E_{\bm Q_{(t+1)} }= \bm Q_{(t+1)}-\bm Q $. Then,
	\begin{align}
	\|\bm E_{\bm Q_{(t+1)}}\|_{\rm F}^2 = \|\bm Q_{(t+1)}-\bm Q\|_{\rm F}^2 &= \|\bm V_{(t+1)}\bm W_{(t+1)}^\T- \bm V\bm W^\T\|_{\rm F}^2 \nonumber\\
	&= \|\bm V_{(t+1)}(\bm W_{(t+1)}^\T- \bm W^\T)+(\bm V_{(t+1)}-\bm V)\bm W^\T\|_{\rm F}^2 \nonumber\\
	&\le 2\|\bm W_{(t+1)}- \bm W\|_{\rm F}^2 + 2\|\bm V_{(t+1)}-\bm V\|_{\rm F}^2, \nonumber
	\end{align}
	{where the last inequality is by the Young's inequality} and the fact that $\|\bm \Theta\bm \Phi\|_{\rm F}^2 \leq \|\bm \Theta\|_2^2\| \bm \Phi \|_{\rm F}^2$ for two matrices $\bm \Theta$ and $\bm \Phi$; we have also used that $\|\W\|_2=\|\bm V_{(t+1)}\|_2=1$.
	The above leads to
	\begin{equation}
	\|\bm E_{\bm Q_{(t+1)}}\|^2_{\rm F} \leq \frac{CK  \left(\eta\|\bm H\|_{\rm F}^4   \| {\bm E}_{\bm Q_{(t)}}\|^2_{\rm F} +  \|\bm H\|_{\rm F}^2\nu^2 \right)}{\gamma_0^2}.\label{eq:EQ_expbound0}
	\end{equation}
	for a certain constant $C>1$.
	Let us denote $\rho := \frac{CK\eta \|\bm H\|_{\rm F}^4}{\gamma_0^2}$. Then we have
	\begin{align}
		\|\bm E_{\bm Q_{(t+1)}}\|^2_{\rm F} \leq \rho   \| {\bm E}_{\bm Q_{(t)}}\|^2_{\rm F} + \frac{\rho \nu^2}{\eta\|\bm H\|_{\rm F}^2 }.\label{eq:EQ_expbound}
	\end{align}
	
	 We can see that if the below condition is satisfied, then $\rho  < 1$:
	\begin{align}
	\eta = \frac{|\bm \varDelta|}{JK}(1+J\zeta ) &\le \frac{\gamma_{0}^2}{CK \|\bm H\|_{\rm F}^4}, \nonumber\\
	\implies |\bm \varDelta| & \le \frac{J\gamma_{0}^2}{C(1+J\zeta ) \|\bm H\|_{\rm F}^4}. \label{eq:cond_omega}
	\end{align}

	{
	Therefore, under the conditions of $\alpha_{(t)}$ in \eqref{eq:cond_1} and \eqref{eq:newcond} and the condition on $|\bm \varDelta|$ in \eqref{eq:cond_omega}, we get the bound for $\|\bm E_{\bm Q_{(t+1)}}\|^2_{\rm F}$ and $\|\bm E_{\bm H_{(t+1)}} \|^2_{\rm F}$ given by \eqref{eq:EQ_expbound} and \eqref{eq:exp_EH_bound4}, respectively, with $\rho <1$ and with probability greater than $1-2\exp\left(   - \frac{ 2|\bm \varDelta|     }{K^2 (1-\frac{|\bm \varDelta|-1}{JK})  }\right)$.


Regarding the feasibility of $\alpha_{(t)}$ satisfying the conditions \eqref{eq:cond_1} and \eqref{eq:newcond}, we have the following lemma:
\begin{lemma} \label{lem:alpha_feas}
Assume that the following conditions are satisfied:
\begin{align*}
    \nu &\le (1-\rho)\sqrt{\eta} \|\bm H\|_{\rm F}\|\bm E_{\bm Q_{(0)}}\|_{\rm F},\quad \quad
    \min_{(j,k) \in \bm \varDelta} [\bm H]_{j,k} > 2\|\bm H\|_{\rm F}\|\bm E_{\bm Q_{(0)}}\|_{\rm F} + 2\nu.
\end{align*}
Then there exists $\alpha_{(t)}=\alpha > 0$, for all $t$, specified as below such that the bounds given by \eqref{eq:EQ_expbound} and \eqref{eq:exp_EH_bound4} hold true:
\begin{align*}
   \|\bm H\|_{\rm F}\|\bm E_{\bm Q_{(0)}}\|_{\rm F} + \nu\leq \alpha \leq \min_{(j,k) \in \bm \varDelta} [\bm H]_{j,k}-\|\bm H\|_{\rm F}\|\bm E_{\bm Q_{(0)}}\|_{\rm F} - \nu. 
\end{align*}
\end{lemma}

The proof can be found in Sec. \ref{app:alpha_feas}.

}
}

\section{Proof of Lemma \ref{lem:inverse}} \label{supp:lem_inverse}
Consider the below:
\begin{align}
    \|(\bm Y+ \bm E)^{-1} - \bm Y^{-1}\|_{2}  &= \|(\bm Y+ \bm E)^{-1} (\bm I- (\bm Y+\bm E)\bm Y^{-1}\|_2 \nonumber\\
    &=\|(\bm Y+ \bm E)^{-1}\bm E \bm Y^{-1}\|_2 \nonumber\\
    &\le \frac{\|\bm E \|_2}{\sigma_{\min}(\bm Y)\sigma_{\min}(\bm Y+\bm E)}. \label{eq:inverseYE}
\end{align}
Next, we consider the following relations for any vector $\bm x \in \mathbb{R}^{K }$ satisfying $\|\bm x\|=1$:
	\begin{align*}
	\|(\bm Y+\bm E)\bm x\|_2 &= \|\bm Y\bm x+\bm E\bm x \|_2\\
	&\ge  \|\bm Y\bm x\|_2- \|\bm E\bm x\|_2,\\
	\implies \underset{\bm x}{\min}~\|(\bm Y+\bm E)\bm x\|_2 &\ge \underset{\bm x}{\min}~ \|\bm Y\bm x\|_2 - \underset{\bm x}{\max}~\|\bm E\bm x\|_2,\\
	\implies \sigma_{\min}(\bm Y+\bm E) &\ge \sigma_{\min}(\bm Y) - \|\bm E|_2,
	\end{align*}
	where the first inequality is by applying the triangle inequality. Using the assumption that $\|\bm E\|_2 \le \sigma_{\min}(\bm Y)/2$, we get $\sigma_{\min}(\bm Y+\bm E) \ge \sigma_{\min}(\bm Y)/2$. Applying this relation in \eqref{eq:inverseYE}, we get the bound in the lemma.
	
\section{Proof of Lemma \ref{lem:sigmaU}}\label{supp:lem_sigmaU}
	Recall the below relation:
	\begin{align} \label{eq:Cr_r1}
	\bm C &=[\R_{m,r}^{\top}, \R_{\ell,r}^{\top}]^{\top}
	= [\bm A_m^{\top},\bm A_{\ell}^{\top}]^{\top}\bm D \bm A_{r}. 
	\end{align}
	
	The SVD of $\bm C$ results the below:
	\begin{align} \label{eq:Cr_r2}
	\bm C =  [\bm U_m^{\top} ,\bm U_{\ell}^{\top}]^{\top}\bm \Sigma_{m,\ell,r}\bm V_{r}^{\top}
	\end{align}
	
	From \eqref{eq:Cr_r1} and \eqref{eq:Cr_r2}, we get that there exists a nonsingular matrix ${\bm \Theta} \in \mathbb{R}^{K \times K}$ such that
	\begin{align} \label{eq:UGM}
	[\bm U_m^{\top} ,\bm U_{\ell}^{\top}]^{\top} =  [\bm A_m^{\top},\bm A_{\ell}^{\top}]^{\top} \bm \Theta,
	\end{align}
	where the matrix $[\bm U_m^{\top} ,\bm U_{\ell}^{\top}]^{\top}$ is semi-orthogonal. Therefore,  we get
	\begin{align} \label{eq:GMM}
	\sigma_{\max}({\bm \Theta}) = \frac{1}{\sigma_{\min}([\bm A_m^{\top},\bm A_{\ell}^{\top}]^{\top})} \quad \text{and}\quad  \sigma_{\min}({\bm \Theta}) = \frac{1}{\sigma_{\max}([\bm A_m^{\top},\bm A_{\ell}^{\top}]^{\top})}.
	\end{align}
	Since $\bm A_m$ is full row-rank, we have
	\begin{align}
	\sigma_{\min}({\bm U}_{m})  &= \underset{\|\bm x\|_2=1}{\min}~\| \bm A_{m}{\bm \Theta}\bm x\|_2 \nonumber\\
	&\ge  \underset{\|\bm x\|_2=1}{\min}~\sigma_{\min}(\bm A_{m})\|{\bm \Theta}\bm x\|_2 = \sigma_{\min}(\bm A_{m})\underset{\|\bm x\|_2=1}{\min}~\|{\bm \Theta}\bm x\|_2 \nonumber\\
	&=\sigma_{\min}(\bm A_{m}) \sigma_{\min}({\bm \Theta}) = \frac{\sigma_{\min}(\bm A_{m})}{\sigma_{\max}([\bm A_{m}^{\top},\bm A_{\ell}^{\top}]^{\top})}. \label{eq:Usigmamin}
	\end{align}
	where we have applied \eqref{eq:GMM} to obtain the last equality.
	
	We proceed to bound $\sigma_{\max}([\bm A_{m}^{\top},\bm A_{\ell}^{\top}]^{\top})$. Under the assumption $\kappa(\bm A_m) \le \gamma$, for all $m$, there exists a positive scalar $\omega_{\max}$ and $\omega_{\min}$, such that for all $m$,
	$$\sigma_{\max}(\bm A_m) \le \omega_{\max}, \quad \sigma_{\min}(\bm A_m) \ge \omega_{\min}, \quad \gamma:= \frac{\omega_{\max}}{\omega_{\min}}.$$ Then we have,
	\begin{align*}
	\sigma^2_{\max}([\bm A_m^{\top},\bm A_{\ell}^{\top}]^{\top})&= \|[\bm A_m^{\top},\bm A_{\ell}^{\top}]^{\top}\|_2^2
	\le \|[\bm A_m^{\top},\bm A_{\ell}^{\top}]^{\top}\|_{\rm F}^2 \\
	& = \|\bm A_m\|_{\rm F}^2 +\|\bm A_{\ell}\|_{\rm F}^2
	\le K \|\bm A_m\|_{2}^2 + K \|\bm A_{\ell}\|_{2}^2 \le 2K \omega_{\max}^2,
	\end{align*}
	where we have utilized the norm equivalence for the first and second inequalities. Hence, we have
	\begin{align*}
	\sigma_{\max}([\bm A_m^{\top},\bm A_{\ell}^{\top}]^{\top}) &\le \sqrt{2K}\omega_{\max}.
	\end{align*}
Applying the above results in \eqref{eq:Usigmamin}, we get
\begin{align*}
    \sigma_{\min}({\bm U}_{m}) \ge \frac{\omega_{min}}{\sqrt{2K}\omega_{\max}} = \frac{1}{\sqrt{2K}\gamma}.
\end{align*}
Similarly, we can easily show the above lower bound for $\sigma_{\min}({\bm U}_{\ell})$.

Next, we consider upper bounding $\sigma_{\max}({\bm U}_{m})$ and $\sigma_{\max}({\bm U}_{\ell})$. From \eqref{eq:UGM} and \eqref{eq:GMM}, we have
\begin{align*}
\sigma_{\max}({\bm U}_{m}) &\le  \sigma_{\max}({\bm \Theta}) \sigma_{\max}(\bm A_{m}) =\frac{\sigma_{\max}(\bm A_{m})}{\sigma_{\min}([\bm A_m^{\top},\bm A_{\ell}^{\top}]^{\top})} \\
&\le\frac{\sigma_{\max}(\bm A_{m})}{\sigma_{\min}(\bm A_m)}
	\le \frac{\omega_{\max}}{\omega_{\min}} = \gamma,
\end{align*}
where we have applied $\sigma_{\min}([\bm A_m^{\top},\bm A_{\ell}^{\top}]^{\top}) \ge \sigma_{\min}(\bm A_m)$ for second inequality.
Similarly, we can easily show the above upper bound for $\sigma_{\max}({\bm U}_{\ell})$.

\section{Proof of Lemma~\ref{lem:svds2}} \label{supp:lem_svds2}
The perturbation theorem in \cite{wedin1972perturbation} gives the below bound if $\|\widehat{\bm C}-\bm C\|_2 \le \widetilde{\gamma}_0/2$,
	\begin{align} \label{eq:wedin_svdresult}
	\sqrt{\sum_{i=1}^r(\sin^2\theta(\widehat{\bm w}_i,\bm w_i)+\sin^2\theta(\widehat{\bm v}_i,\bm v_i))} \le \frac{2\|\widehat{\bm C}-\bm C\|_{2}}{\widetilde{\gamma}_0},
	\end{align}
	where $\theta(\widehat{\bm w}_i,\bm w_i)$ is the canonical angle between the left singular vectors $\widehat{\bm w}_i$ and $\bm w_i$.
	We can easily see that 
	\begin{align}\label{eq:wedin_svdresult2}
		\max\{\sin\theta(\widehat{\bm w}_i,\bm w_i),\sin\theta(\widehat{\bm v}_i,\bm v_i) \} \le \sqrt{\sin^2\theta(\widehat{\bm w}_i,\bm w_i)+\sin^2\theta(\widehat{\bm v}_i,\bm v_i)} \le \sqrt{\sum_{i=1}^r(\sin^2\theta(\widehat{\bm w}_i,\bm w_i)+\sin^2\theta(\widehat{\bm v}_i,\bm v_i))} .
	\end{align}
Also, consider the below:
	\begin{align*}
	\|\widehat{\bm w}_i-\bm w\|^2_2 &= 2-2{ \widehat{\bm w}_i^{\top}\bm w}\\
	&\le 2(1-\cos\theta(\widehat{\bm w}_i,\bm w_i))\\
	&\le 2(1-\cos^2\theta(\widehat{\bm w}_i,\bm w_i))\\
	&= 2\sin^2\theta(\widehat{\bm w}_i,\bm w_i)\\
	\implies \|\widehat{\bm w}_i-\bm w\|_2 &\le \sqrt{2}\sin\theta(\widehat{\bm w}_i,\bm w_i).
	\end{align*}
The above inequality combined with \eqref{eq:wedin_svdresult} and \eqref{eq:wedin_svdresult2} gives the bound in Lemma \ref{lem:svds2}.
	
	\section{Proof of Lemma \ref{lem:alpha_feas}} \label{app:alpha_feas}
{	
The conditions on $\alpha_{(t)}$ given by \eqref{eq:cond_1} and \eqref{eq:newcond} can be re-written as:
\begin{align} 
 \max_{(j,k)}~[\bm H\bm E_{\bm Q_{(t)}}+\bm N\bm Q_{(t)}]_{j,k}\leq \alpha_{(t)} \leq \min_{(j,k) \in \bm \varDelta} [\bm H]_{j,k}+\min_{(j,k) \in \bm \varDelta}[\bm H\bm E_{\bm Q_{(t)}}+\bm N\bm Q_{(t)}]_{j,k}.\label{eq:alpha_init}
\end{align}
We can bound the term $\min_{(j,k) \in \bm \varDelta}[\bm H\bm E_{\bm Q_{(t)}}+\bm N\bm Q_{(t)}]_{j,k}$ as below:
\begin{align} \label{eq:alpha_bound1}
   \min_{(j,k) \in \bm \varDelta}[\bm H\bm E_{\bm Q_{(t)}}+\bm N\bm Q_{(t)}]_{j,k} &\ge -\max_{(j,k) \in \bm \varDelta}\left|[\bm H\bm E_{\bm Q_{(t)}}+\bm N\bm Q_{(t)}]_{j,k}\right| \ge -\max_{(j,k) }\left|[\bm H\bm E_{\bm Q_{(t)}}+\bm N\bm Q_{(t)}]_{j,k}\right|.
\end{align}
Using \eqref{eq:alpha_bound1}, we can re-write the conditions on $\alpha_{(t)}$ in \eqref{eq:alpha_init} as below:
\begin{align} \label{eq:alpha_bound2}
\max_{(j,k)}~|[\bm H\bm E_{\bm Q_{(t)}}+\bm N\bm Q_{(t)}]_{j,k}|\leq \alpha_{(t)} \leq \min_{(j,k) \in \bm \varDelta} [\bm H]_{j,k}-\max_{(j,k)}|[\bm H\bm E_{\bm Q_{(t)}}+\bm N\bm Q_{(t)}]_{j,k}|.
\end{align}
To proceed, we bound the term $\max_{(j,k)}~|[\bm H\bm E_{\bm Q_{(t)}}+\bm N\bm Q_{(t)}]_{j,k}|$ as below:
\begin{align}\label{eq:alpha_bound3}
    \max_{(j,k) }\left|[\bm H\bm E_{\bm Q_{(t)}}+\bm N\bm Q_{(t)}]_{j,k}\right| \le \|\bm H\bm E_{\bm Q_{(t)}}+\bm N\bm Q_{(t)}\|_{\rm F} \le \|\bm H\|_{\rm F}\|\bm E_{\bm Q_{(t)}}\|_{\rm F} + \nu,
\end{align}
where we used $\|\bm N\|_{\rm F}=\nu$ and the orthogonality of $\bm Q_{(t)}$ to obtain the last inequality.
Applying \eqref{eq:alpha_bound3} in \eqref{eq:alpha_bound2}, we can further re-write the conditions as:
\begin{align} \label{eq:alpha_bound4}
\|\bm H\|_{\rm F}\|\bm E_{\bm Q_{(t)}}\|_{\rm F} + \nu\leq \alpha_{(t)} \leq \min_{(j,k) \in \bm \varDelta} [\bm H]_{j,k}-\|\bm H\|_{\rm F}\|\bm E_{\bm Q_{(t)}}\|_{\rm F} - \nu.
\end{align}

Next, we proceed to bound $\|\bm E_{\bm Q_{(t)}}\|_{\rm F}$ using $\|\bm E_{\bm Q_{(0)}}\|_{\rm F}$. To accomplish this,
	we can recursively apply the results in \eqref{eq:EQ_expbound}  to obtain the below relation for any $t > 1$:
	\begin{align}
	\|\bm E_{\bm Q_{(t)}}\|^2_{\rm F} = \|\bm Q_{(t)}-\bm Q\|^2_{\rm F} &\le \rho^{t}\|\bm E_{\bm Q_{(0)}}\|^2_{\rm F} + \frac{ \nu^2}{\eta \|\bm H\|^2_{\rm F}}\sum_{q=1}^{t}\rho^q, \nonumber\\
	&= \rho^{t}\|\bm E_{\bm Q_{(0)}}\|^2_{\rm F} + \frac{  \nu^2(1-\rho^{t+1})}{\eta \|\bm H\|^2_{\rm F}(1-\rho)}.\label{eq:EQ_EH_bound_1}
	\end{align}

With the above result, we consider the following:
\begin{align}
	\|\bm E_{\bm Q_{(t)}}\|^2_{\rm F} - 	\|\bm E_{\bm Q_{(0)}}\|^2_{\rm F}  &\le \left(\rho^{t}\|\bm E_{\bm Q_{(0)}}\|^2_{\rm F} + \frac{ \nu^2}{\eta \|\bm H\|^2_{\rm F}}\sum_{q=1}^{t}\rho^q\right) - \|\bm E_{\bm Q_{(0)}}\|^2_{\rm F} \nonumber\\
	&= \left((\rho^{t}-1)\|\bm E_{\bm Q_{(0)}}\|^2_{\rm F} + \frac{ \nu^2}{\eta \|\bm H\|^2_{\rm F}}\sum_{q=1}^{t}\rho^q\right),\label{eq:EQ_diff}
\end{align}
 where we applied \eqref{eq:EQ_EH_bound_1} to get the first inequality.
If the R.H.S of \eqref{eq:EQ_diff} is smaller than zero, then we have $\|\bm E_{\bm Q_{(t)}}\|^2_{\rm F} \le 	\|\bm E_{\bm Q_{(0)}}\|^2_{\rm F}$. The condition to make the R.H.S of \eqref{eq:EQ_diff} smaller than zero can be written as below:
\begin{align}
(\rho^{t}-1)\|\bm E_{\bm Q_{(0)}}\|^2_{\rm F} &+ \frac{ \nu^2}{\eta \|\bm H\|^2_{\rm F}}\sum_{q=1}^{t}\rho^q \le 0 \nonumber\\
 \implies \frac{ \nu^2}{\eta \|\bm H\|^2_{\rm F}}\sum_{q=1}^{t}\rho^q &\le (1-\rho^{t})\|\bm E_{\bm Q_{(0)}}\|^2_{\rm F} \nonumber\\
\implies  \frac{ \nu^2}{\eta \|\bm H\|^2_{\rm F}}\frac{1}{1-\rho} &\le (1-\rho)\|\bm E_{\bm Q_{(0)}}\|^2_{\rm F} \nonumber\\
\implies \nu &\le (1-\rho)\sqrt{\eta} \|\bm H\|_{\rm F}\|\bm E_{\bm Q_{(0)}}\|_{\rm F}, \label{eq:cond_nu}
 \end{align}
 where the third inequality is obtained using the facts that $\sum_{q=1}^{t}\rho^q \le \sum_{q=1}^{\infty}\rho^q \le \frac{1}{1-\rho}$ and $1-\rho^t \ge 1-\rho$ since $\rho < 1$. It implies that if the conditions on $\nu$ given by \eqref{eq:cond_nu} is satisfied, 
 \begin{align} \label{eq:EQ_shrinking}
     \|\bm E_{\bm Q_{(t)}}\|^2_{\rm F} \le 	\|\bm E_{\bm Q_{(0)}}\|^2_{\rm F},\quad \forall t.
 \end{align}
 Applying \eqref{eq:EQ_shrinking} in \eqref{eq:alpha_bound4}, the condition on $\alpha_{(t)}$ can be further re-written as:
\begin{align} \label{eq:alpha_bound5}
\|\bm H\|_{\rm F}\|\bm E_{\bm Q_{(0)}}\|_{\rm F} + \nu\leq \alpha_{(t)} \leq \min_{(j,k) \in \bm \varDelta} [\bm H]_{j,k}-\|\bm H\|_{\rm F}\|\bm E_{\bm Q_{(0)}}\|_{\rm F} - \nu.
\end{align}
From \eqref{eq:alpha_bound5}, it is clear that we can find $\alpha_{(t)} = \alpha$ for every iteration $t$ as long as 
\begin{align*}
    \min_{(j,k) \in \bm \varDelta} [\bm H]_{j,k} > 2\|\bm H\|_{\rm F}\|\bm E_{\bm Q_{(0)}}\|_{\rm F} + 2\nu.
\end{align*}
}

	%

\end{document}